\documentclass[oneside,11pt]{article}

\usepackage{url,mathrsfs,algorithm}
\usepackage{amsmath,wrapfig,color}
\usepackage{amsfonts}
\usepackage{graphicx}
\usepackage{subfigure}

\begin{document}
\title{\huge Nystrom Method for Approximating the GMM Kernel}

\author{ \bf{Ping Li} \\
         Department of Statistics and Biostatistics\\
         Department of Computer Science\\
       Rutgers University\\
          Piscataway, NJ 08854, USA\\
       \texttt{pingli@stat.rutgers.edu}\\
}

\date{}

\maketitle

\begin{abstract}

\noindent The GMM (generalized min-max) kernel was recently proposed~\cite{Report:Li_GMM16} as a  measure of data similarity and was demonstrated  effective in  machine learning tasks. In order to use the GMM kernel for large-scale datasets, the prior work resorted to the (generalized) consistent weighted sampling (GCWS) to convert the GMM kernel to linear kernel.  We call this approach as ``GMM-GCWS''.  \\

\noindent  In the machine learning literature, there is a popular algorithm which we call ``RBF-RFF''. That is, one can use the ``random Fourier features'' (RFF) to convert the ``radial basis function'' (RBF) kernel to linear kernel. It was empirically shown in~\cite{Report:Li_GMM16} that RBF-RFF typically  requires substantially more samples than GMM-GCWS in order to achieve comparable accuracies. \\

\noindent The Nystrom method  is a general tool for computing nonlinear kernels, which again converts nonlinear kernels into linear kernels.  We apply the Nystrom method for approximating the GMM kernel, a strategy which we name as ``GMM-NYS''. In this study, our extensive experiments on a set of fairly large datasets confirm that GMM-NYS is also a strong competitor of RBF-RFF.

\end{abstract}

\section{Introduction}

The ``generalized min-max'' (GMM) kernel was recently proposed in~\cite{Report:Li_GMM16} as an effective measure of data similarity. Consider the original ($D$-dim) data vector $u_i$, $i=1$ to $D$.  The first step is to expand the data vector to a vector of $2D$ dimensions:
\begin{align}\label{eqn_transform}
 \left\{\begin{array}{cc}
\tilde{u}_{2i-1} = u_i,\hspace{0.1in} \tilde{u}_{2i} = 0&\text{if } \ u_i >0\\
\tilde{u}_{2i-1} = 0,\hspace{0.1in} \tilde{u}_{2i} =  -u_i &\text{if } \ u_i \leq 0
\end{array}\right.
\end{align}
For example, if $u = [2\ \ -1\ \ 3]$, then the transformed data vector becomes $\tilde{u} = [2\ \ 0\ \ 0\ \ 1\ \ 3\ \ 0]$. After the transformation, the GMM similarity between two vectors $u, v\in\mathbb{R}^D$ is defined as
\begin{align}\label{eqn_GMM}
&GMM(u,v) = \frac{\sum_{i=1}^{2D} \min(\tilde{u}_i,\ \tilde{v}_i)}{\sum_{i=1}^{2D} \max(\tilde{u}_i,\ \tilde{v}_i)}
\end{align}

It was shown in~\cite{Report:Li_GMM16}, through extensive experiments on a large collection of publicly available datasets, that using the GMM kernel can often produce excellent results in classification tasks. On the other hand, it is generally  nontrivial to scale nonlinear kernels for  large data~\cite{Book:Bottou_07}. In a sense, it is not  practically meaningful to discuss nonlinear kernels without knowing how to compute them efficiently (e.g., via hashing).  \cite{Report:Li_GMM16} focused on  the generalized consistent weighted sampling (GCWS).

\subsection{Generalized Consistent Weighted Sampling (GCWS) and 0-bit GCWS}

Algorithm~\ref{alg_GCWS} summarizes the ``(0-bit) generalized consistent weighted sampling'' (GCWS).  Given two  data vectors  $u$ and $v$, we transform them into nonnegative vectors  $\tilde{u}$ and $\tilde{v}$ as in (\ref{eqn_transform}). We then  apply the  ``(0-bit) consistent weighted sampling'' (0-bit CWS)~\cite{Report:Manasse_CWS10,Proc:Ioffe_ICDM10,Proc:Li_KDD15} to generate  random integers: $i^*_{\tilde{u},j},\  i^*_{\tilde{v},j}$, $j=1, 2, ..., k$. According to the result  in~\cite{Proc:Li_KDD15}, the following approximation
\begin{align}\label{eqn_GCWS_Prob}
\mathbf{Pr}\left\{i^*_{\tilde{u},j} =  i^*_{\tilde{v},j}\right\} \approx GMM({u},{v})
\end{align}
is  accurate in practical settings and makes the implementation  convenient.

\begin{algorithm}{
\textbf{Input:} Data vector $u$ = ($i=1$ to $D$)

Generate vector $\tilde{u}$ in $2D$-dim by (\ref{eqn_transform})

\vspace{0.08in}

For $i$ from 1 to $2D$

\hspace{0.25in}$r_i\sim Gamma(2, 1)$, \ $c_i\sim Gamma(2, 1)$,  $\beta_i\sim Uniform(0, 1)$

\hspace{0.2in} $t_i\leftarrow \lfloor \frac{\log \tilde{u}_i }{r_i}+\beta_i\rfloor$, \ $z_i\leftarrow \exp(r_i(t_i - \beta_i))$,\  $a_i\leftarrow c_i/(z_i \exp(r_i))$

End For

\textbf{Output:} $i^* \leftarrow arg\min_i \ a_i$
}\caption{(0-Bit) Generalized Consistent Weighted Sampling (GCWS)}
\label{alg_GCWS}
\end{algorithm}

For each data vector $u$, we obtain $k$ random samples $i^*_{\tilde{u},j}$, $j=1$ to $k$. We store only the lowest $b$ bits of $i^*$, based on the idea of~\cite{Proc:HashLearning_NIPS11}. We need to view those $k$ integers as locations (of the nonzeros) instead of numerical values. For example, when $b=2$, we should view $i^*$ as a  vector of length $2^b=4$. If $i^*=3$, then we code it as $[1\ 0\ 0\ 0]$; if $i^*=0$, we code it as $[0\ 0\ 0\ 1]$, etc. We  concatenate all $k$ such vectors into a binary vector of length $2^b\times k$, which contains exactly $k$ 1's.  After we have generated such new data vectors for all data points, we feed them to a linear SVM or  logistic regression solver. We can, of course,  also use the new data for many other tasks including clustering, regression, and near neighbor search.

\vspace{0.1in}

Note that for linear learning methods, the storage and computational cost is largely determined by the number of nonzeros in each data vector, i.e., the $k$ in our case. It is thus crucial not to use a too large $k$. For the other parameter $b$, we recommend to use a fairly large value if possible.

\subsection{The RBF Kernel and Random Fourier Features (RFF)}

The natural competitor of the GMM kernel is the RBF (radial basis function) kernel, whose definition involves a crucial tuning parameter $\gamma>0$.  In this study, for convenience (e.g.,  parameter tuning), we use the following version of the RBF kernel:
\begin{align}
RBF(u,v;\gamma) = e^{-\gamma(1-\rho)},\hspace{0.5in} \text{where } \
\rho = \rho(u,v) = \frac{\sum_{i=1}^D u_i v_i }{\sqrt{\sum_{i=1}^D u_i^2}\sqrt{\sum_{i=1}^Dv_i^2}}
\end{align}
Based on Bochner’s Theorem~\cite{Book:Rudin_90}, it is known~\cite{Proc:Rahimi_NIPS07} that, if we sample $w\sim uniform(0,2\pi)$, $r_{i}\sim N(0,1)$ i.i.d., and let  $x = \sum_{i=1}^D u_i r_{ij}$, $y = \sum_{i=1}^D v_i r_{ij}$, where $\|u\|_2=\|v\|_2=1$, then we have
\begin{align}\label{eqn_RFF}
E\left(\sqrt{2}\cos(\sqrt{\gamma} x+w)\sqrt{2}\cos(\sqrt{\gamma} y+w)\right)
= e^{-\gamma(1-\rho)}
\end{align}
This provides an elegant mechanism for linearizing the RBF kernel and the so-called RFF method has become  popular in machine learning, computer vision, and beyond.  

It turns out that, for nonnegative data, one can simplify (\ref{eqn_RFF}) by removing the random variable  $w$, due to the following fact:
\begin{align}\label{eqn_fRBF}
E\left(\cos(\sqrt{\gamma} x)\cos(\sqrt{\gamma} y)\right)
= \frac{1}{2}e^{-\gamma(1-\rho)}+\frac{1}{2}e^{-\gamma(1+\rho)} = fRBF(u,v;\gamma)
\end{align}
which is monotonic when $\rho\geq 0$. This creates a new  nonlinear kernel called ``folded RBF'' (fRBF).\\

A major issue with the RFF method is the  high variance. Typically a large number of samples (i.e., large $k$) would be needed in order to reach a satisfactory accuracy, as validated in~\cite{Report:Li_GMM16}. Usually, ``GMM-GCWS'' (i.e., the GCWS algorithm for approximating the GMM kernel) requires substantially fewer samples than ``RBF-RFF'' (i.e., the RFF method for approximating the RBF kernel).\\

 In this paper, we will introduce the Nystrom method~\cite{Article:Nystrom1930} for approximating the GMM kernel, which we call ``GMM-NYS''. We will show that GMM-NYS is also a strong competitor of RBF-RFF.

\section{The Nystrom Method for Kernel Approximation}

The Nystrom method~\cite{Article:Nystrom1930} is a  sampling  scheme for kernel approximation~\cite{Proc:Willimas_NIPS01}. For example, \cite{Proc:Yang_NIPS12} applied the Nystrom method for approximating the RBF kernel, which we call ``RBF-NYS''. Analogously, we propose  ``GMM-NYS'', which is the use of the Nystrom method for approximating the GMM kernel. This paper will show that GMM-NYS is  a strong competitor of RBF-RFF.

To help interested readers repeat our experiments, here  we post the matlab script for generating $k$ samples using the Nystrom method. This piece of code contains various small tricks to make the implementation fairly efficient without hurting its readability.
\begin{verbatim}
%%%%%%%%%%%%%%%%%%%%%%%%%%%%%%%%%%%%%%%%%%
function Hash = GenNysGmm(k, X1, X2)
% k = number of samples
% X1 = data matrix to be sampled from
% X2 = data matrix to be hashed

Xs = X1(randsample(size(X1,1),k),:);
KernelXs = zeros(size(Xs,1),size(Xs,1));
for i = 1:size(Xs,1)
    U = sparse(ones(size(Xs,1),1))*Xs(i,:);
    Min = min(U, Xs);    Max = max(U, Xs);
    KernelXs(i,:) = sum(Min,2)./(sum(Max,2) + eps);
end

[v,d] = eig(KernelXs); T = inv(d.^0.5)*v';
Kernel = zeros(size(Xs,1),size(X2,1));
for i = 1:size(Xs,1)
    U = sparse(ones(size(X2,1),1))*Xs(i,:);
    Min = min(U, X2);    Max = max(U, X2);
    Kernel(i,:) = sum(Min,2)./(sum(Max,2) + eps);
end
Hash = Kernel'*T';
%%%%%%%%%%%%%%%%%%%%%%%%%%%%%%%%%%%%%%%%%%%
\end{verbatim}

Firstly, we randomly sample $k$ data points from the (training) data matrix. Then we generate a $k\times k$ data kernel matrix and compute its eigenvalues and eigenvectors. We then produce a new representation of given data matrix based on the eigenvalues and eigenvectors of the sampled kernel matrix. The new representation will be of exactly $k$ dimensions (i.e., $k$ nonzeros per data point). \\

In this paper, we will show that GMM-NYS is a strong competitor of RBF-RFF. This is  actually not surprising. Random projection based algorithms always have (very) high variances and often do not perform as well as sampling based algorithms~\cite{Proc:Li_Church_Hastie_NIPS06}.

\section{Experiments}

We provide an extensive experimental study on a collection of fairly large classification datasets which are publicly available. We report the classification results for four methods: 1) GMM-GCWS, 2) GMM-NYS, 3) RBF-NYS,  4) RBF-RFF. Even though we focus on reporting classification results,  we should mention that our methods generate new data presentations and can be used for (e.g.,) classification, regression, clustering.  Note that, due to the discrete nature of the hashed values, GMM-GCWS can also be directly used for building hash tables and efficient near neighbor search.

\subsection{Datasets}

Table~\ref{tab_data} summarizes the datasets for our experimental study. Because the Nystrom method is a sampling algorithm, we know that it will recover the original kernel result if the number of samples ($k$) approaches the number of training examples. Thus, it makes sense to compare the algorithms on  larger datasets. Nevertheless, we still provide the experimental reuslt on {\em SVMGuide3}, a tiny dataset with merely $1,243$ training data points, merely for a sanity check.

\begin{table}[h!]
\caption{\textbf{Public datasets  and kernel SVM results}. The datasets were downloaded  from the UCI repository or the LIBSVM website. For the first 3 datasets (which are small enough), we report the test classification accuracies for the linear kernel, the RBF kernel (with the best $\gamma$ value in parentheses), and the GMM kernel, at the best SVM $l_2$-regularization  $C$ values. For GMM and RBF, We use LIBSVM pre-computed kernel functionality. For the other datasets, we only report the linear SVM results and the best $\gamma$ values obtained from a sub-sample of each dataset.  }
\begin{center}{
{\begin{tabular}{l r r r c l c}
\hline \hline
Dataset     &\# train  &\# test  &\# dim &linear (\%) &RBF (\%) ($\gamma$) &GMM (\%)\\
\hline
SVMGuide3 &1,243 &41 &21 &36.5 &\textbf{100} (120) &\textbf{100}\\
Letter &15,000 &5,000 &16 &61.7 &\textbf{97.4} (11) &97.3\\
Covertype25k &25,000 &25,000 &54 &71.5 &\textbf{84.7} (150)   &84.5\\
SensIT &78,823 &19,705 &100 &80.5 &-- (0.1) &--\\
Webspam &175,000 &175,000 &254 &93.3 & -- (35) &--\\
PAMAP105 &185,548 &185,548 &51 &83.4  & -- (18)  & --\\
PAMAP101 &186,581 &186,580 &51 &79.2  & -- (1.5)  & -- \\
Covertype &290,506 &290,506 &54 &71.3  & -- (150)  & --\\
RCV1 &338,699 &338,700 &47,236 &97.7 &-- (2.0) & --\\
\hline\hline
\end{tabular}}
}
\end{center}\label{tab_data}
\end{table}

When using modern linear algorithms (especially online learning), the storage and computational cost are mainly determined by the number of nonzero entries per data point. In our study, after hashing (sampling), the storage and computational cost are mainly dominated by $k$, the number of samples. We will report the experimental results for $k \in\{32, 64, 128, 256, 512, 1024\}$, as we believe that for most practical applications, $k\geq 1024$ would not be so desirable (and takes too much time/space to complete the experiments). Nevertheless, for RCV1, we also report the  results for $k\in\{2048, 4096\}$, for the comparison purpose (and for our curiosity). We always use the LIBLINEAR package~\cite{Article:Fan_JMLR08} for training linear SVMs on the original data as well as the hashed data.

\subsection{Experimental Results}

\begin{figure}[h!]
\begin{center}
\mbox{
\includegraphics[width=2.3in]{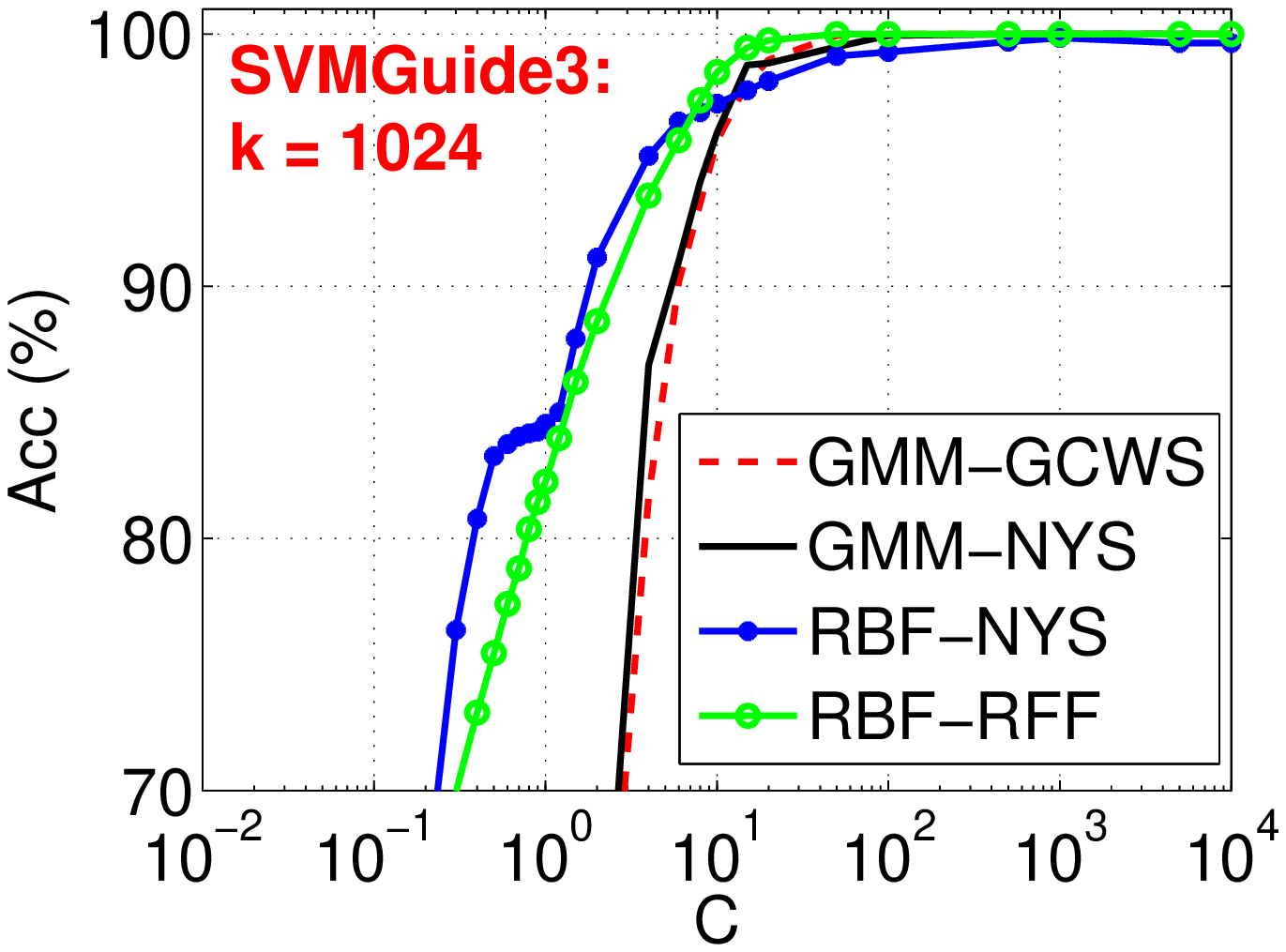}\hspace{-0.12in}
\includegraphics[width=2.3in]{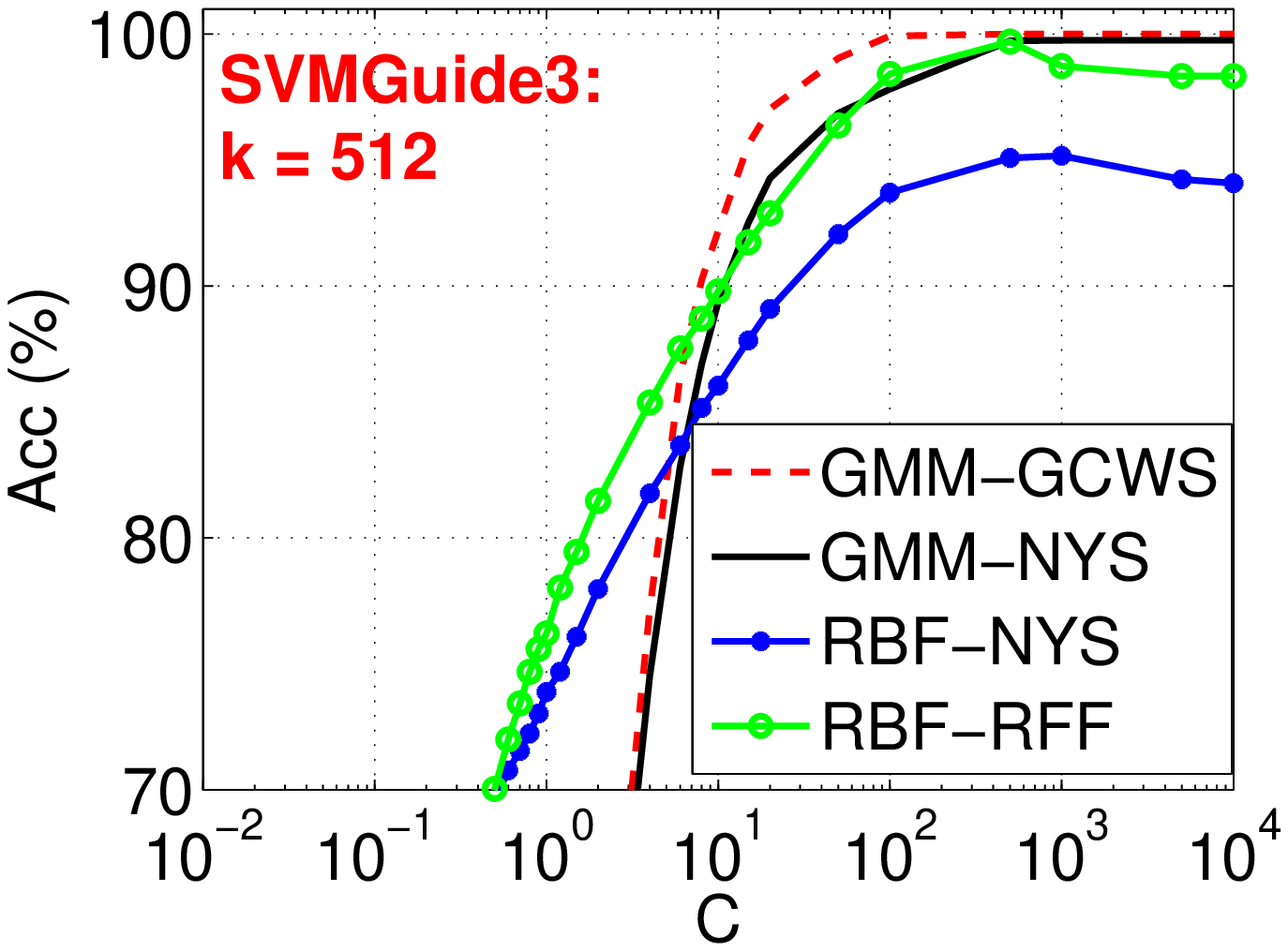}\hspace{-0.12in}
\includegraphics[width=2.3in]{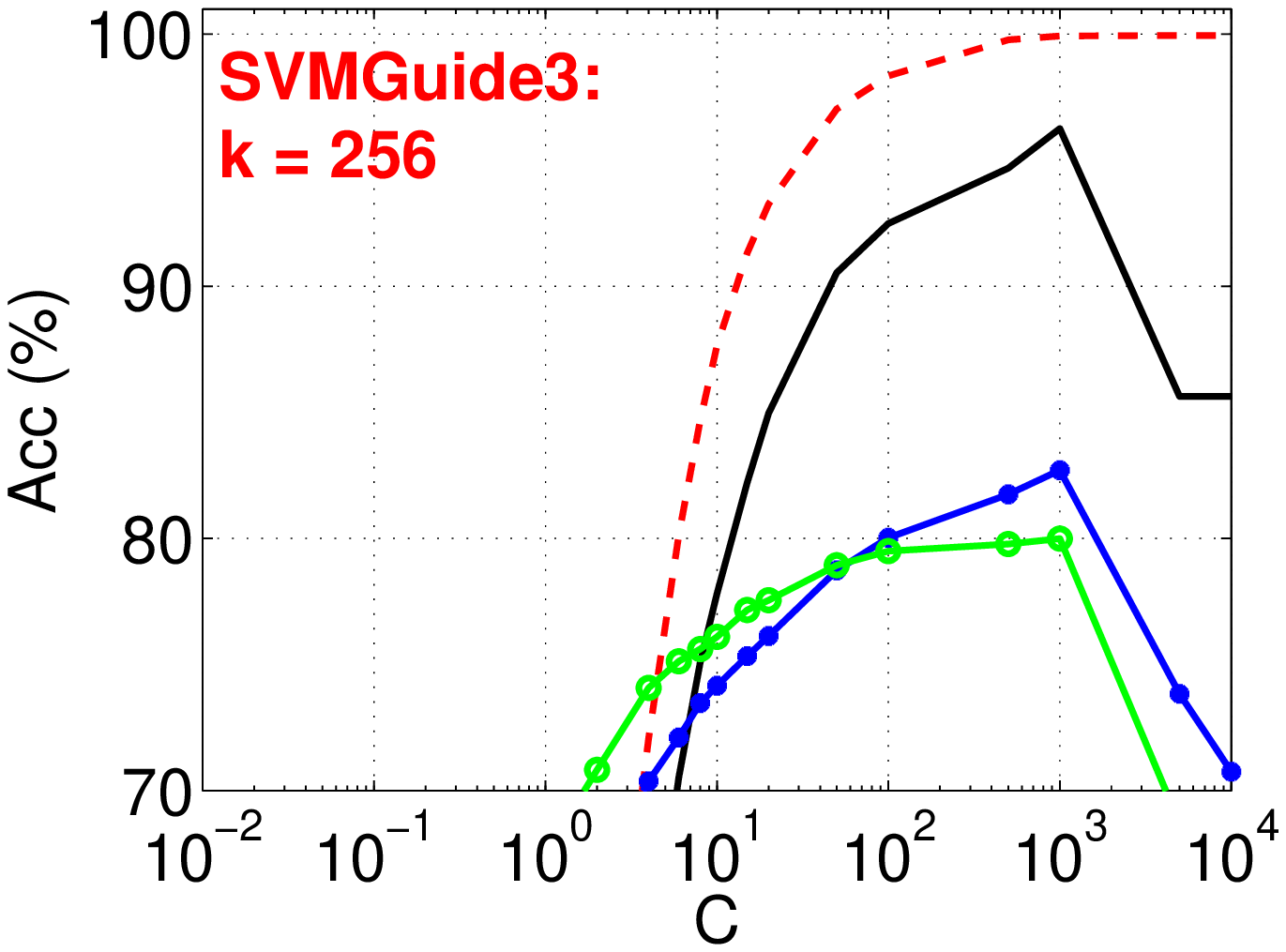}
}

\mbox{
\includegraphics[width=2.3in]{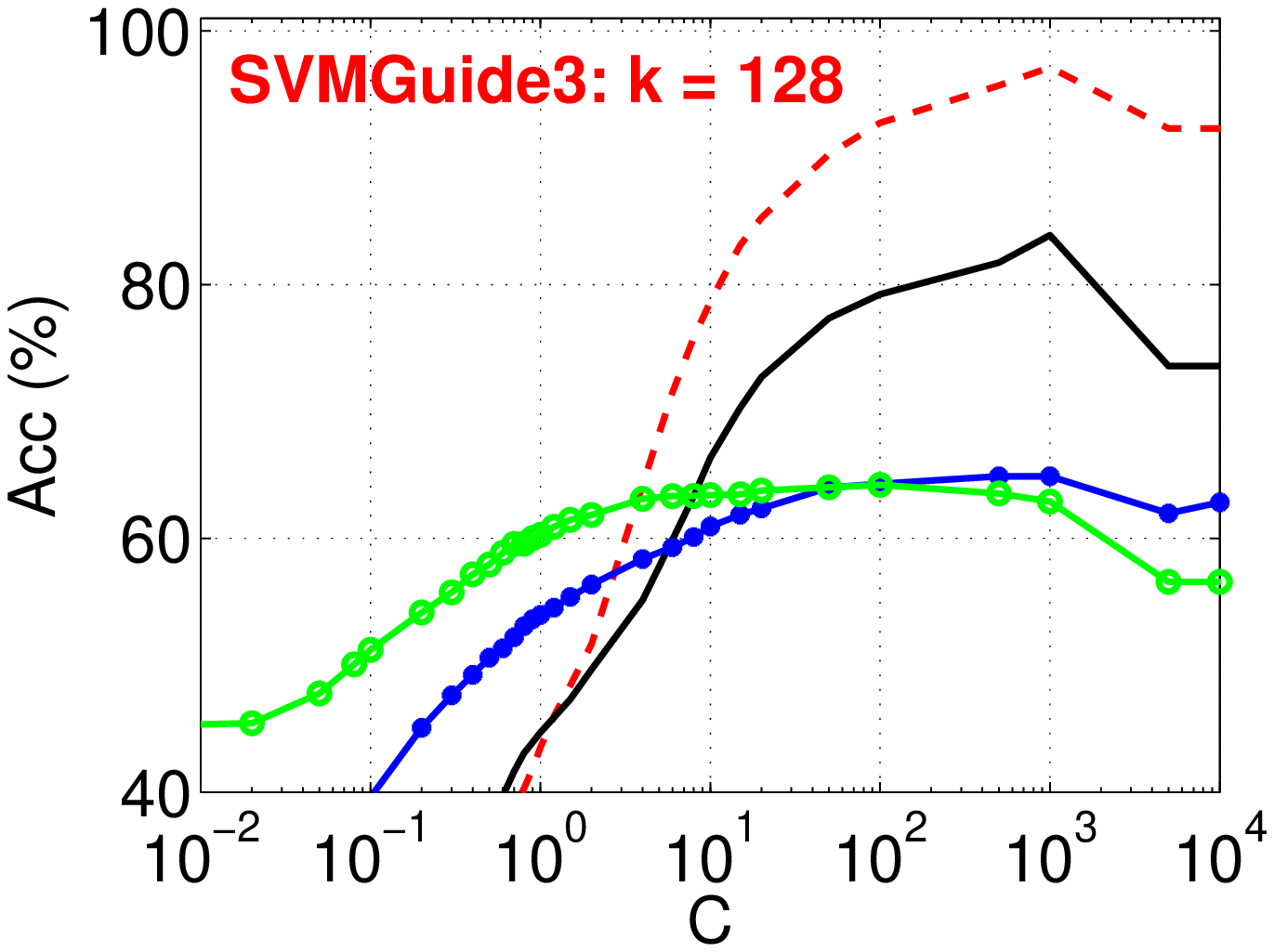}\hspace{-0.12in}
\includegraphics[width=2.3in]{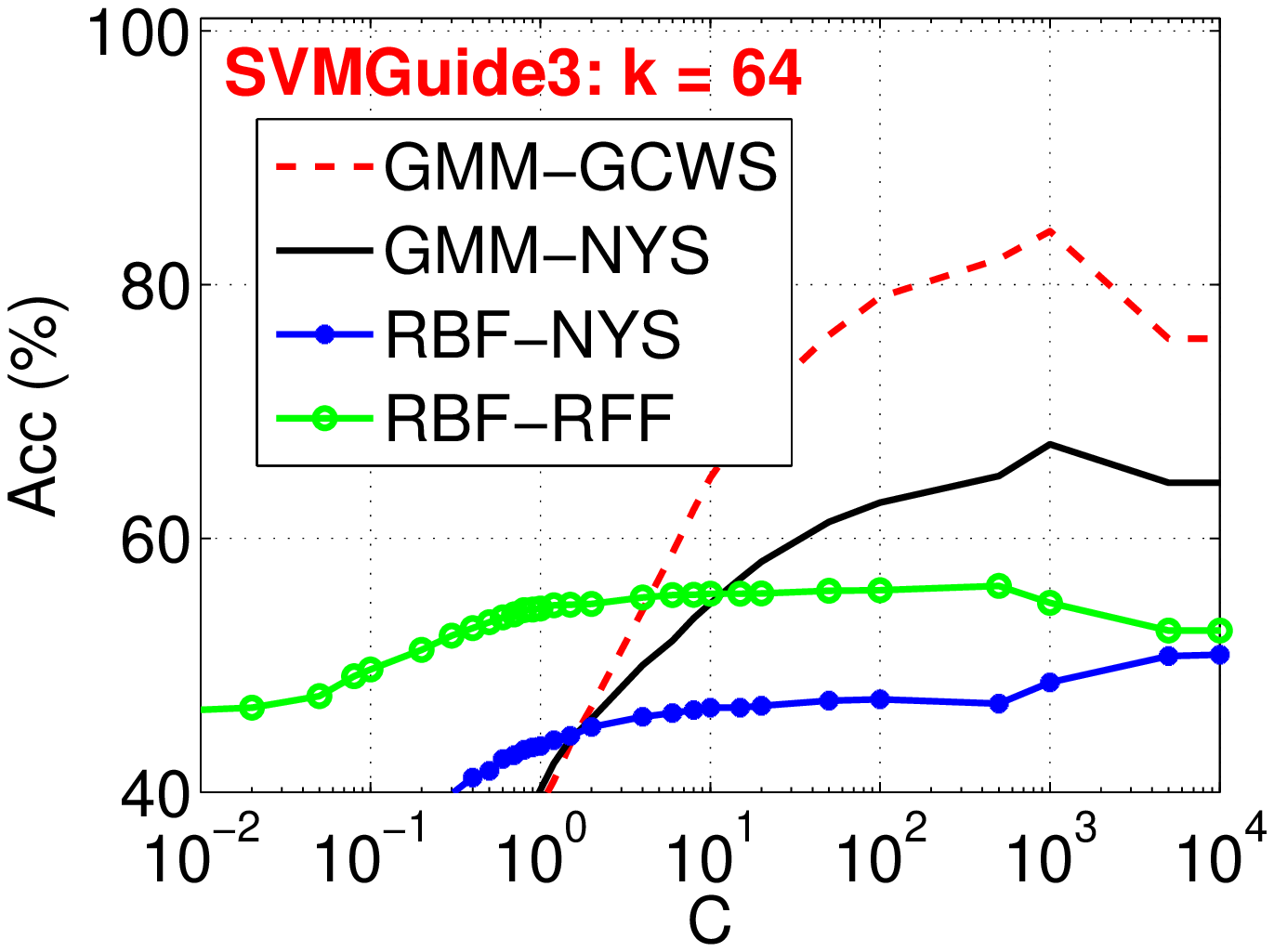}\hspace{-0.12in}
\includegraphics[width=2.3in]{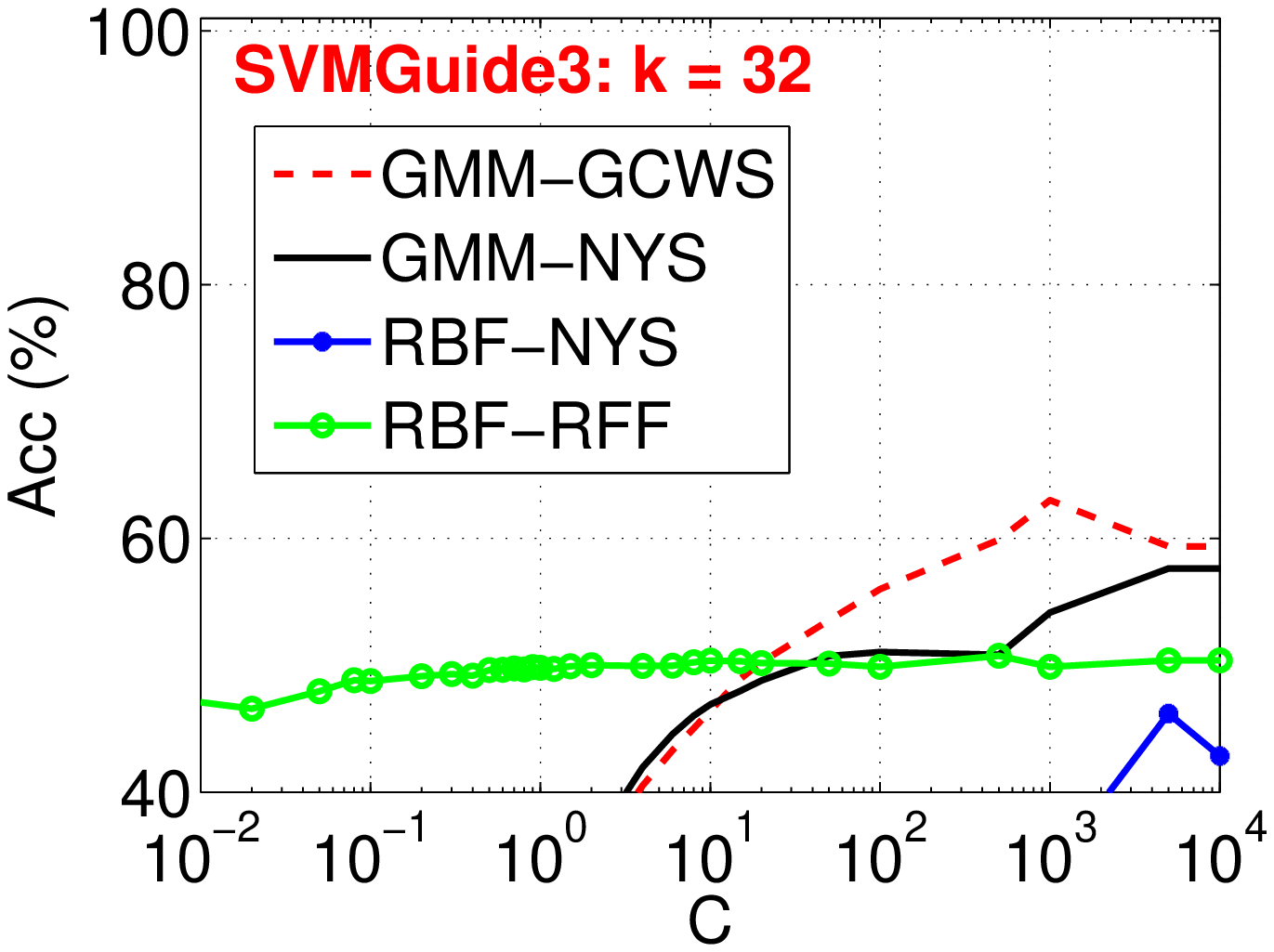}
}
\end{center}
\vspace{-0.3in}
\caption{\textbf{SVMGuide3:}\ Test classification accuracies for 6 different $k$ values and 4 different hashing algorithms: 1) GMM-GCWS, 2) GMM-NYS, 3) RBF-NYS,  4) RBF-RFF. After the hashed data are generated, we use  the LIBLINEAR package~\cite{Article:Fan_JMLR08} for training linear SVMs for a wide range of $l_2$-regularization $C$ values (i.e., the x-axis). The classification results are averaged from 100 repetitions (at each $k$ and $C$). We can see that, at the same sample size $k$ (and when $k$ is not too large), GMM-NYS produces substantially more accurate results than RBF-RFF. }\label{fig_SVMGuide3}\vspace{0.1in}
\end{figure}

Figure~\ref{fig_SVMGuide3} reports the test classification accuracies on  {\em SVMGuide3}, for 6 different samples sizes ($k$) and 4 different algorithms:  1) GMM-GCWS, 2) GMM-NYS, 3) RBF-NYS,  4) RBF-RFF. Again, we should emphasize that the storage and computational cost are largely determined by the sample size $k$, which is also the number of nonzero entries per data vector in the transformed space.  \\

Because the Nystrom method is based on sampling, we know that if $k$ is large enough, the performance will approach that of the original kernel method. In particular, for this tiny dataset, since there are only 1,243 training data points, it is expected that when $k=1024$, the classification results of GMM-NYS and RBF-NYS should be (close to) $100\%$, as validated in the upper left panel of Figure~\ref{fig_SVMGuide3}. It is thus more meaningful to examine the classification results for smaller $k$ values.

From Figure~\ref{fig_SVMGuide3}, it is obvious that when $k$ is not too large, GMM-NYS performs substantially better than RBF-RFF.  Note that in order to show reliable (and smooth) curves, for this (tiny) dataset, we repeat each experiment (at each $k$ and each $C$) 100 times and we report the averaged results. For other datasets, we report the averaged results from 10 repetitions.\\

Note that for {\em SVMGuide3}, the original classification accuracy using linear SVM is  low ($36.5\%$, see Table~\ref{tab_data}). These  hashing methods produce substantially better results even when $k=32$ only.\\

\begin{figure}[h!]
\begin{center}
\mbox{
\includegraphics[width=2.3in]{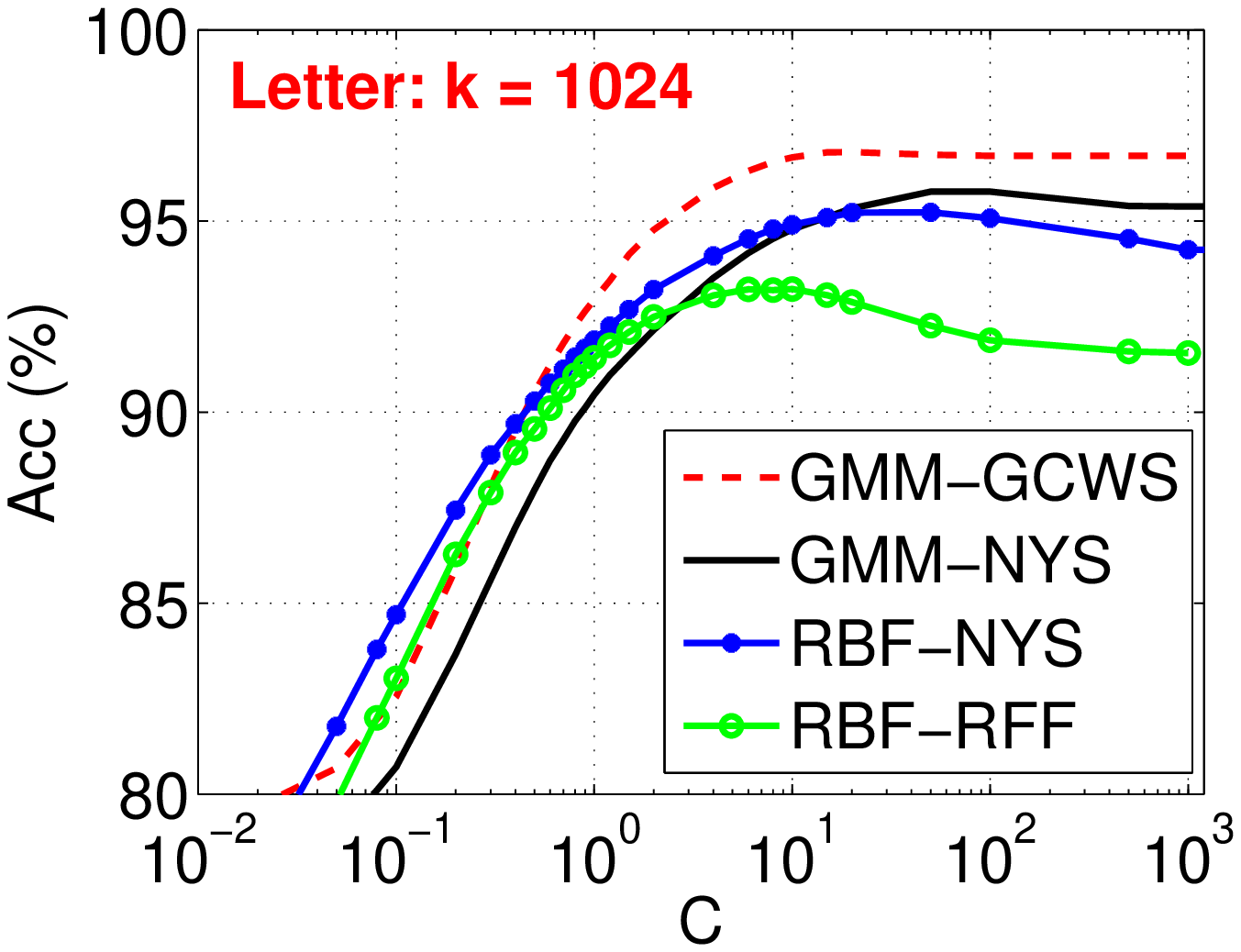}\hspace{-0.12in}
\includegraphics[width=2.3in]{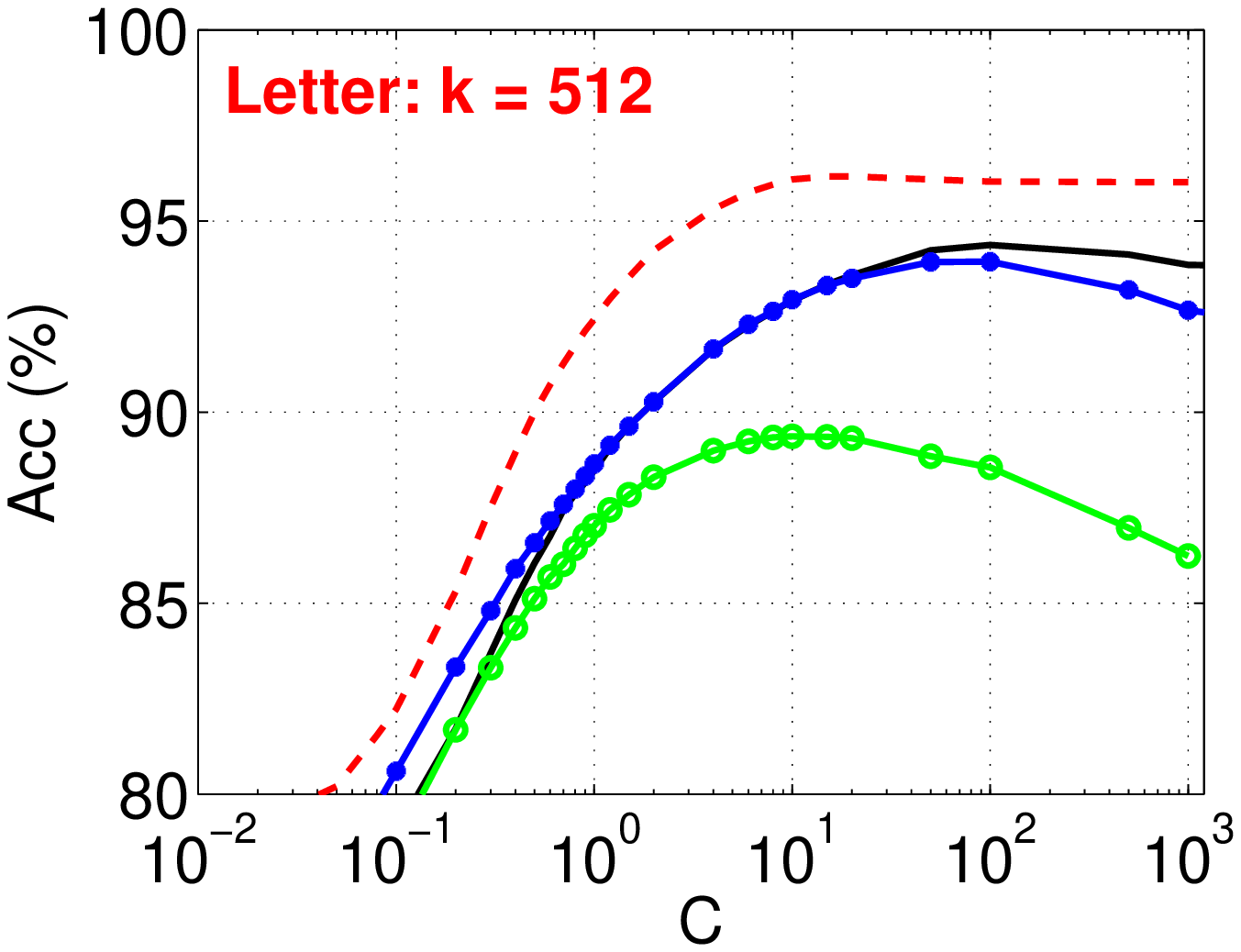}\hspace{-0.12in}
\includegraphics[width=2.3in]{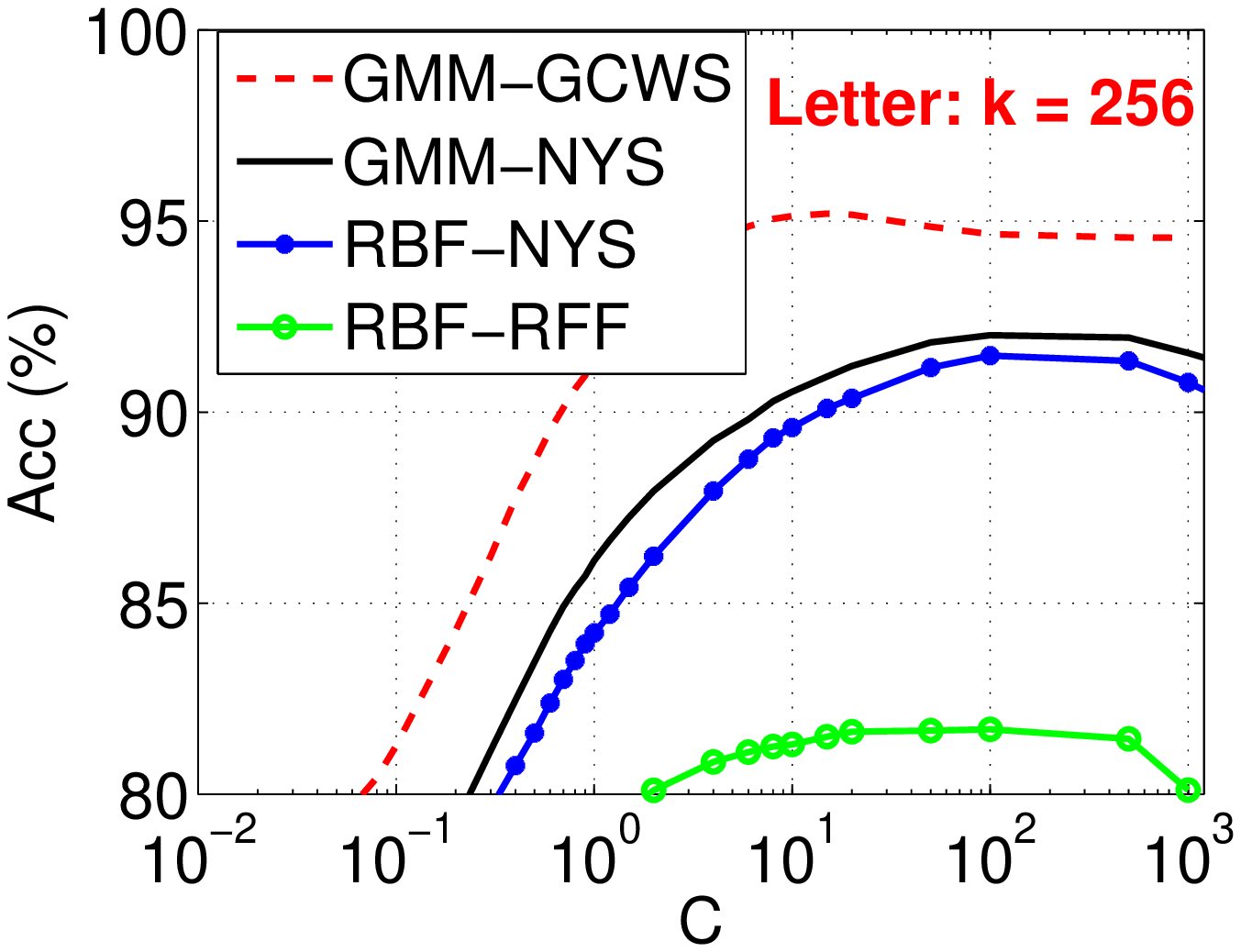}
}

\mbox{
\includegraphics[width=2.3in]{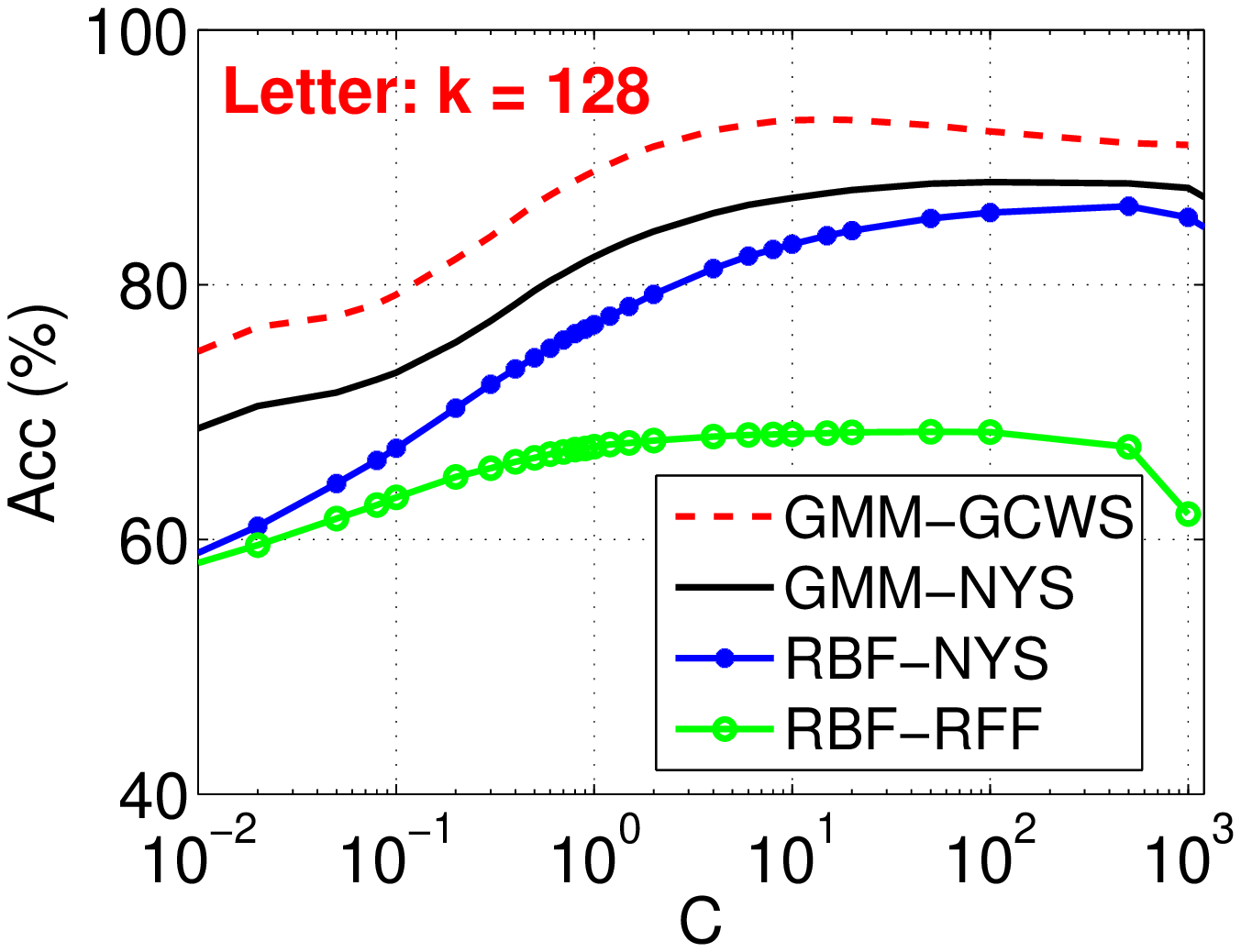}\hspace{-0.12in}
\includegraphics[width=2.3in]{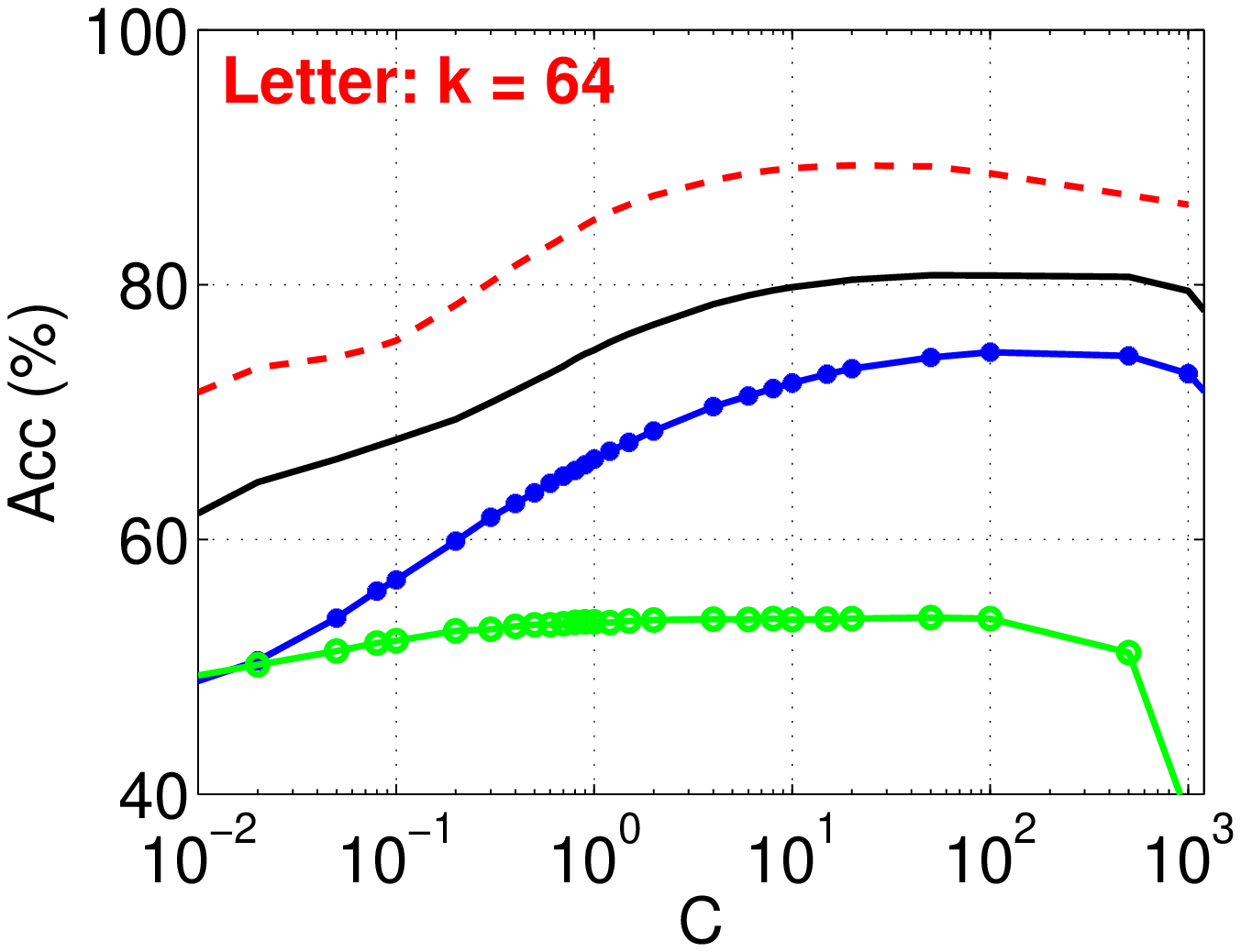}\hspace{-0.12in}
\includegraphics[width=2.3in]{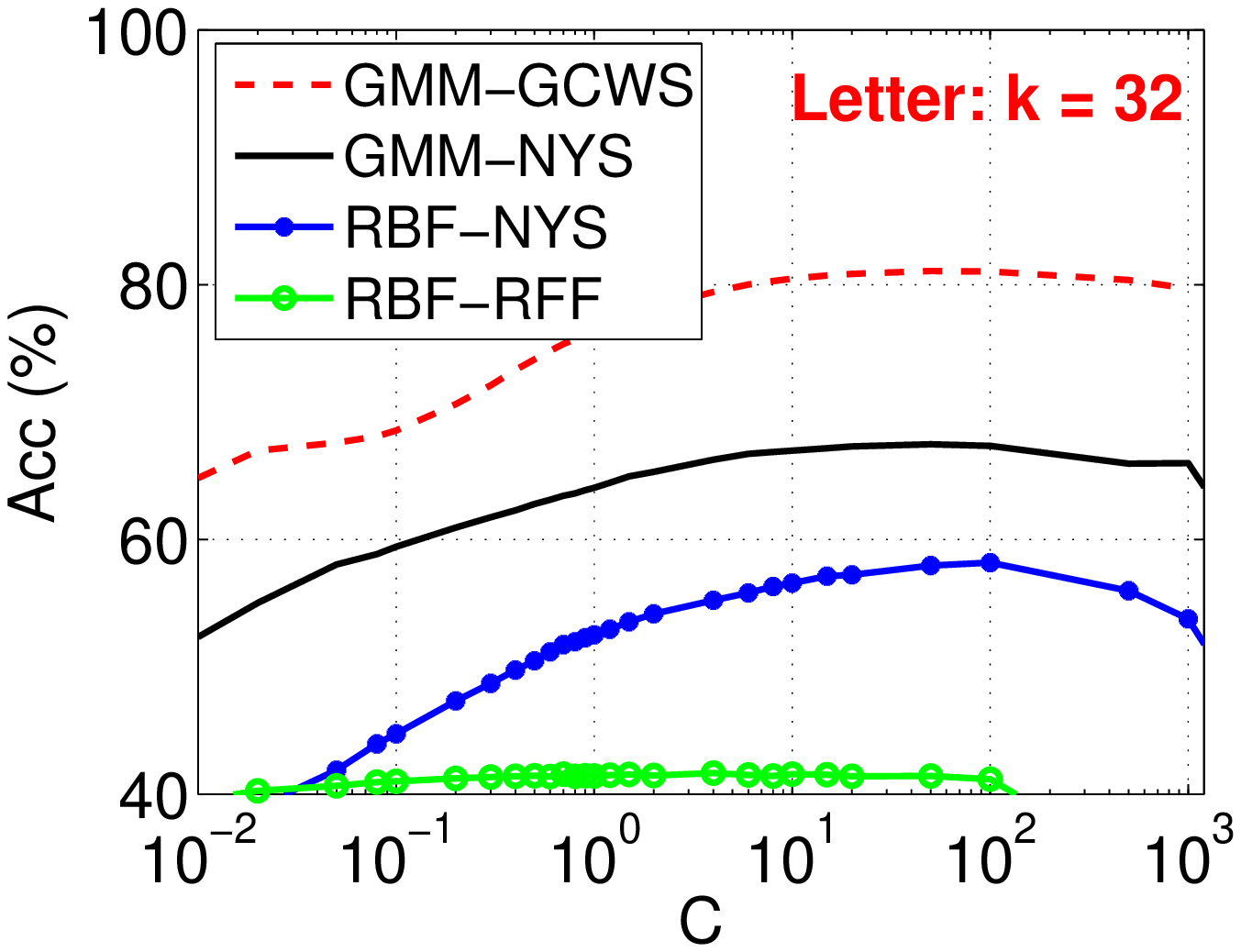}
}
\end{center}
\vspace{-0.3in}
\caption{\textbf{Letter:}\ Test classification accuracies for 6 different $k$ values and 4 different  algorithms.}\label{fig_Letter}\vspace{0.15in}
\end{figure}

Figure~\ref{fig_Letter} reports the test classification results for {\em Letter}, which also confirm that GMM-NYS produces substantially more accurate results than RBF-RFF.  Again, while the original test classification accuracy using linear SVM is low ($61.7\%$, see Table~\ref{tab_data}), GMM-NYS with merely $k=32$ samples already becomes more accurate.

\begin{figure}
\begin{center}
\mbox{
\includegraphics[width=2.3in]{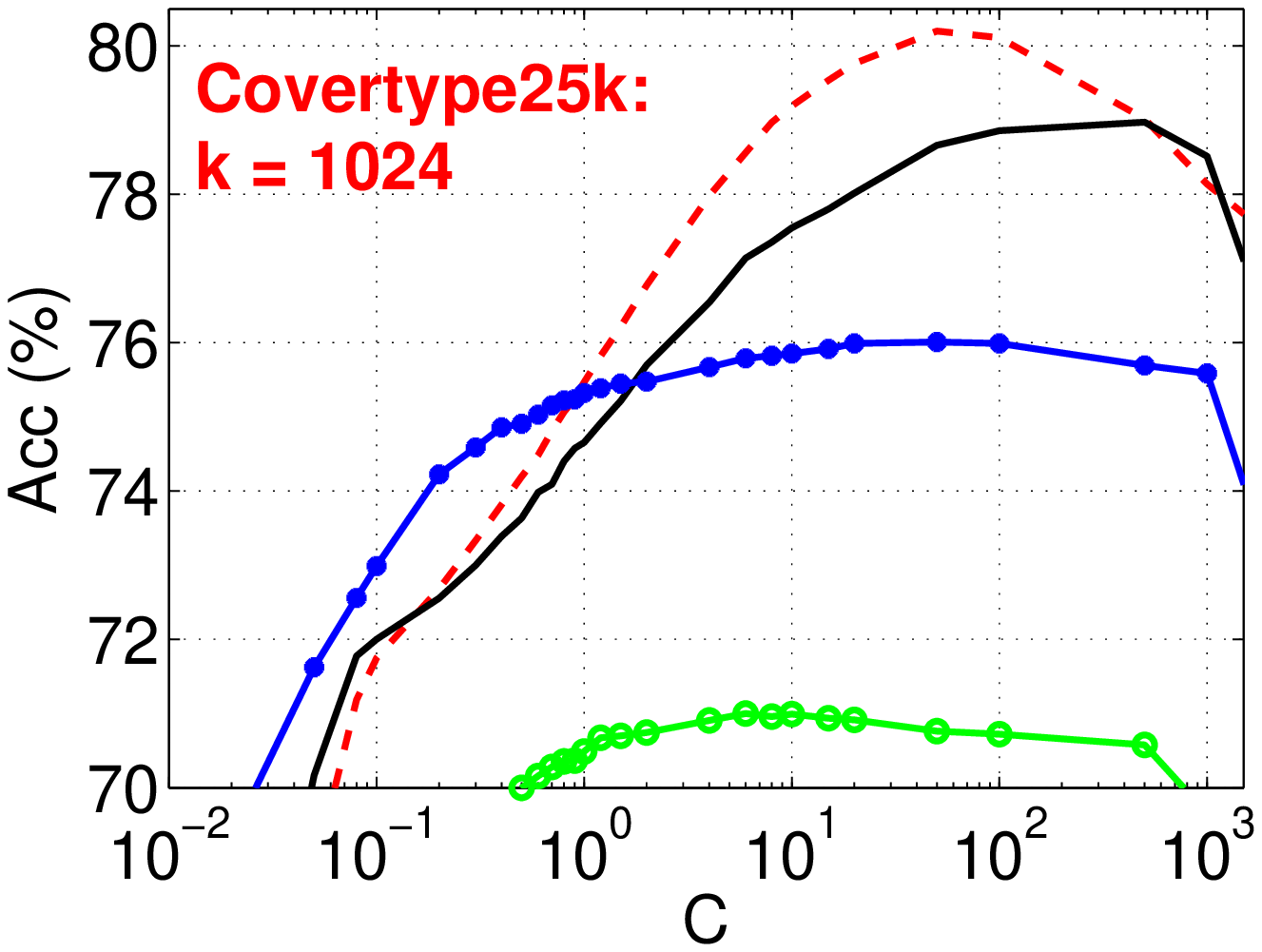}\hspace{-0.12in}
\includegraphics[width=2.3in]{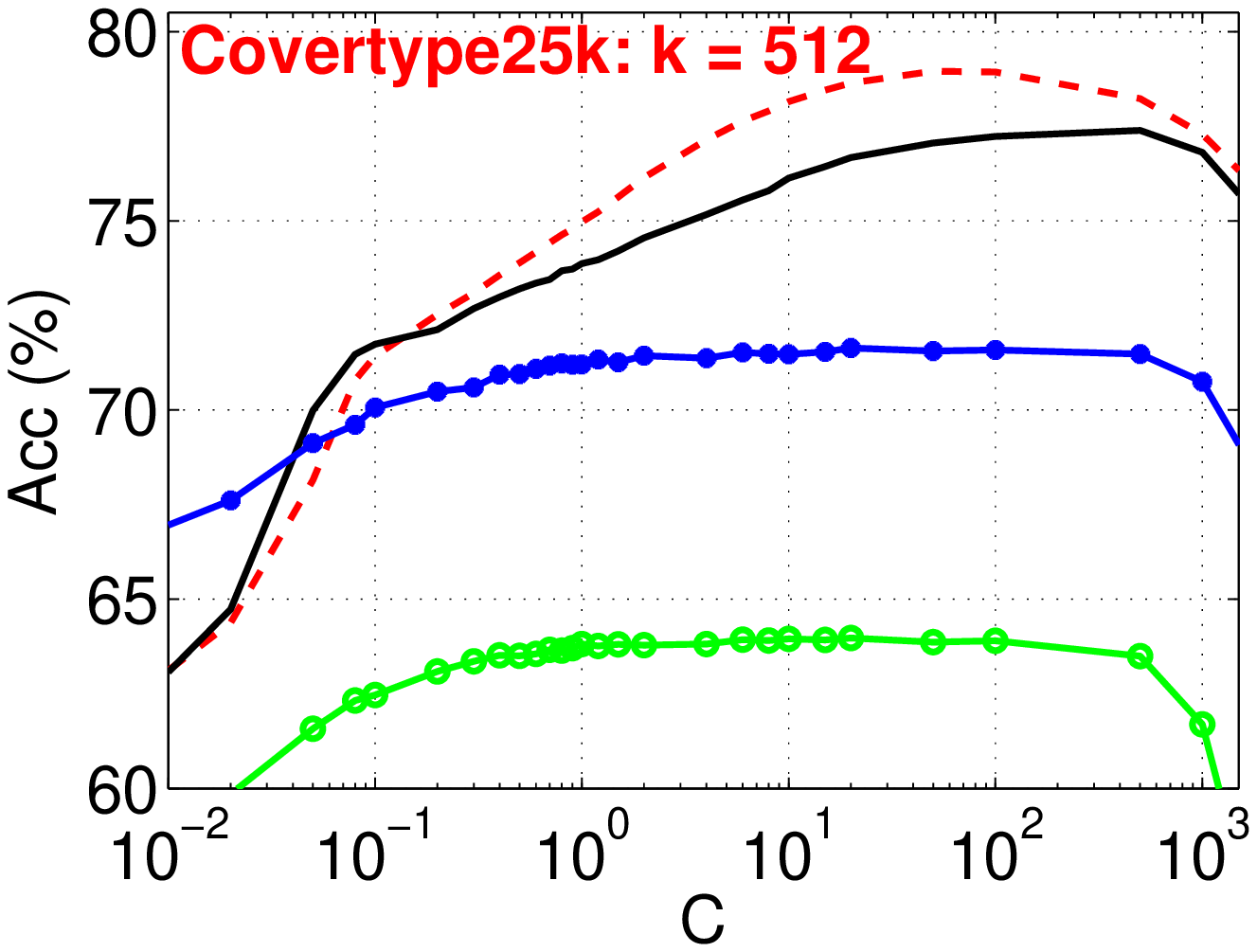}\hspace{-0.12in}
\includegraphics[width=2.3in]{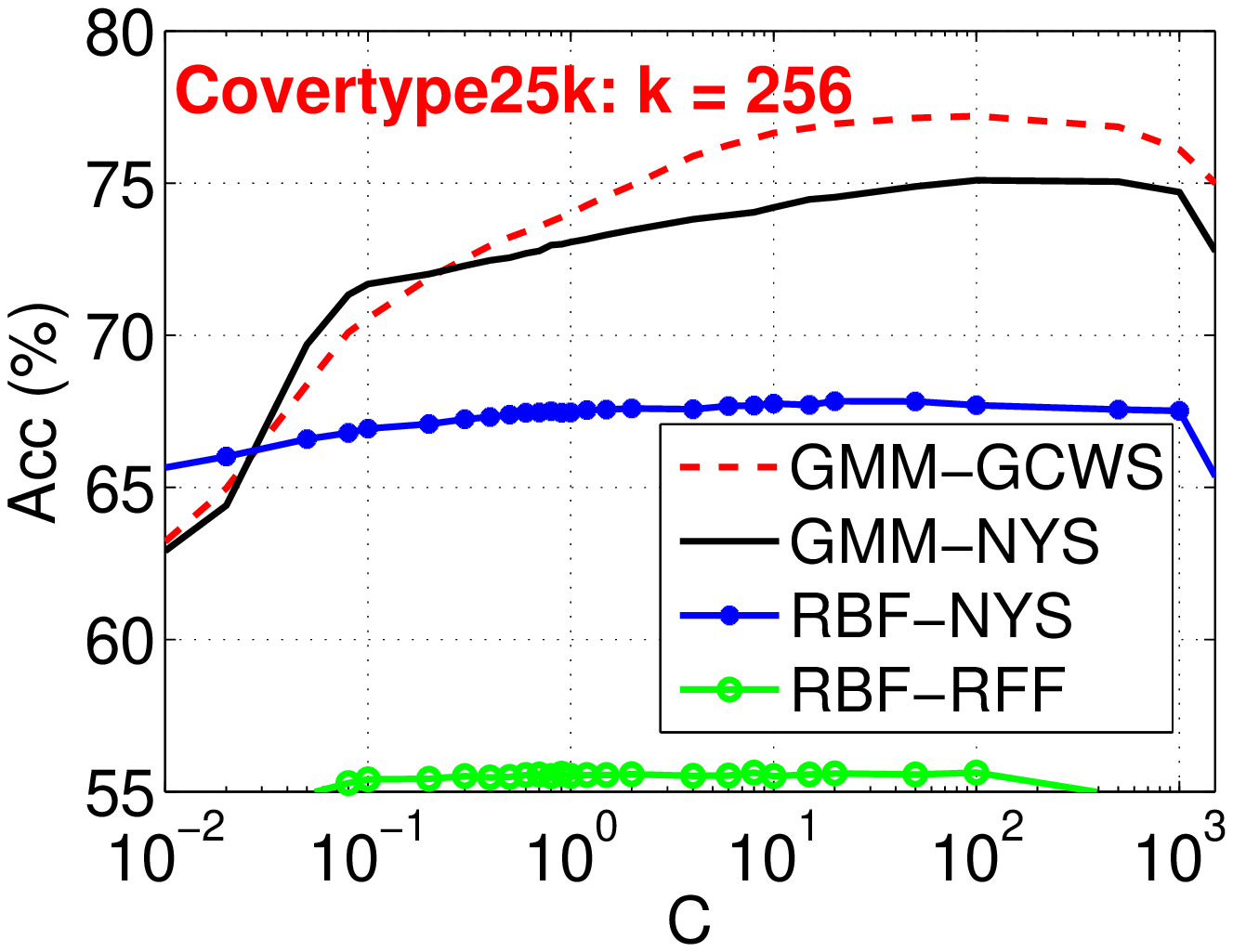}
}

\mbox{
\includegraphics[width=2.3in]{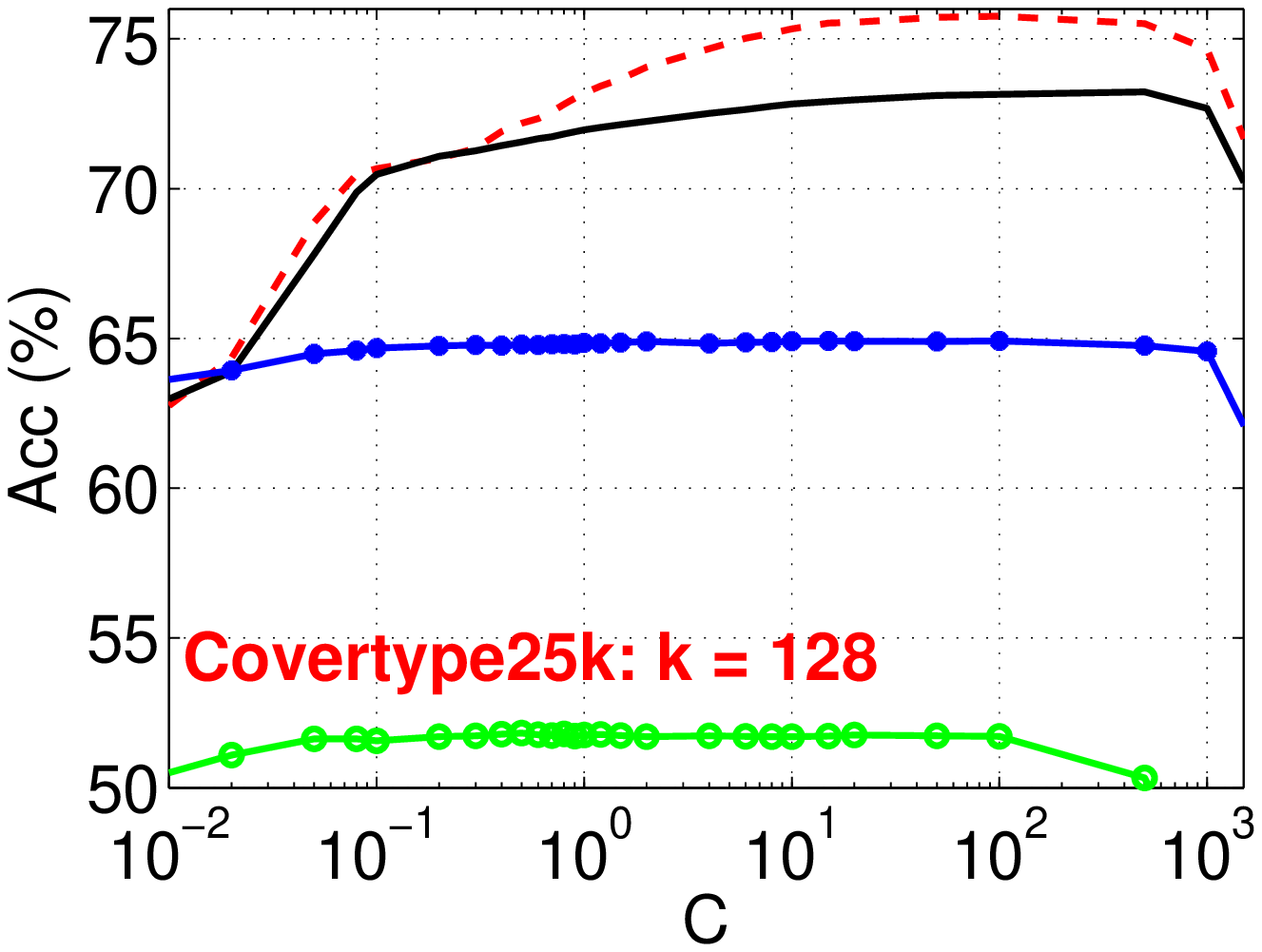}\hspace{-0.12in}
\includegraphics[width=2.3in]{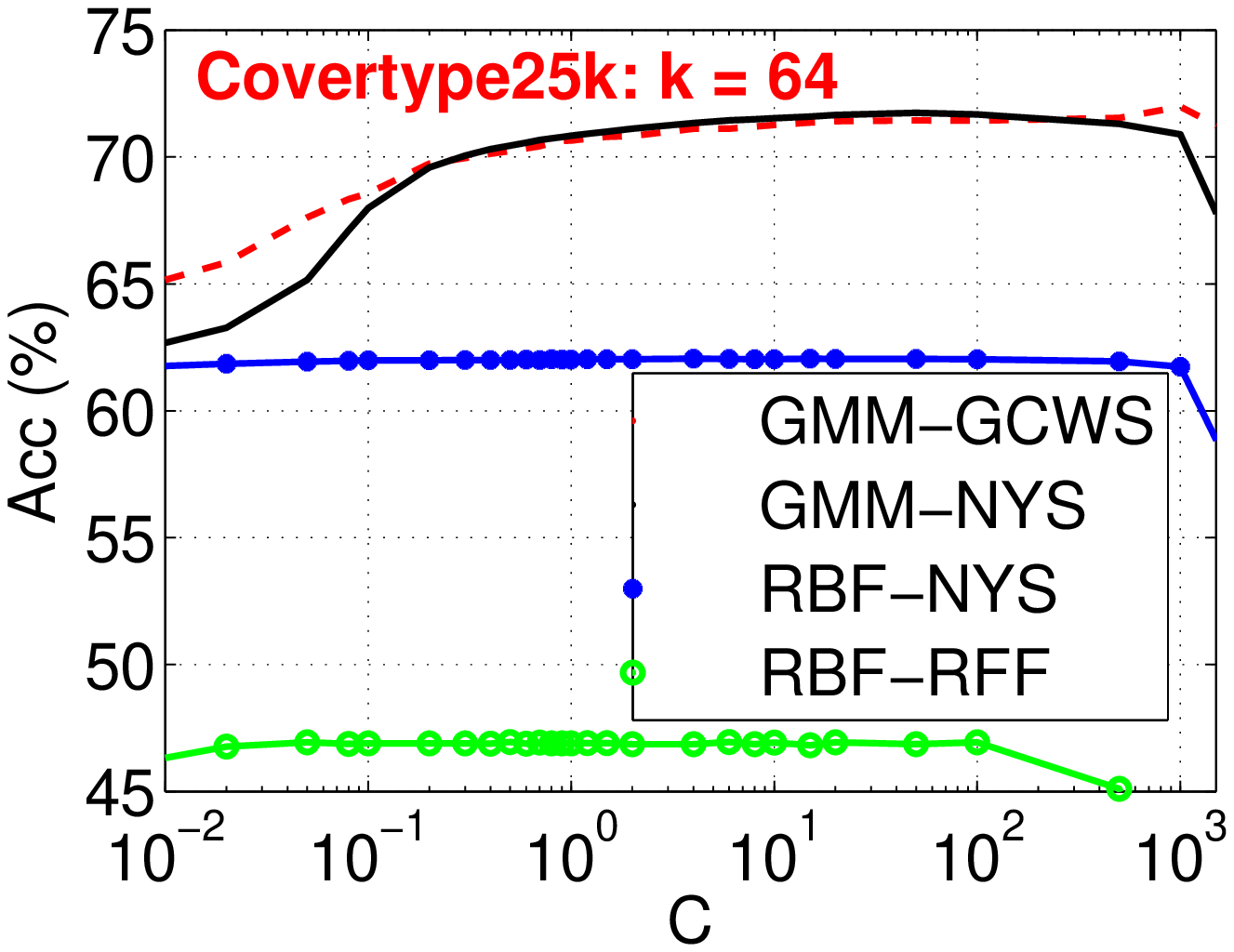}\hspace{-0.12in}
\includegraphics[width=2.3in]{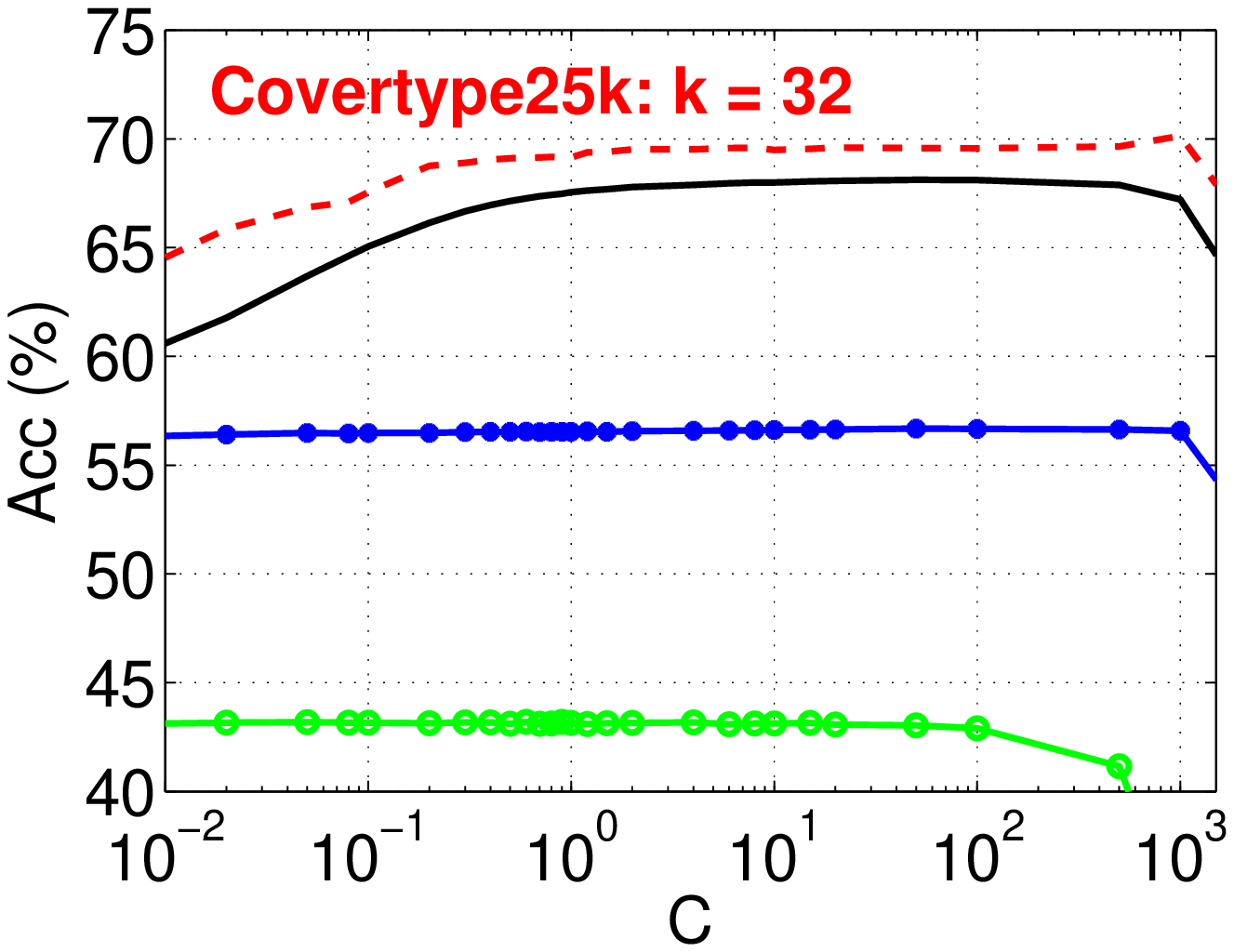}
}
\end{center}
\vspace{-0.3in}
\caption{\textbf{Covertype25k:}\  Test classification accuracies for 6  $k$ values and 4 different algorithms.  }\label{fig_Covertype25k}
\end{figure}

Figure~\ref{fig_Covertype25k} reports the test classification accuracies for {\em Covertype25k}. Once again, the results confirm that GMM-NYS produces substantially more accurate results than RBF-RFF. GMM-NYS becomes more accurate than linear SVM on the original data after $k\geq 64$.\\

\begin{figure}
\begin{center}
\mbox{
\includegraphics[width=2.3in]{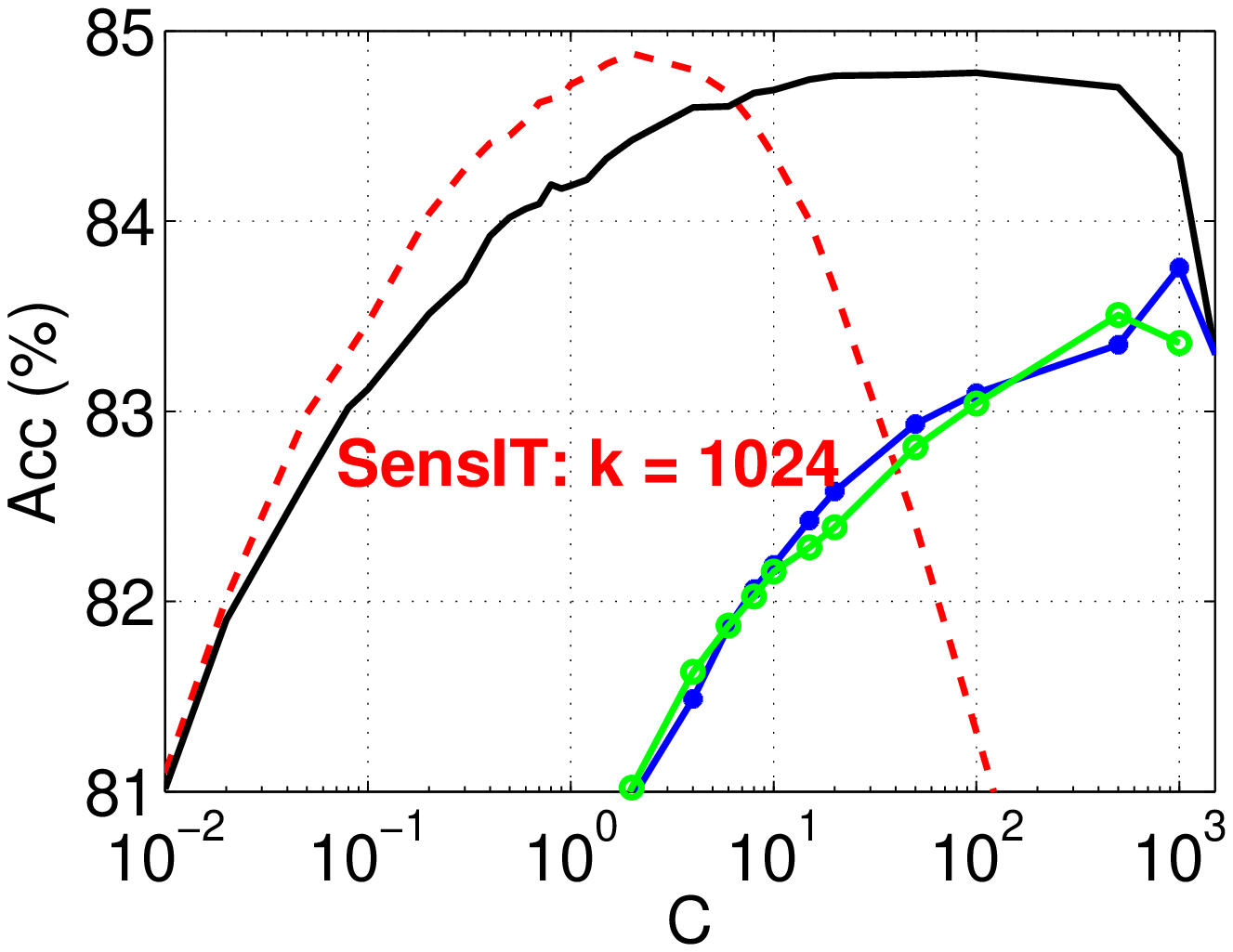}\hspace{-0.12in}
\includegraphics[width=2.3in]{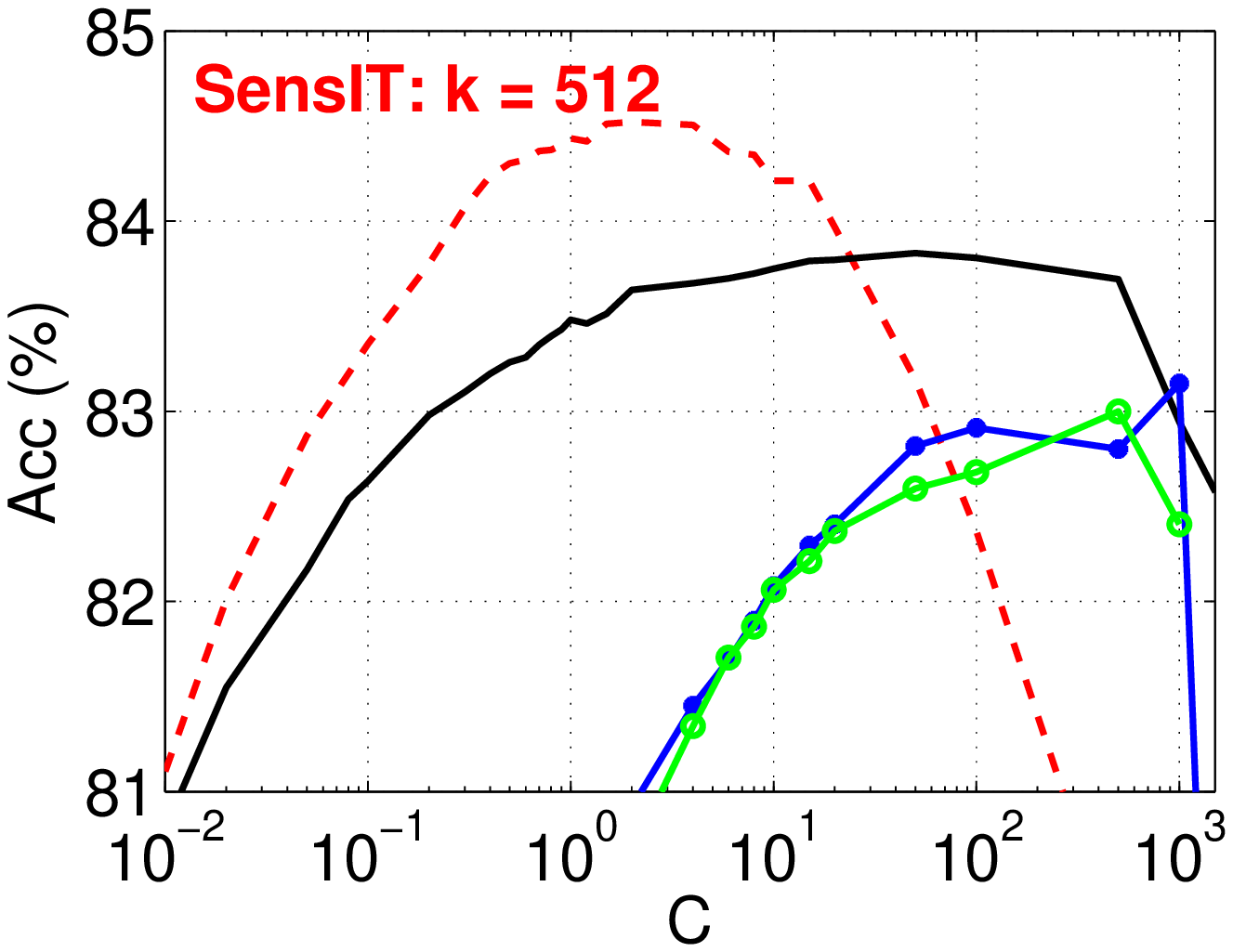}\hspace{-0.12in}
\includegraphics[width=2.3in]{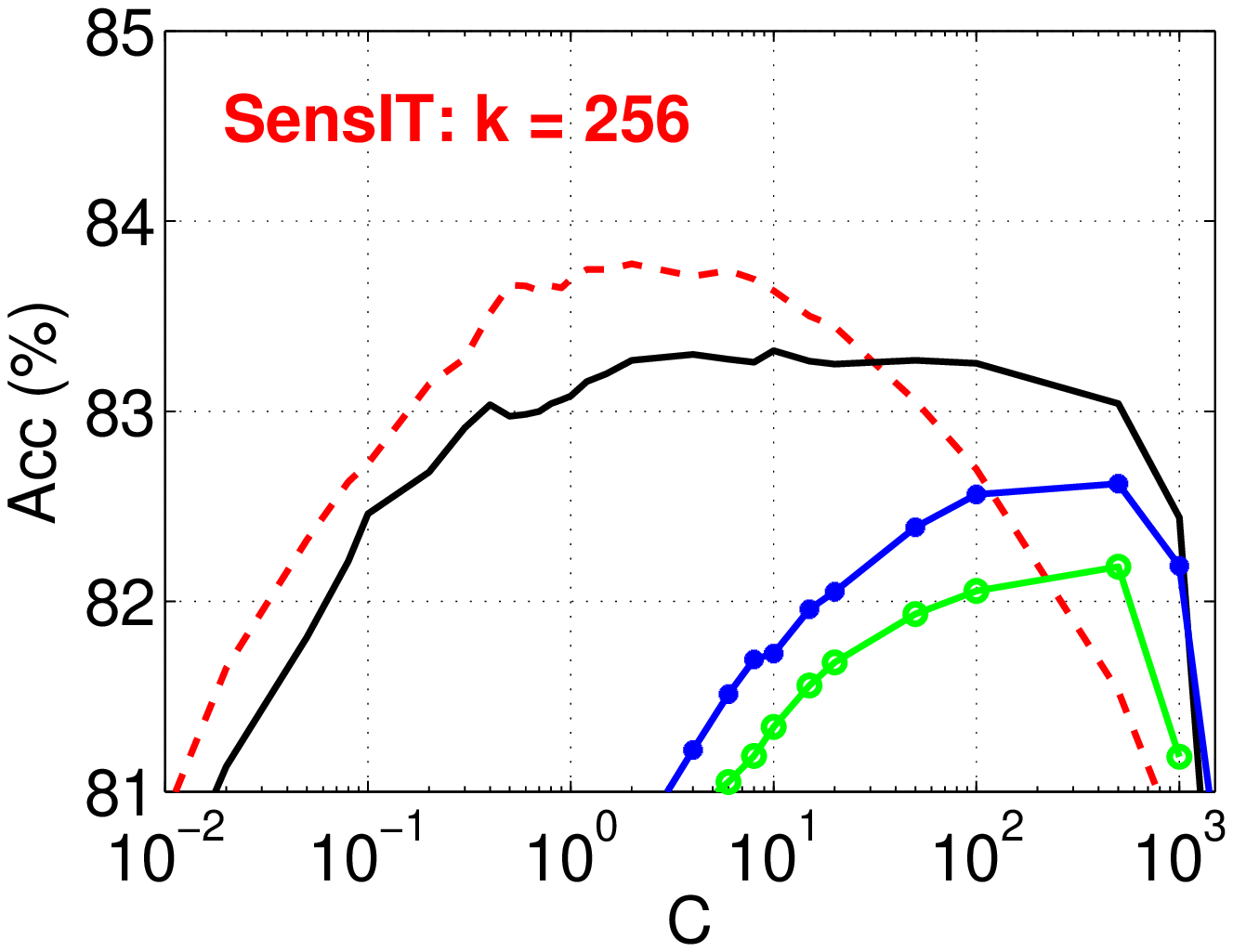}
}

\mbox{
\includegraphics[width=2.3in]{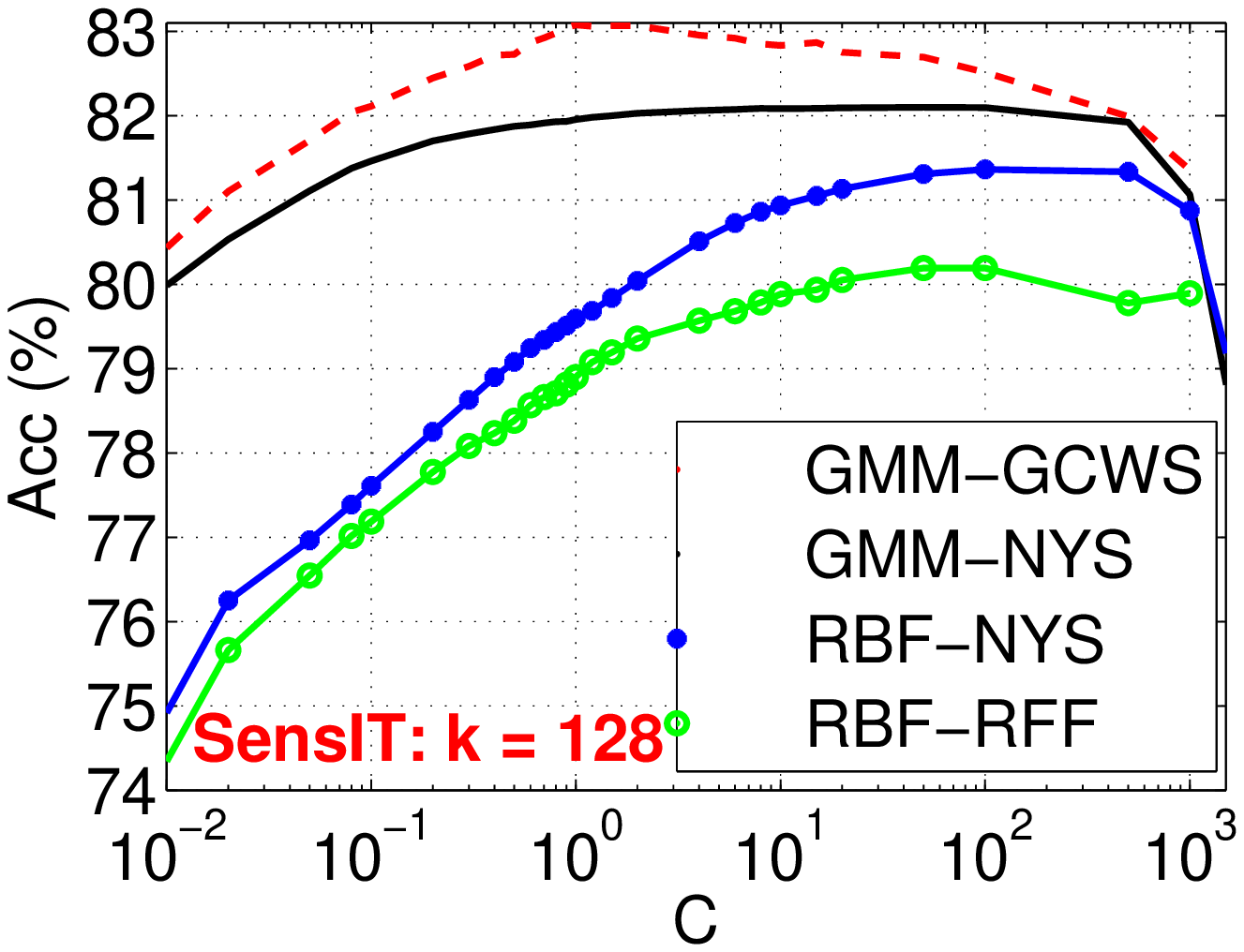}\hspace{-0.12in}
\includegraphics[width=2.3in]{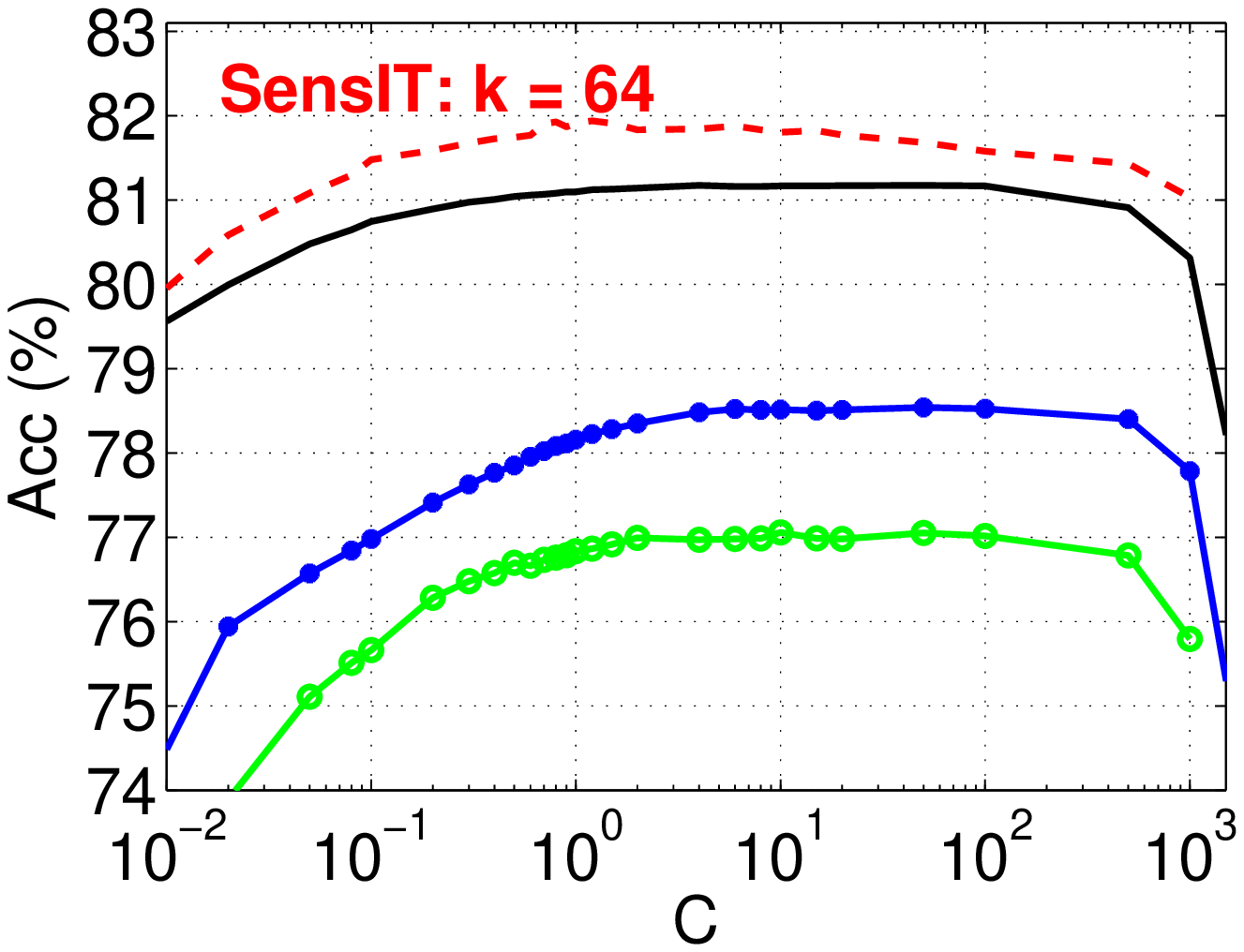}\hspace{-0.12in}
\includegraphics[width=2.3in]{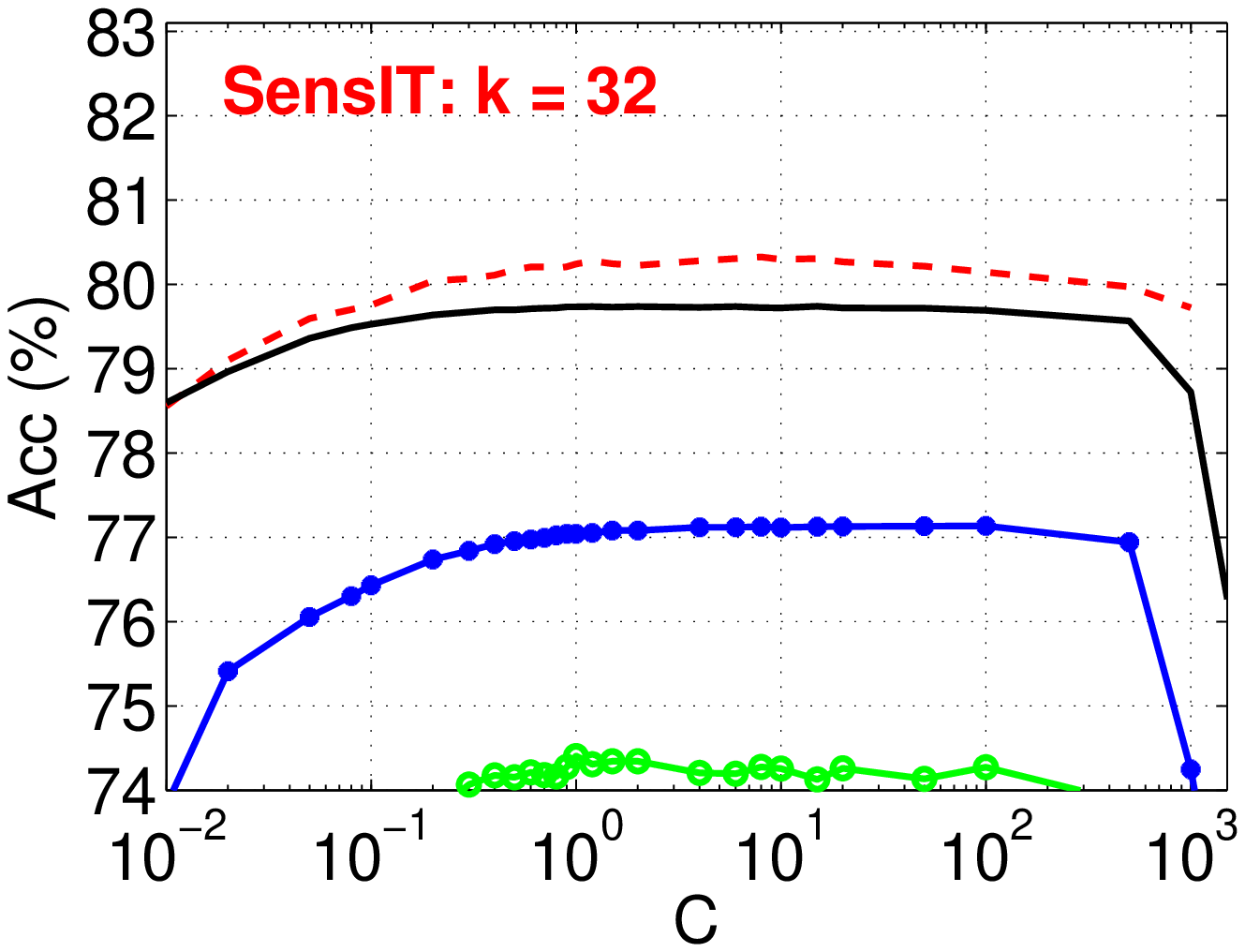}
}
\end{center}
\vspace{-0.2in}
\caption{\textbf{SensIT:} \ Test classification accuracies for 6  $k$ values and 4 different  algorithms.}\label{fig_SensIT}
\end{figure}

\begin{figure}
\begin{center}
\mbox{
\includegraphics[width=2.3in]{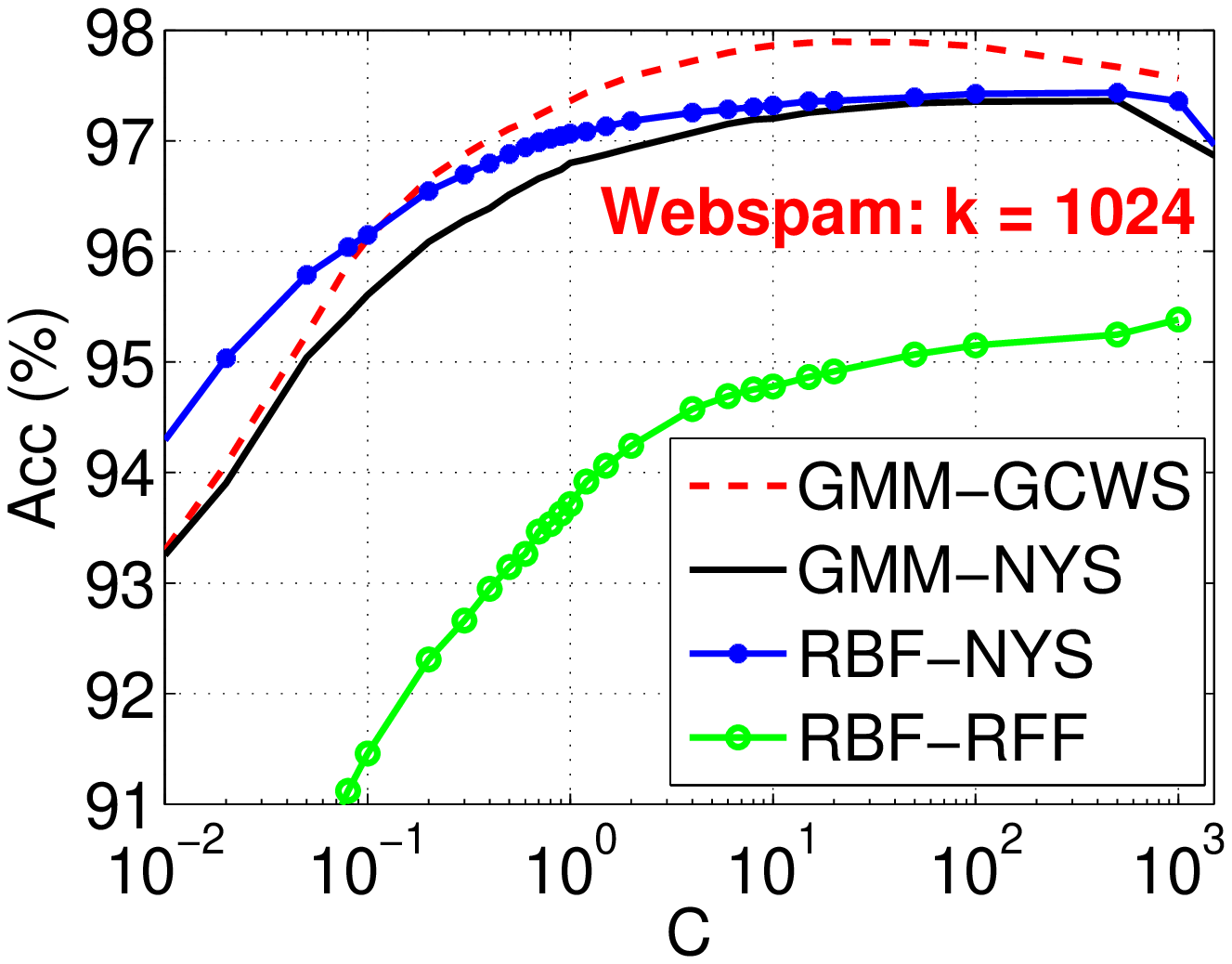}\hspace{-0.12in}
\includegraphics[width=2.3in]{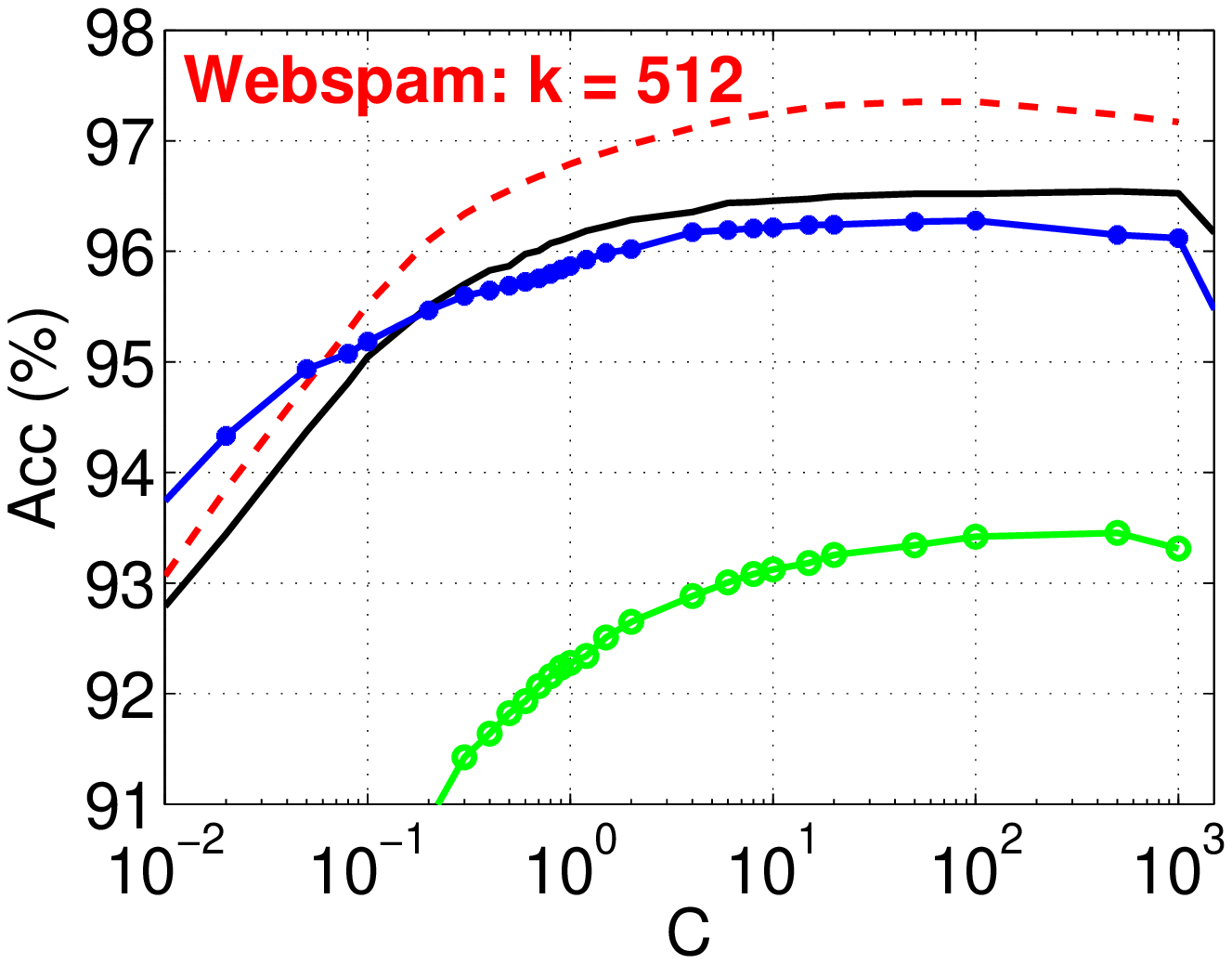}\hspace{-0.12in}
\includegraphics[width=2.3in]{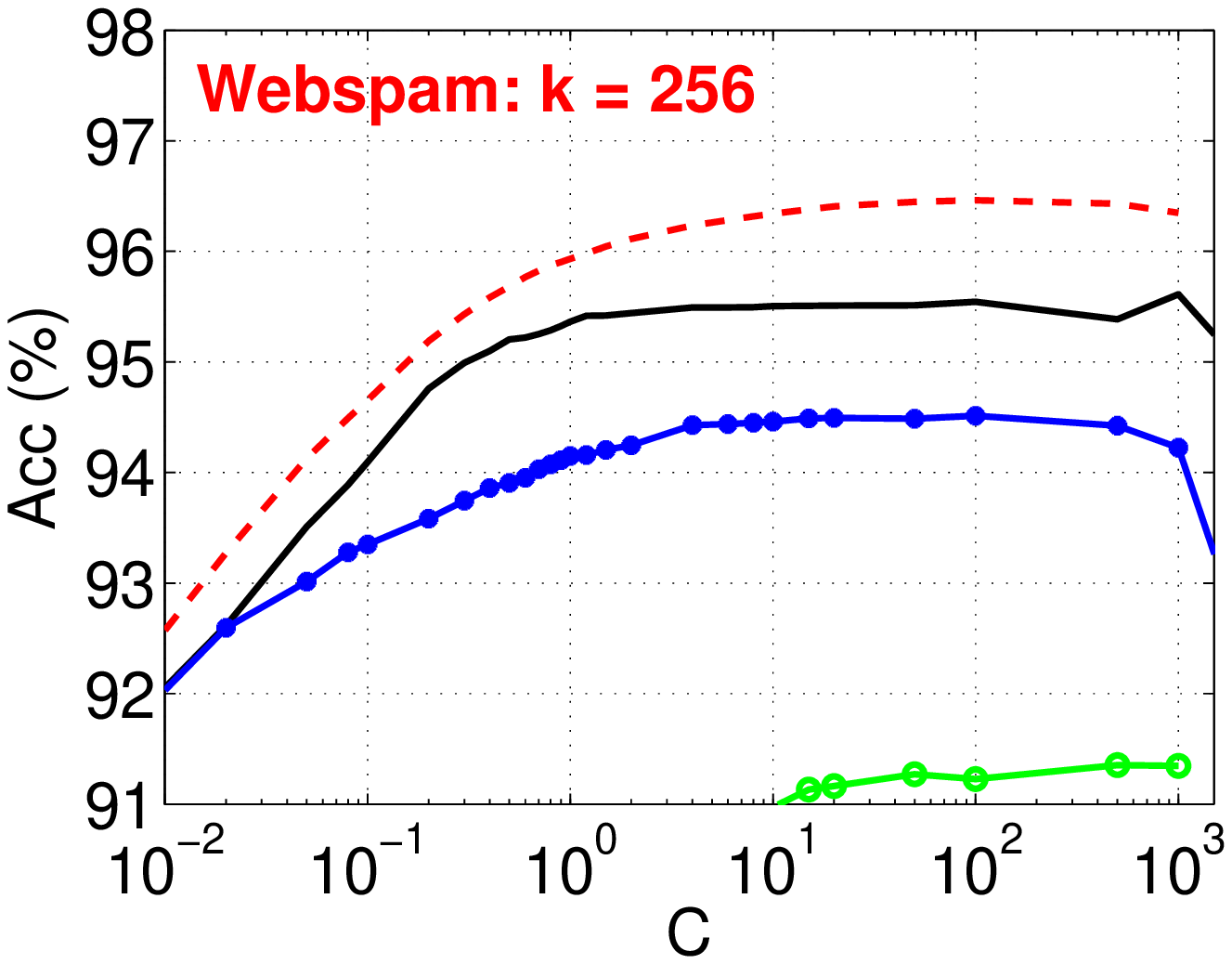}
}

\mbox{
\includegraphics[width=2.3in]{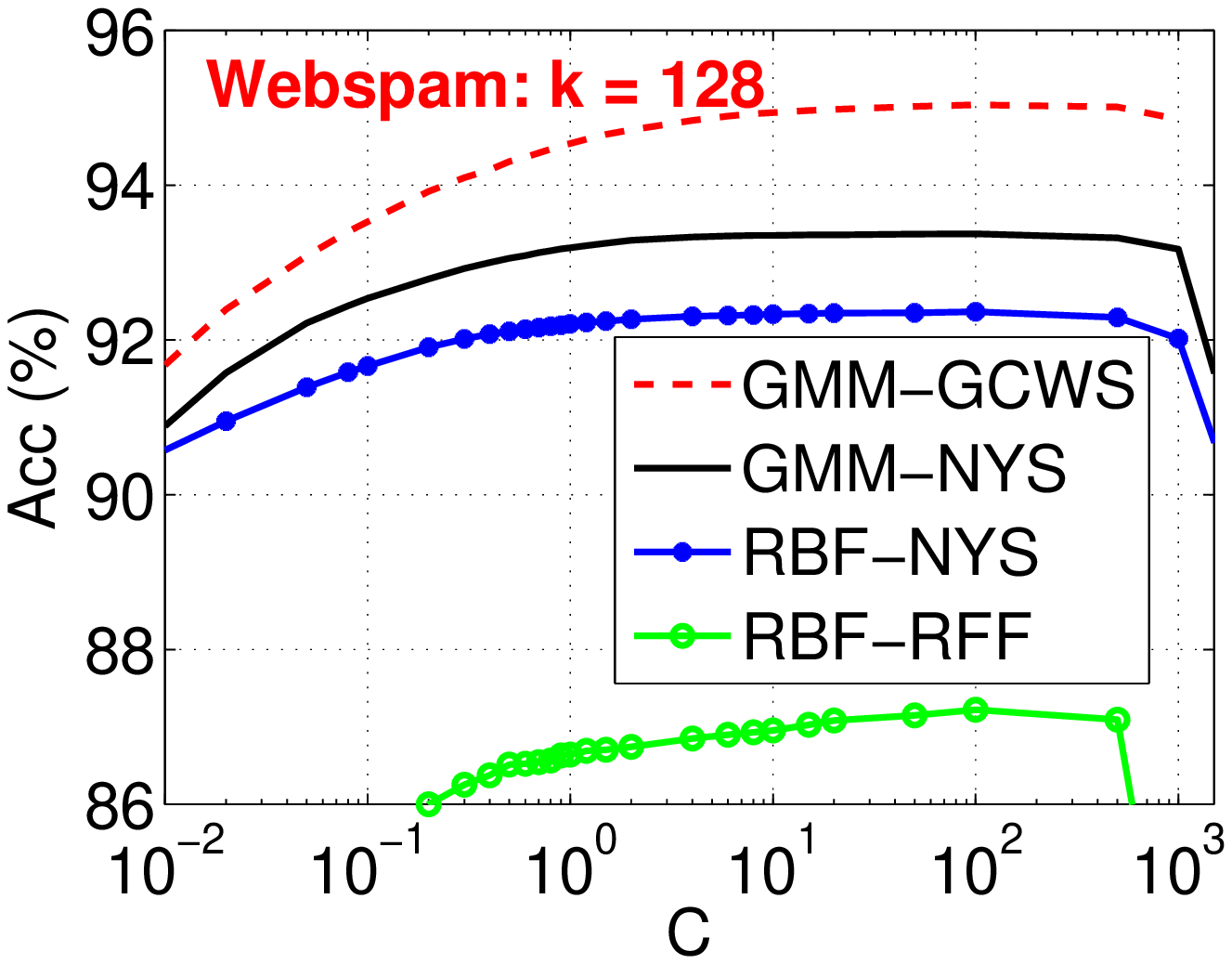}\hspace{-0.12in}
\includegraphics[width=2.3in]{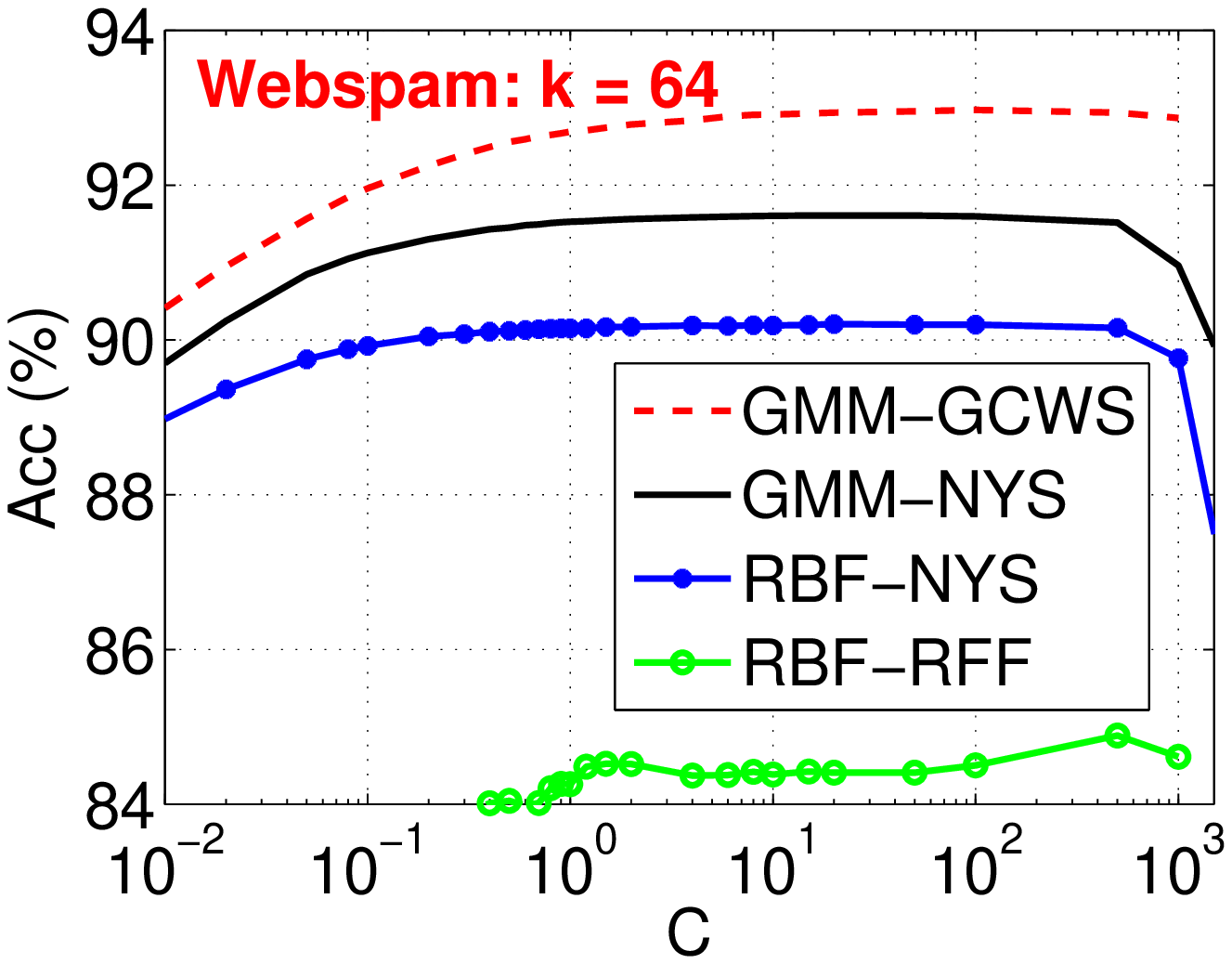}\hspace{-0.12in}
\includegraphics[width=2.3in]{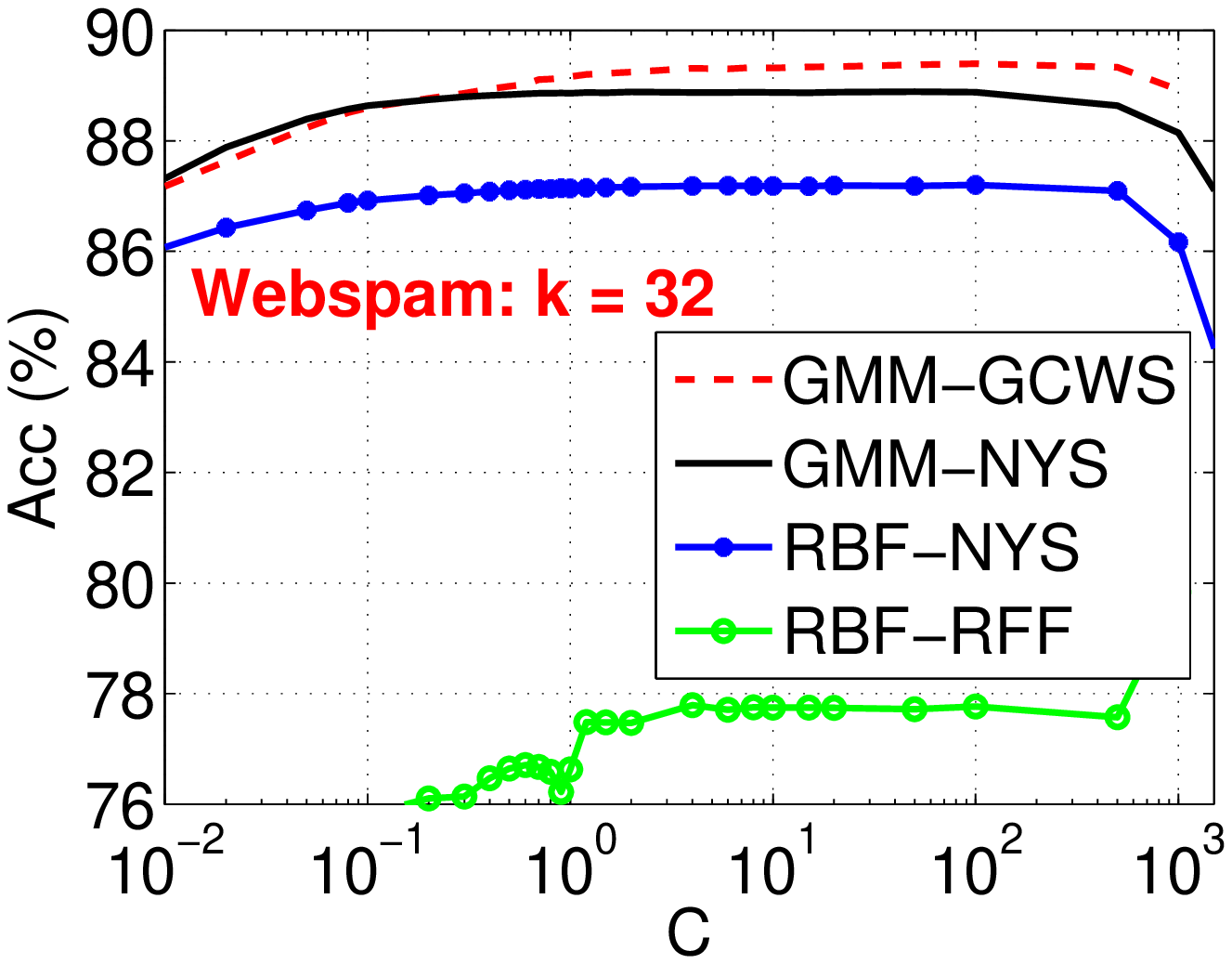}
}
\end{center}
\vspace{-0.3in}
\caption{\textbf{Webspam:} Test classification accuracies for 6  $k$ values and 4 different  algorithms}\label{fig_Webspam}
\end{figure}

\begin{figure}
\begin{center}
\mbox{
\includegraphics[width=2.3in]{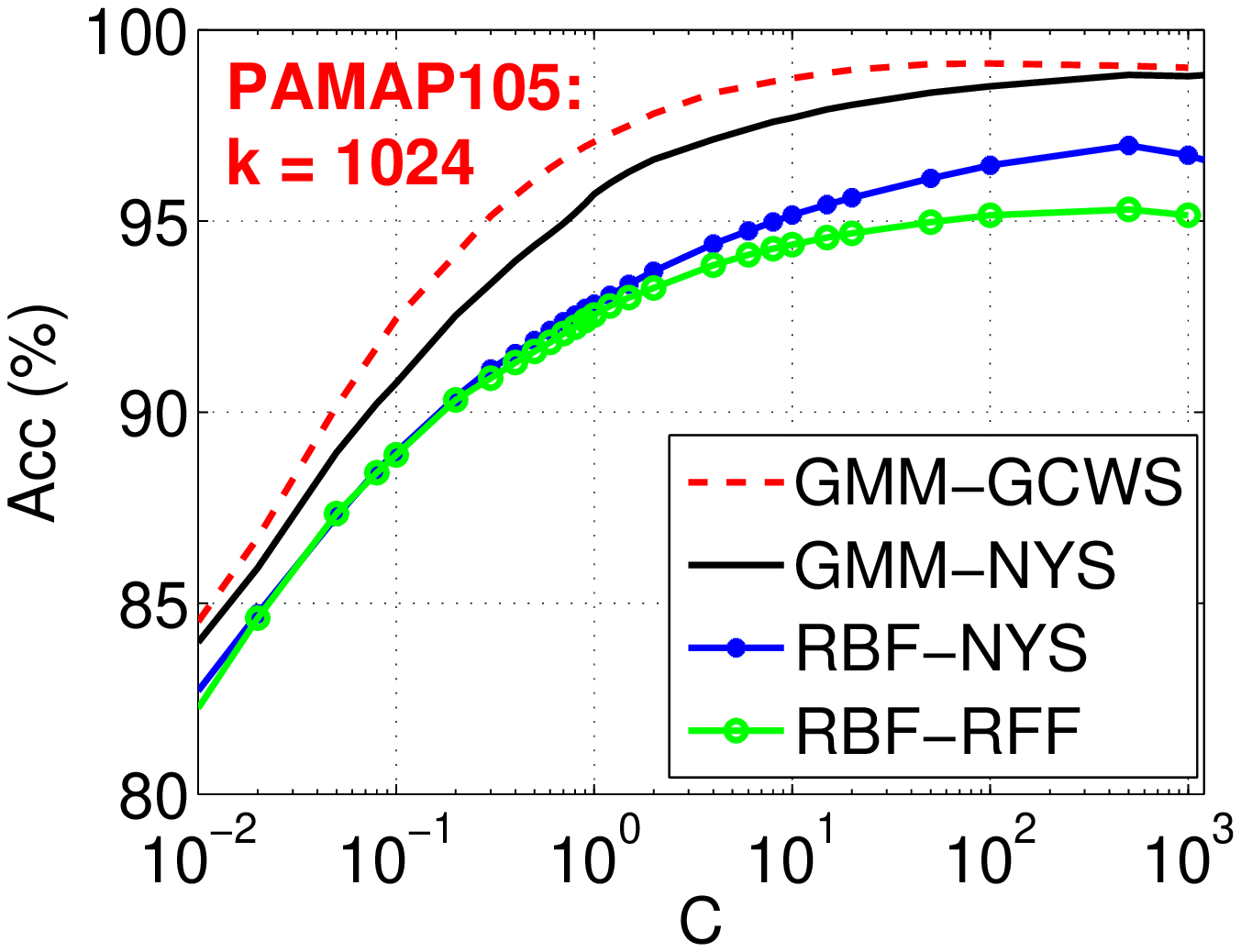}\hspace{-0.12in}
\includegraphics[width=2.3in]{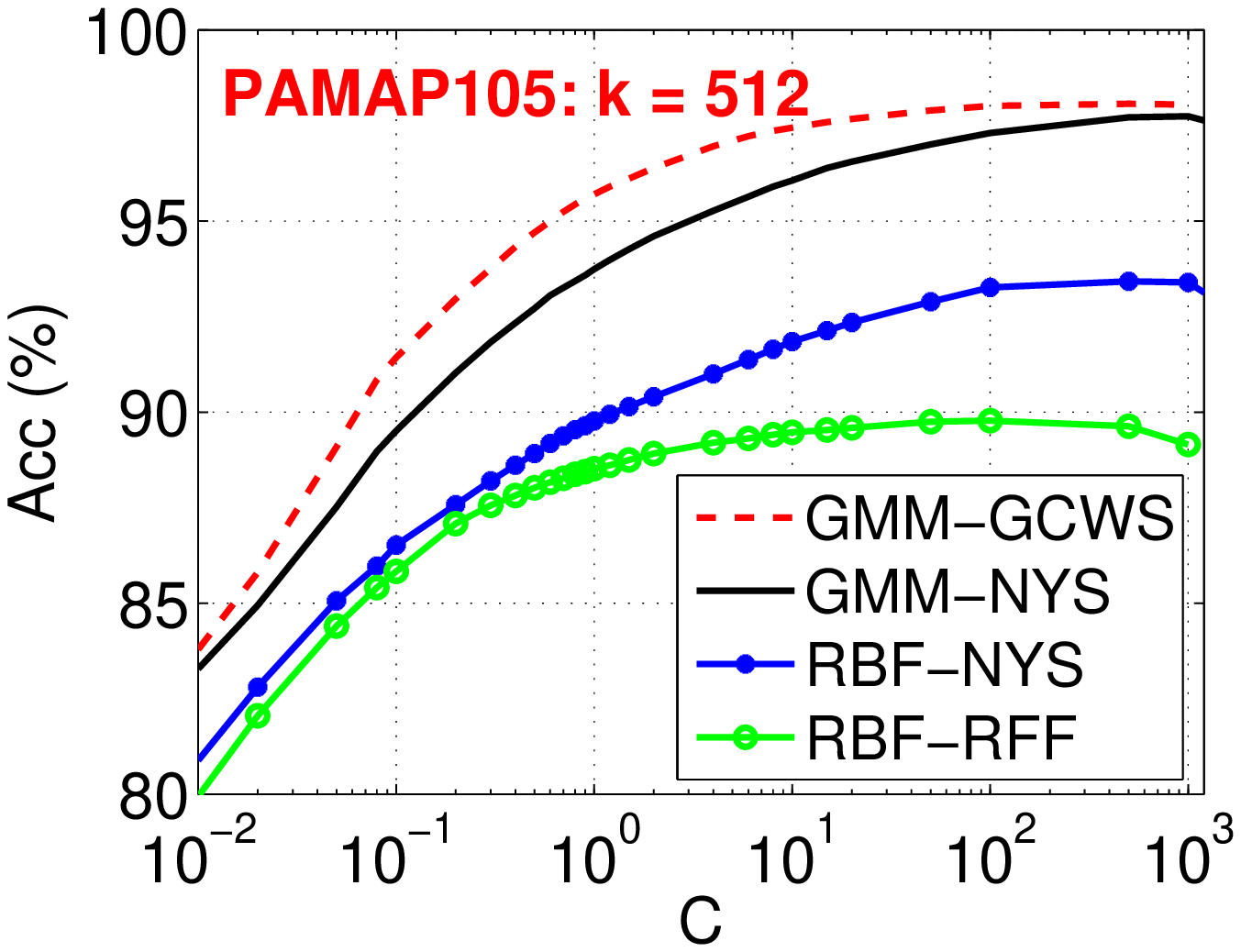}\hspace{-0.12in}
\includegraphics[width=2.3in]{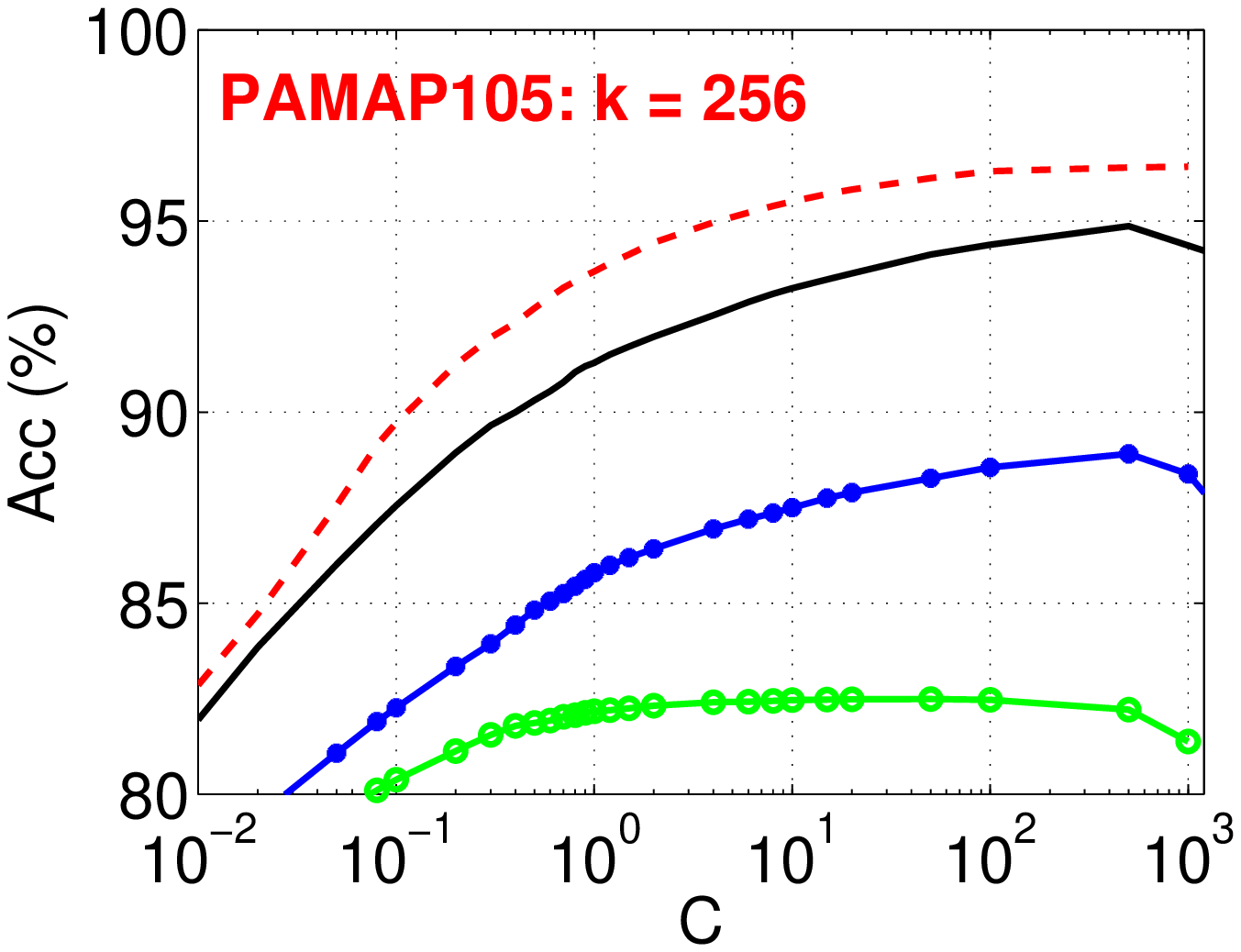}
}

\mbox{
\includegraphics[width=2.3in]{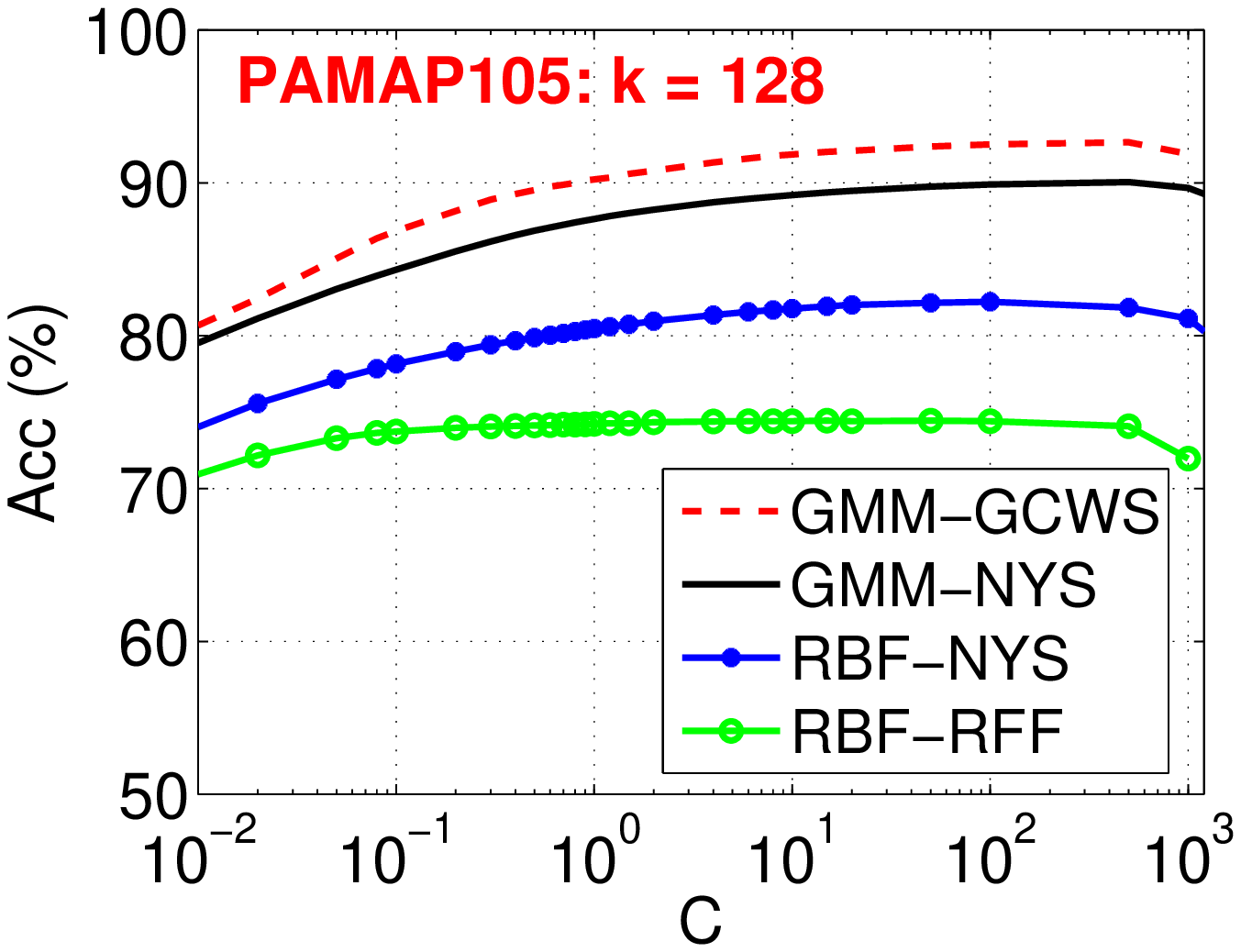}\hspace{-0.12in}
\includegraphics[width=2.3in]{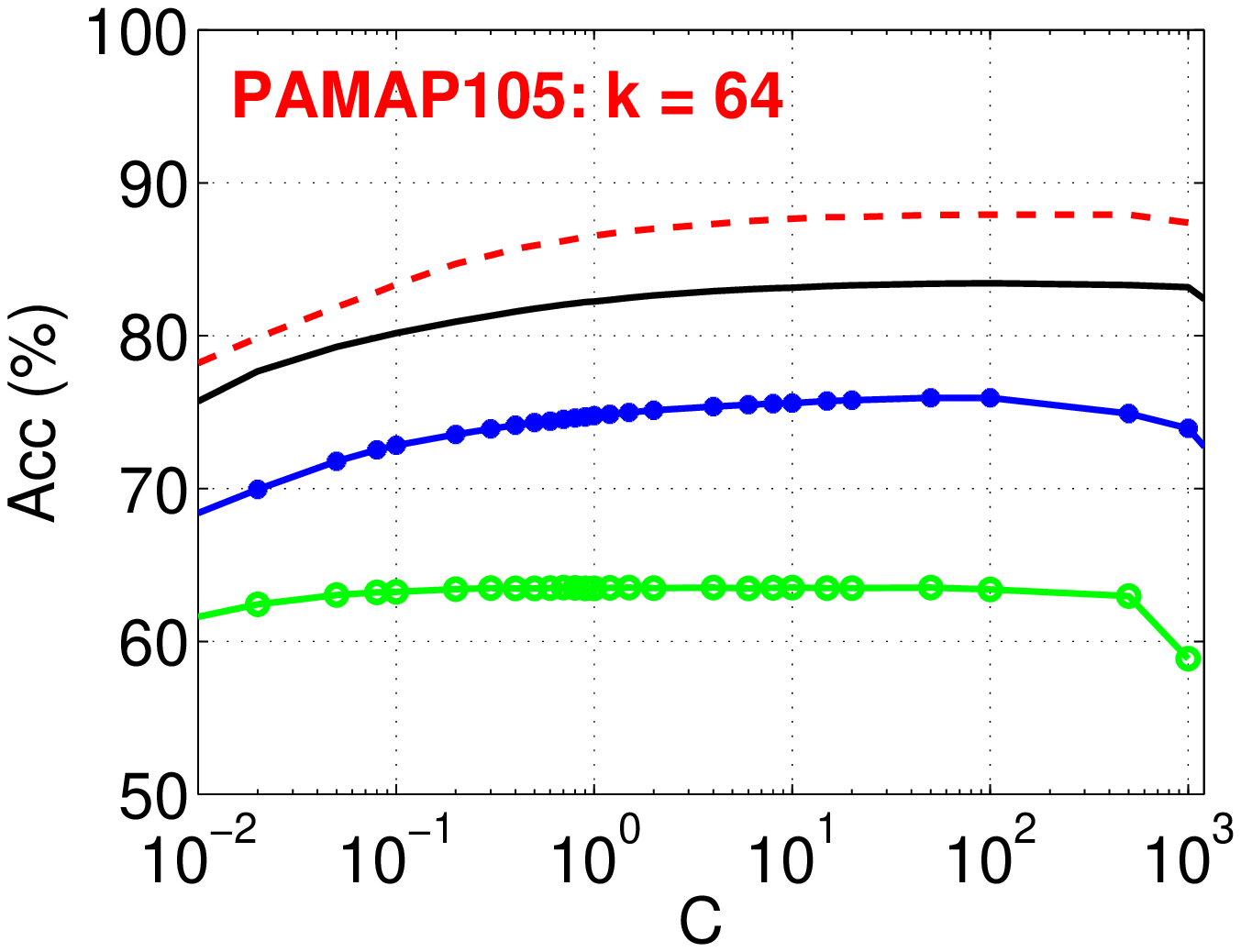}\hspace{-0.12in}
\includegraphics[width=2.3in]{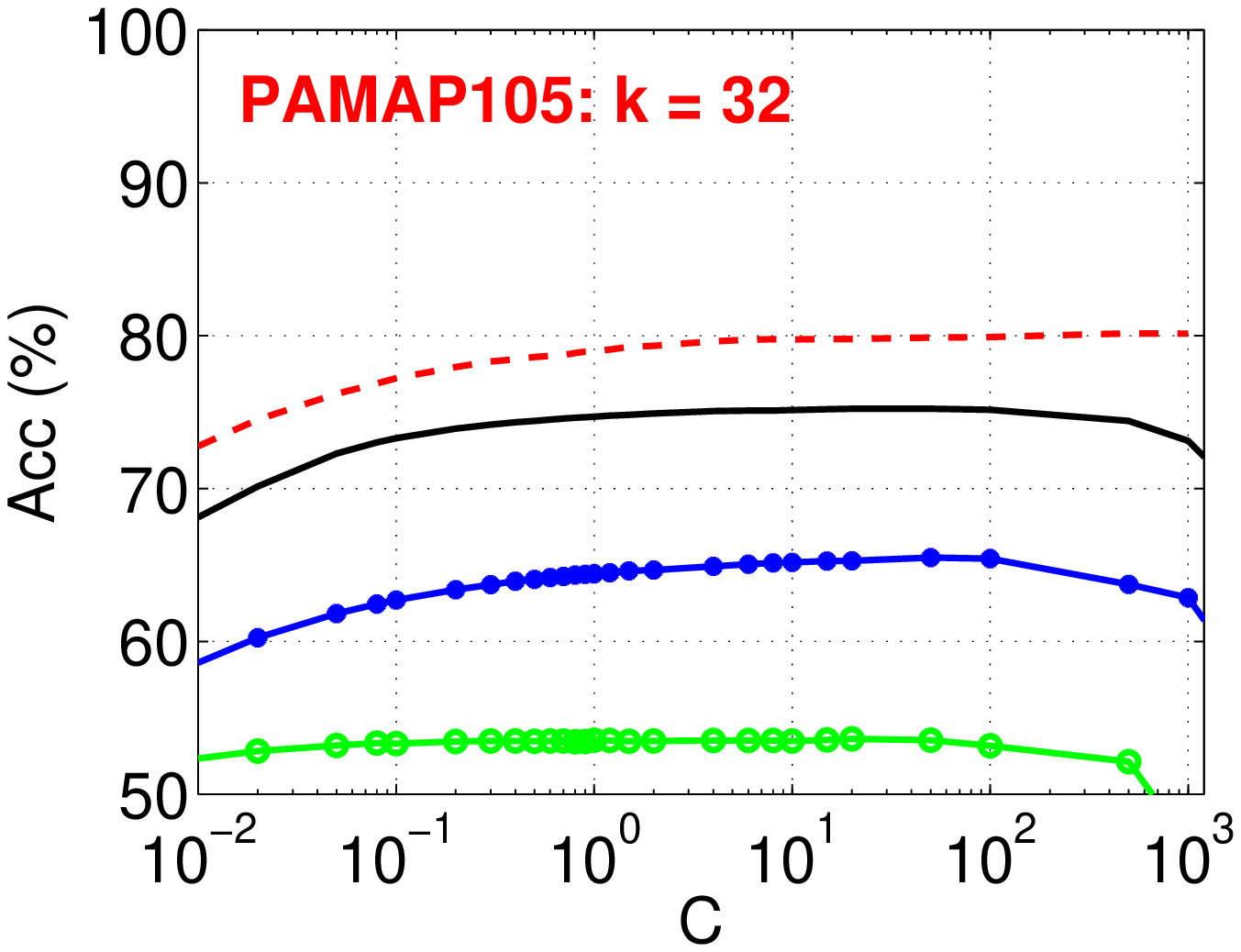}
}
\end{center}
\vspace{-0.3in}
\caption{\textbf{PAMAP105:}\ Test classification accuracies for 6  $k$ values and 4 different  algorithms.}\label{fig_PAMAP105}
\end{figure}

\begin{figure}
\begin{center}
\mbox{
\includegraphics[width=2.3in]{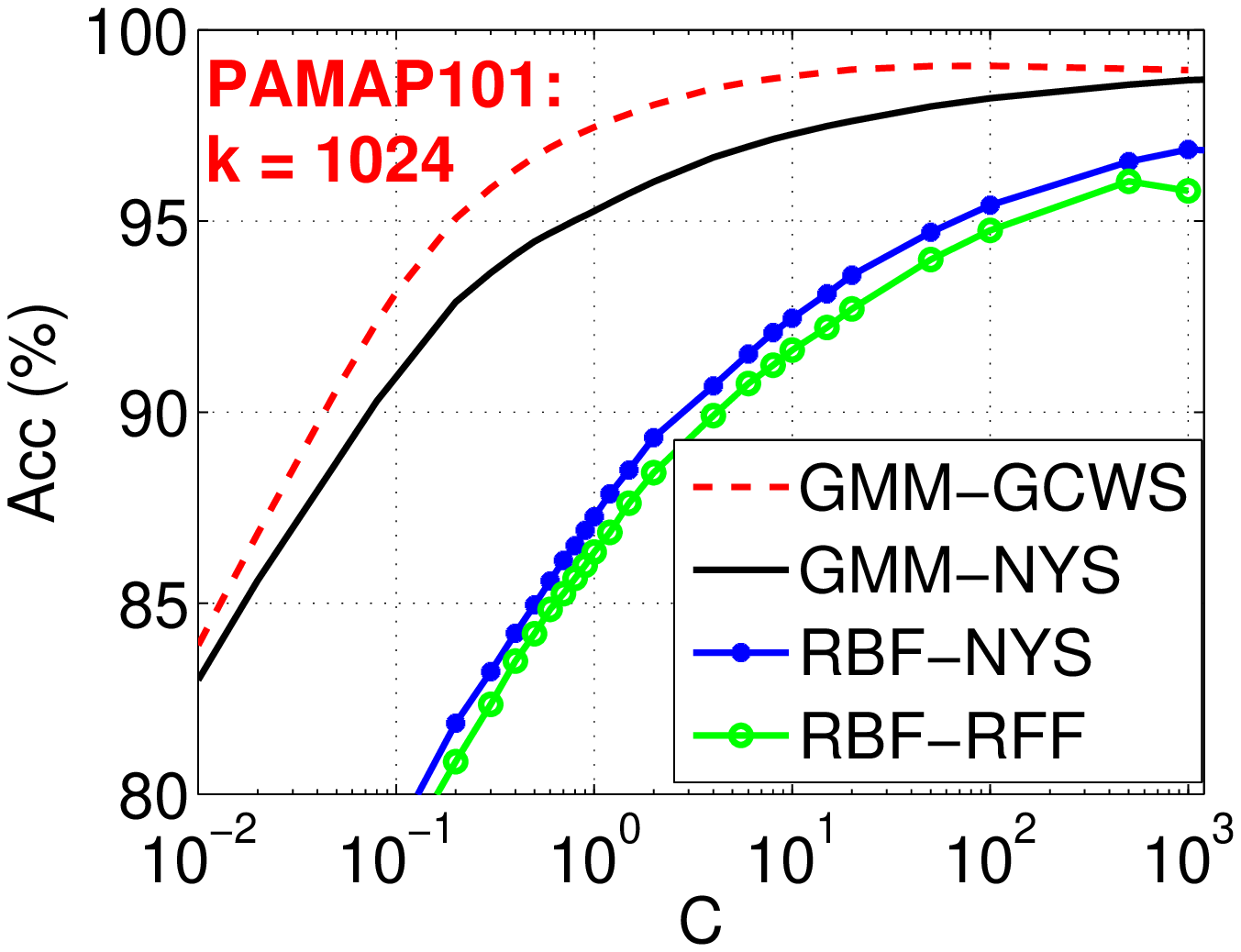}\hspace{-0.12in}
\includegraphics[width=2.3in]{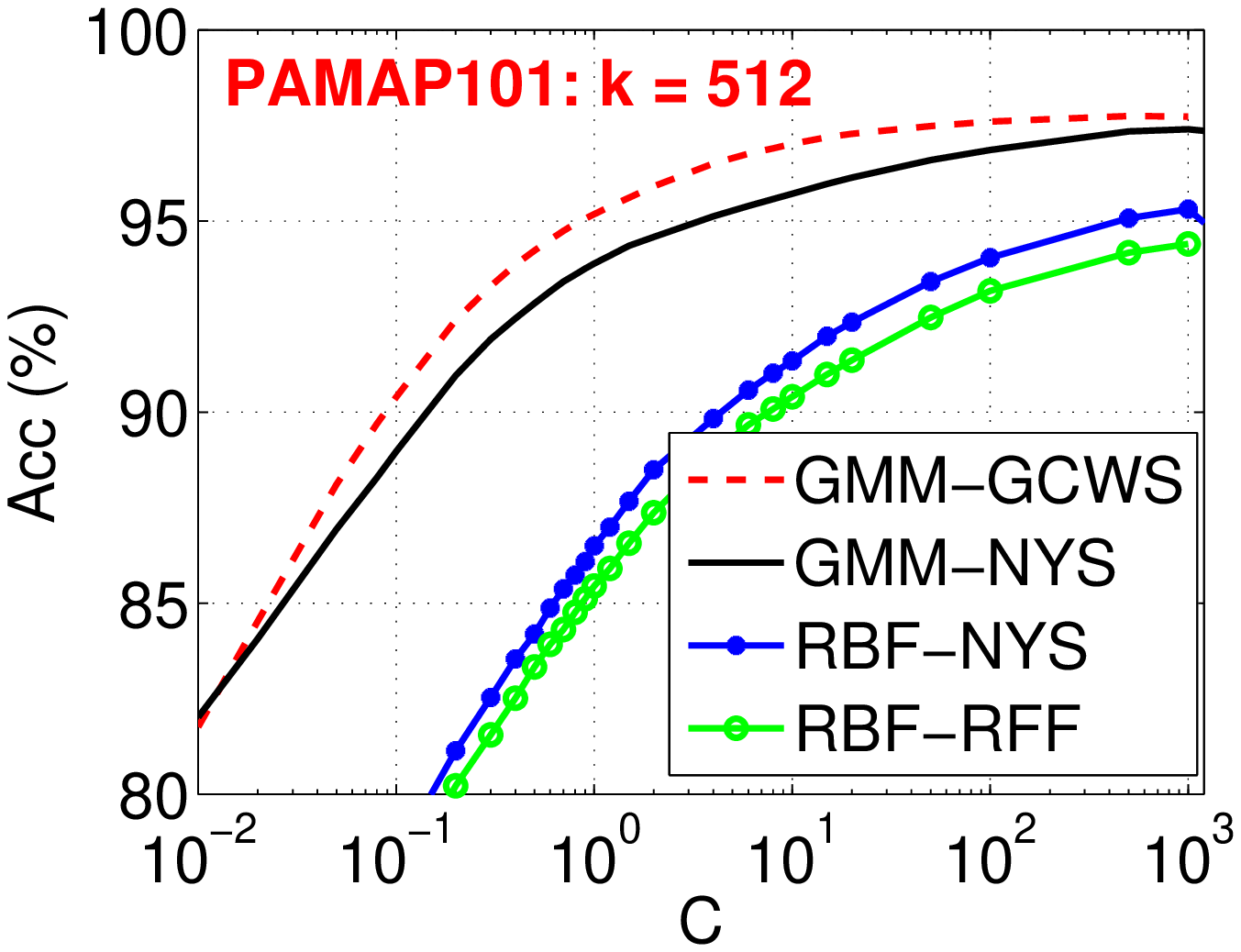}\hspace{-0.12in}
\includegraphics[width=2.3in]{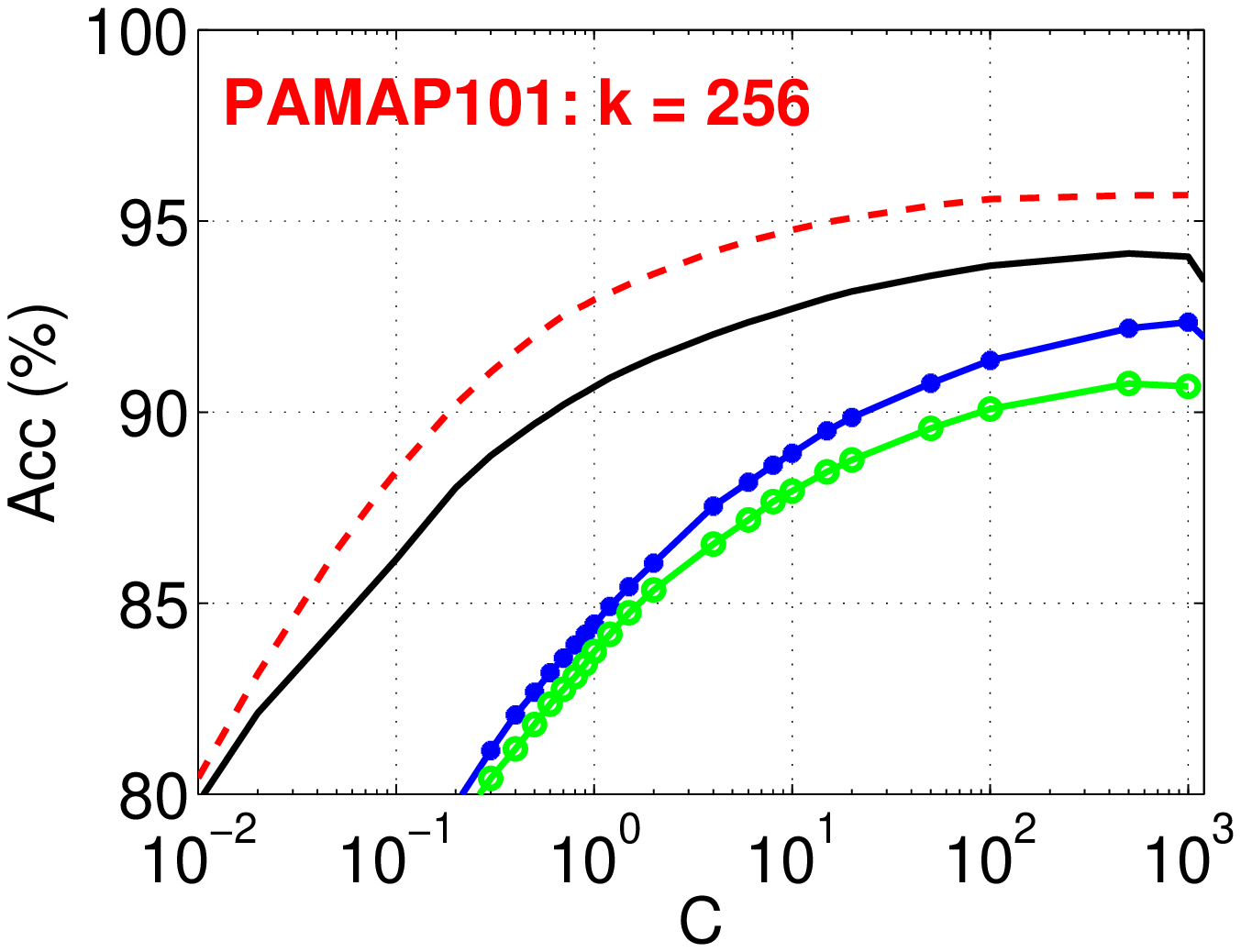}
}

\mbox{
\includegraphics[width=2.3in]{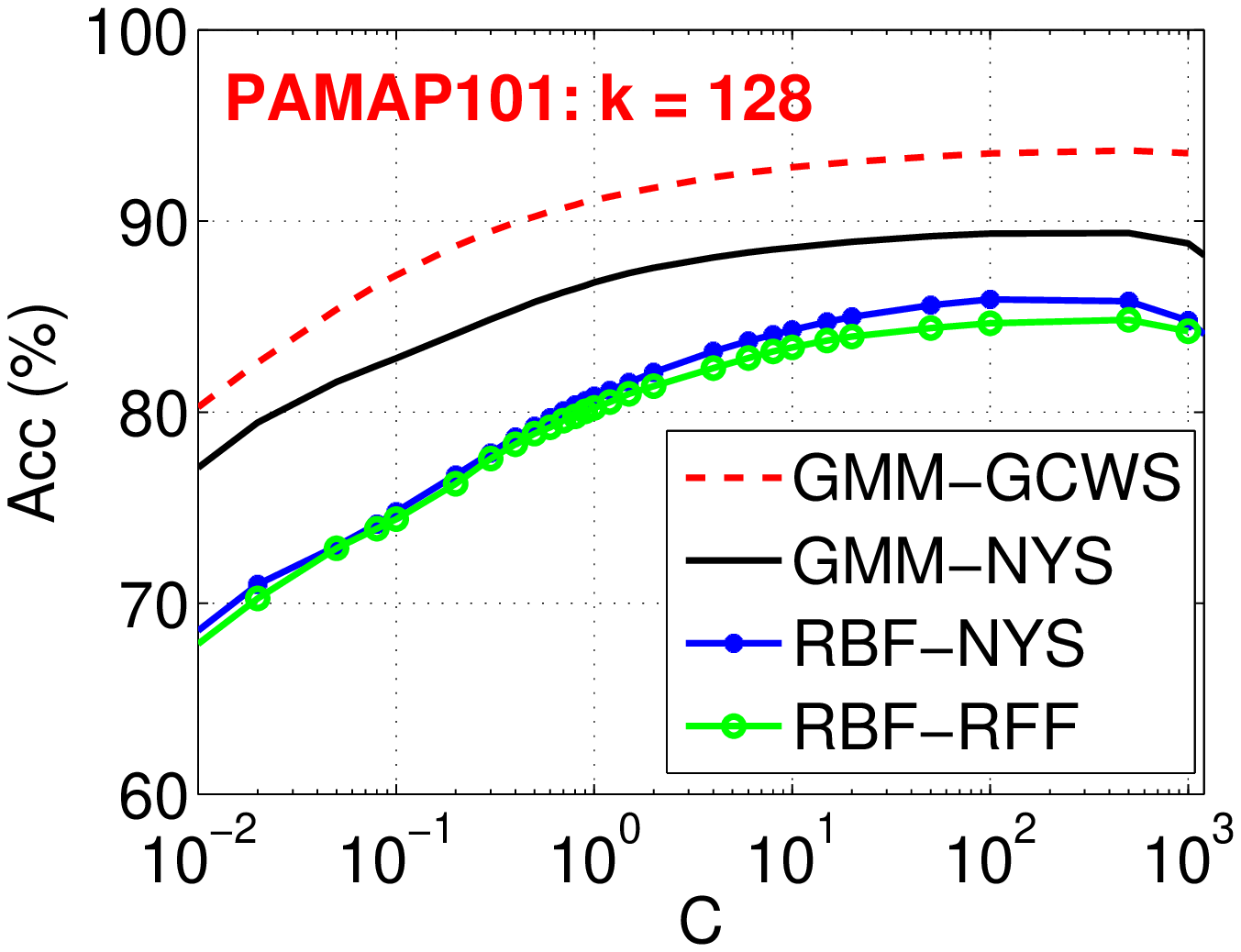}\hspace{-0.12in}
\includegraphics[width=2.3in]{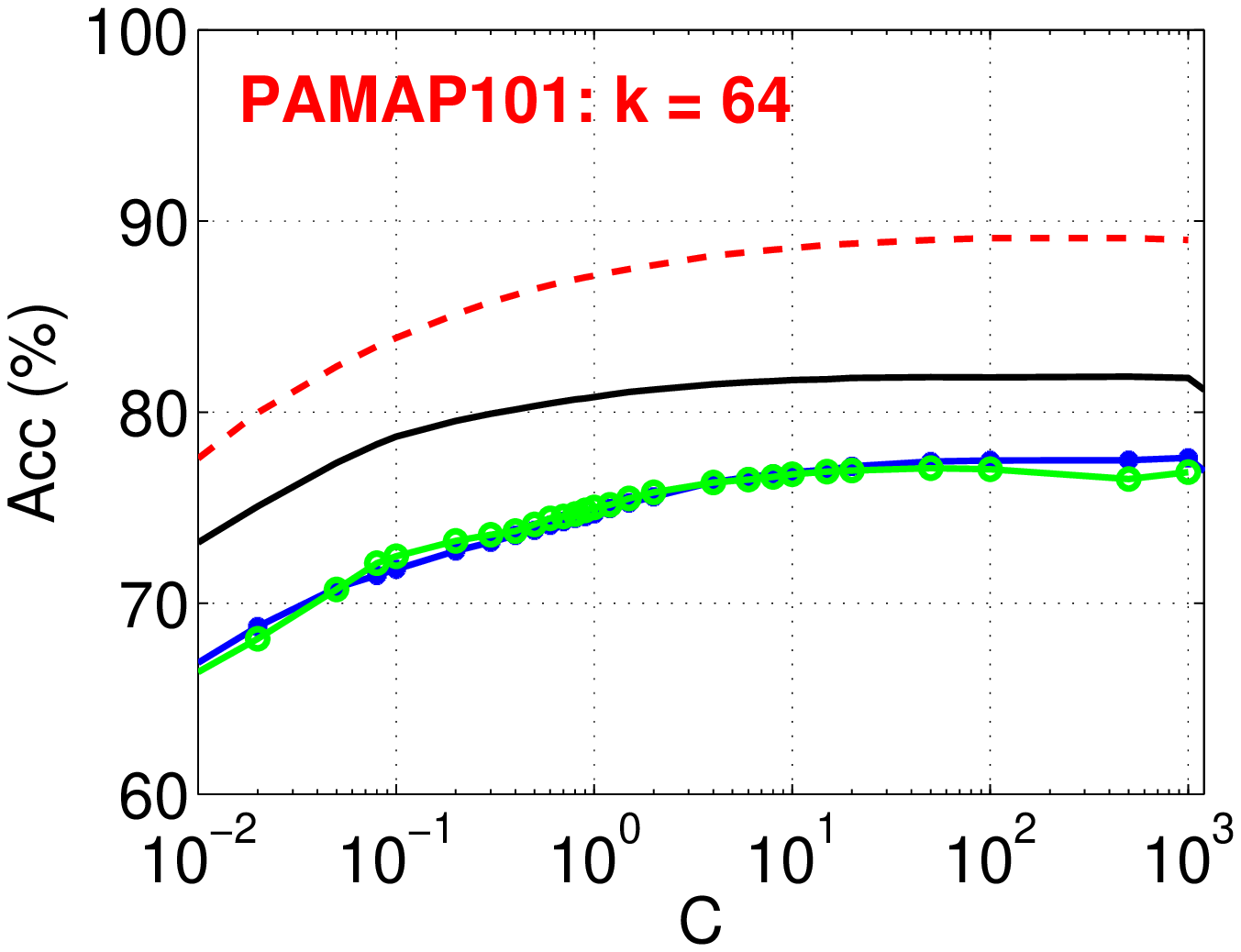}\hspace{-0.12in}
\includegraphics[width=2.3in]{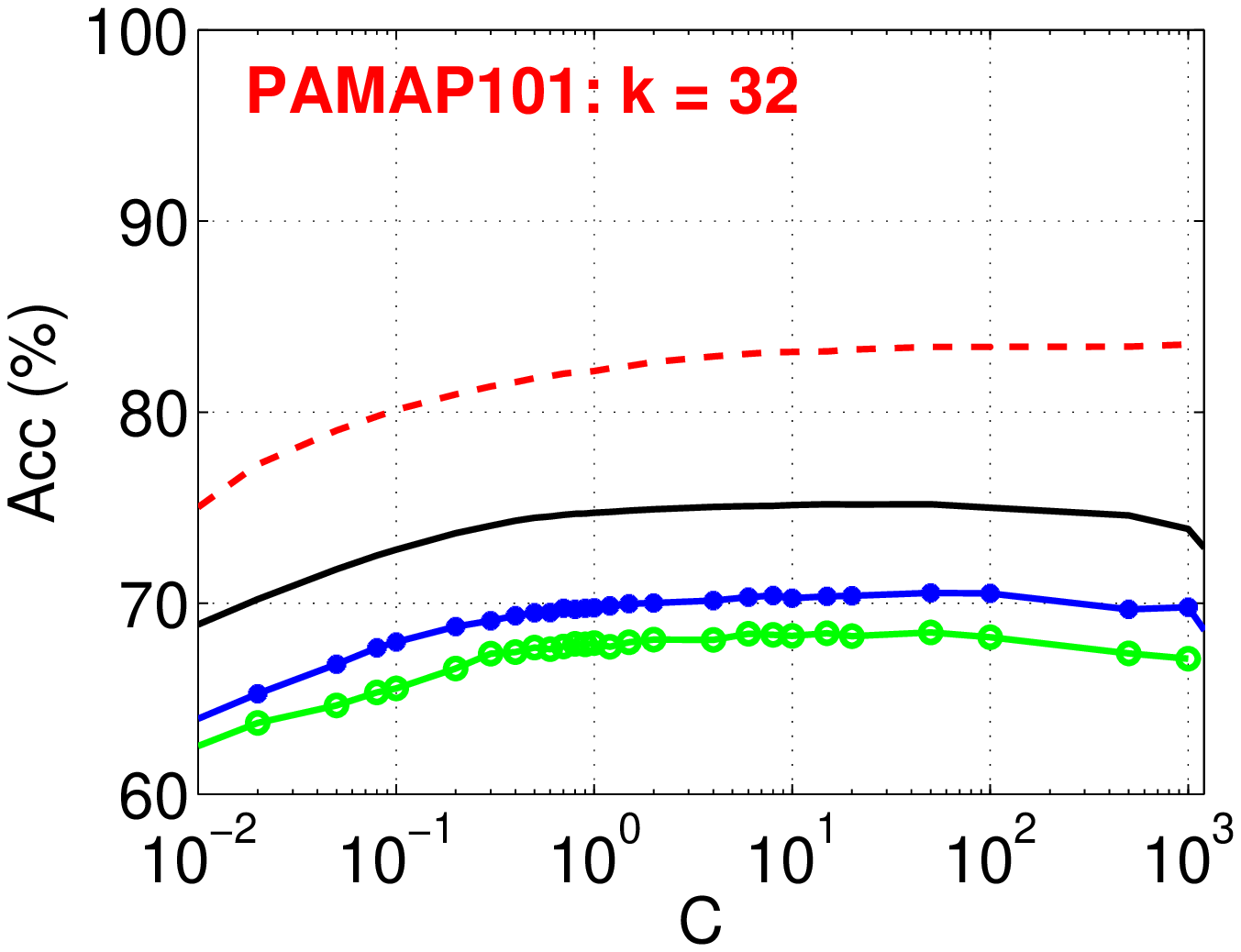}
}
\end{center}
\vspace{-0.3in}
\caption{\textbf{PAMAP101:}\ Test classification accuracies for 6  $k$ values and 4 different  algorithms.}\label{fig_PAMAP101}
\end{figure}

\begin{figure}
\begin{center}
\mbox{
\includegraphics[width=2.3in]{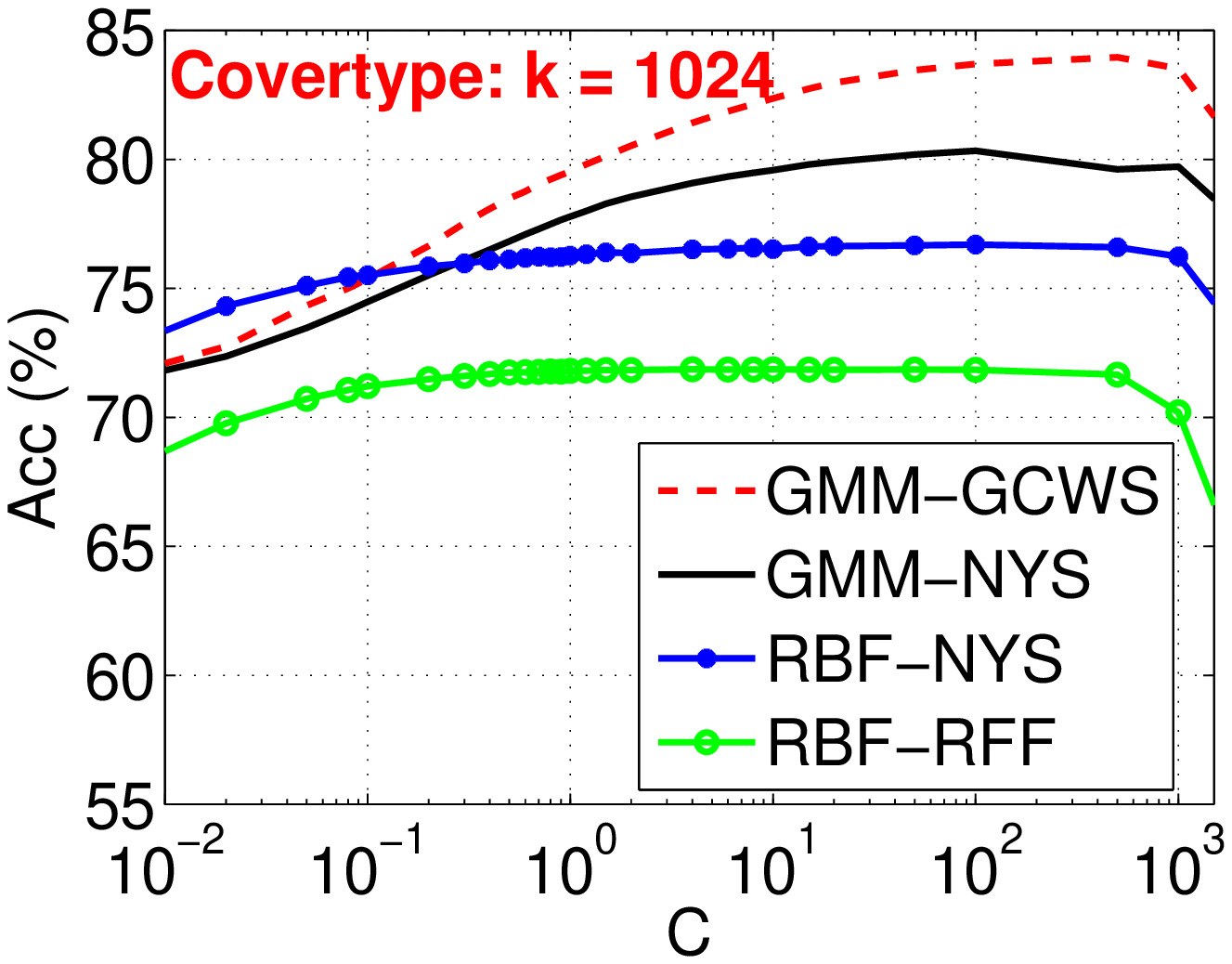}\hspace{-0.12in}
\includegraphics[width=2.3in]{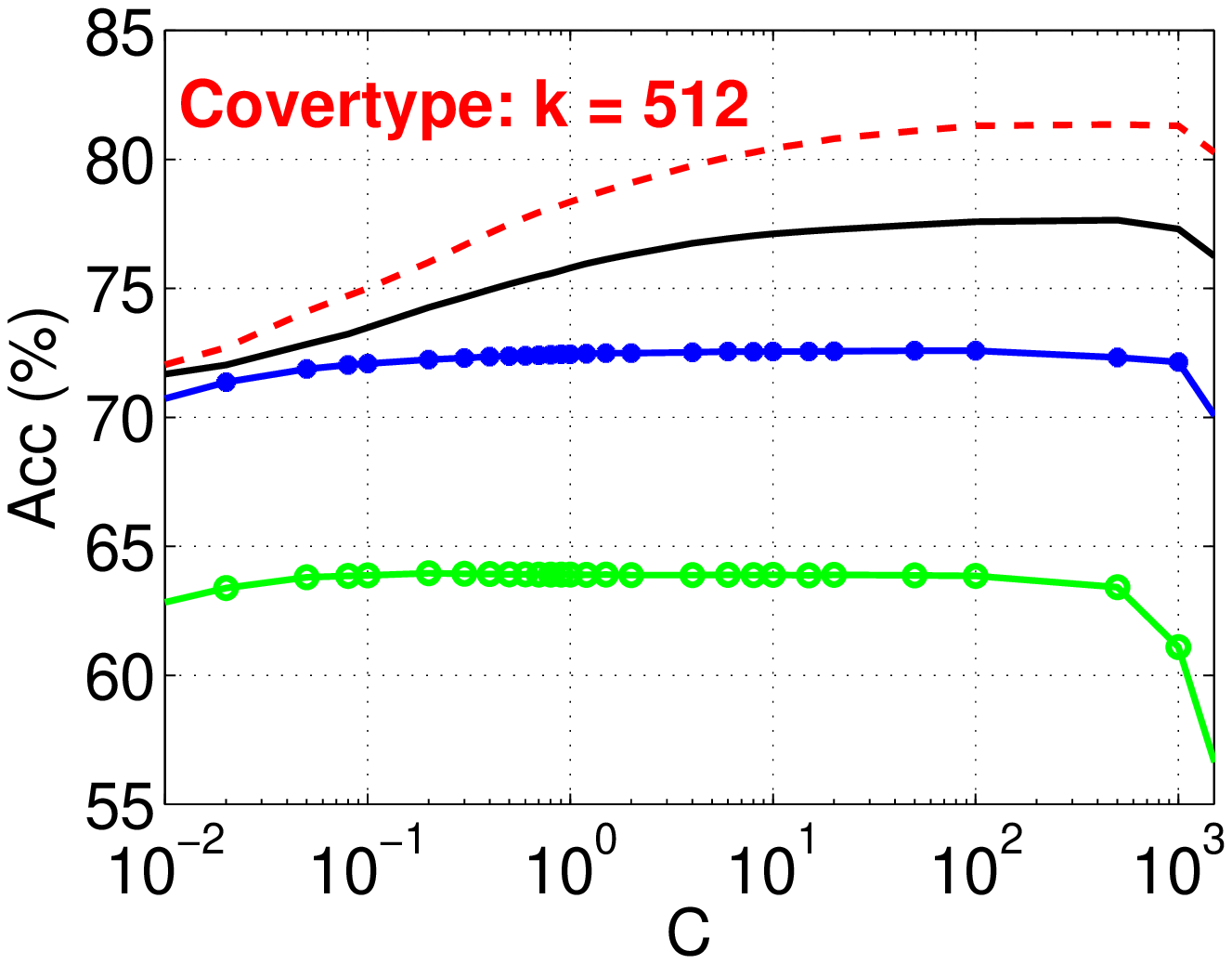}\hspace{-0.12in}
\includegraphics[width=2.3in]{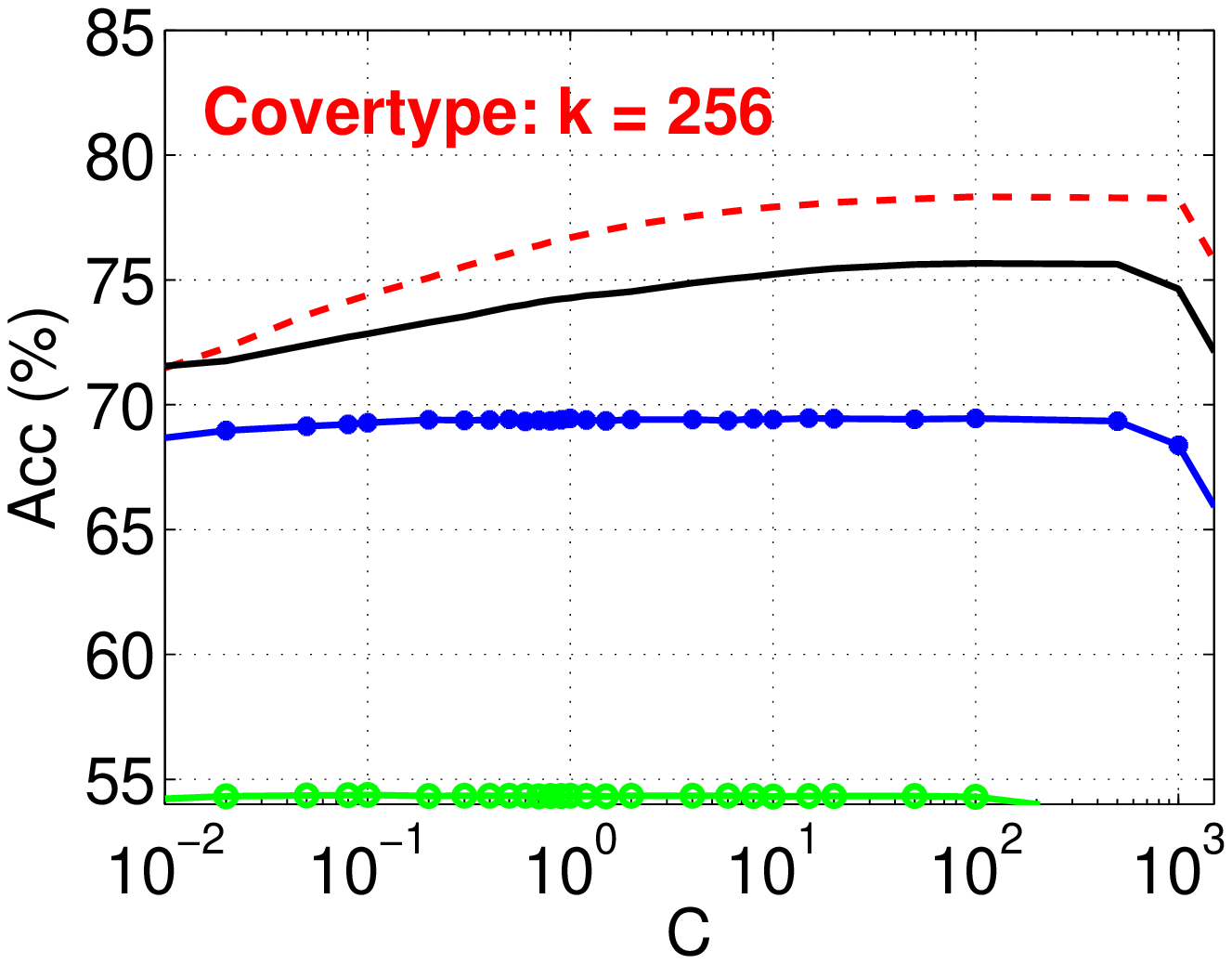}
}

\mbox{
\includegraphics[width=2.3in]{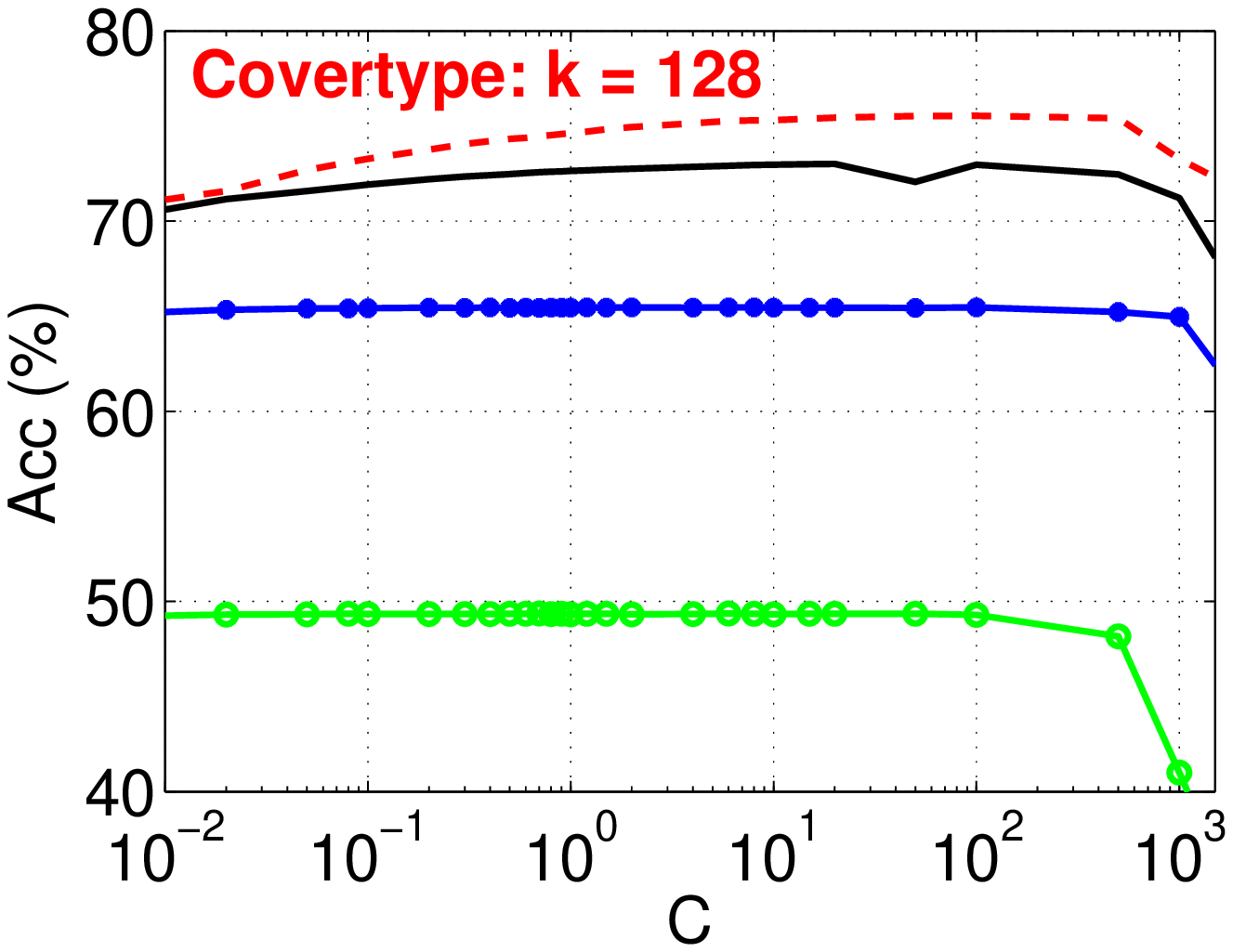}\hspace{-0.12in}
\includegraphics[width=2.3in]{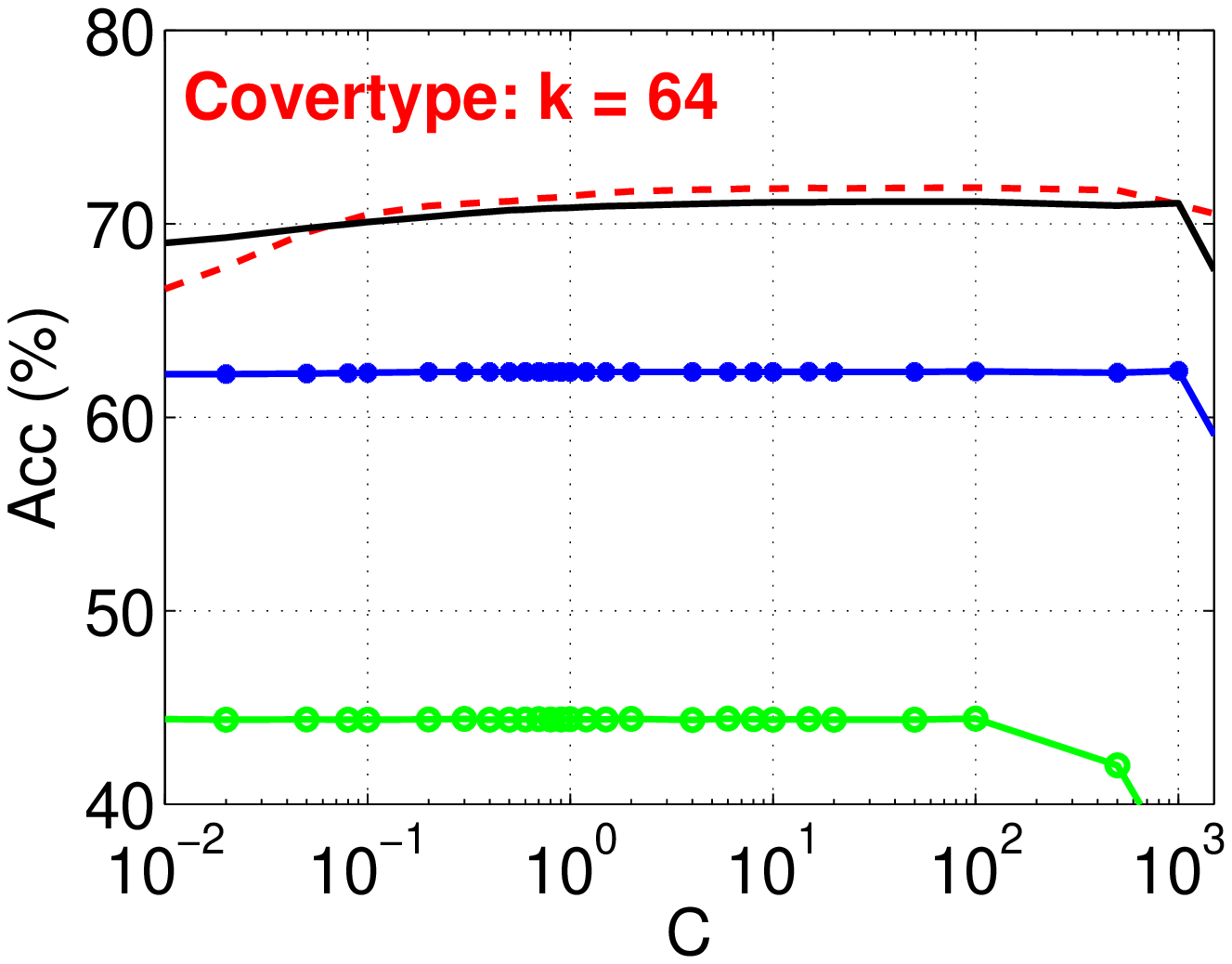}\hspace{-0.12in}
\includegraphics[width=2.3in]{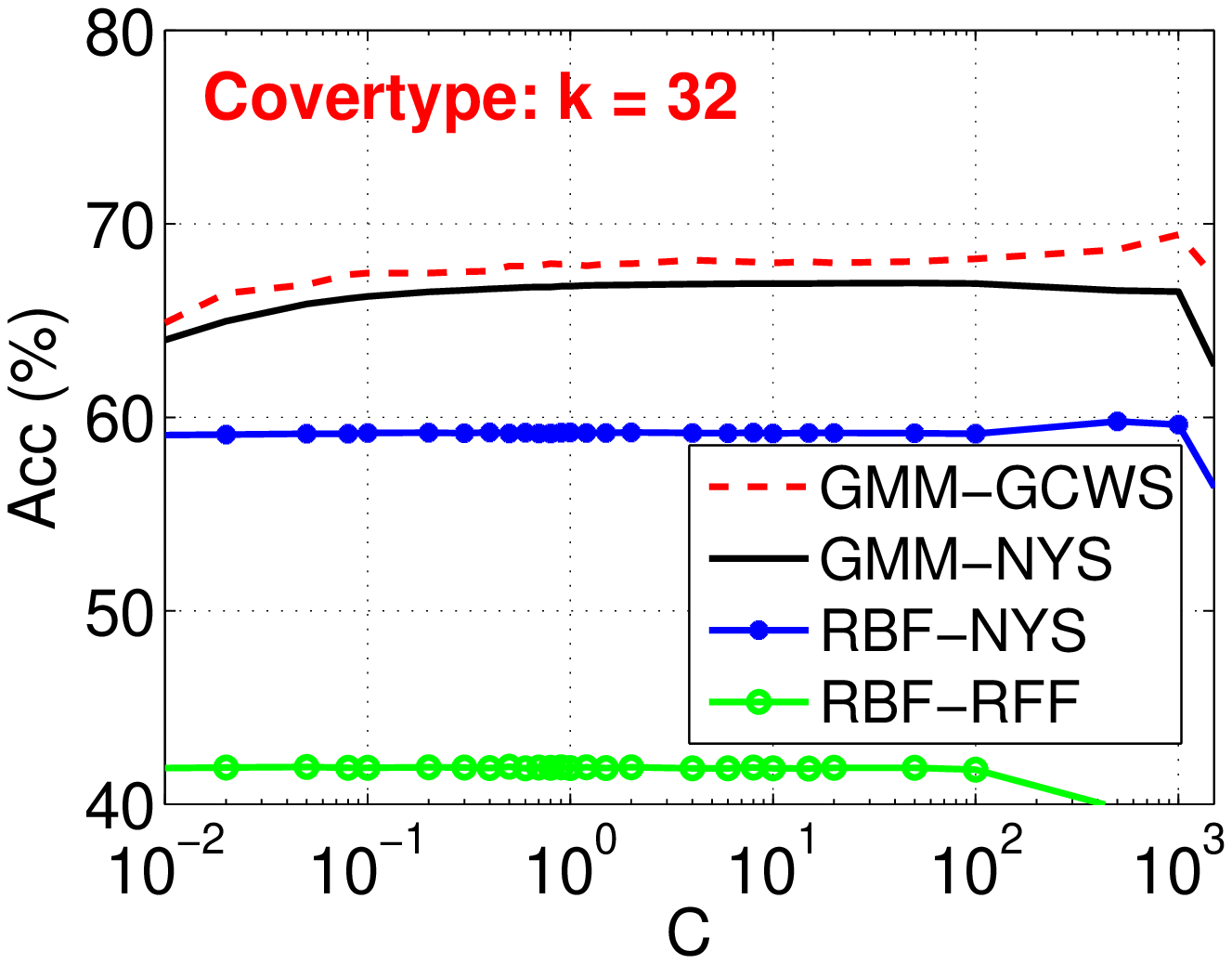}
}
\end{center}
\vspace{-0.3in}
\caption{\textbf{Covertype:}\ Test classification accuracies for 6  $k$ values and 4 different  algorithms.}\label{fig_Covertype}
\end{figure}

Figures~\ref{fig_SensIT}, \ref{fig_Webspam}, \ref{fig_PAMAP105}, \ref{fig_PAMAP101}, \ref{fig_Covertype} report the test classification accuracies for {\em SensIT}, {\em Webspam}, {\em PAMAP105}, {\em PAMAP101}, {\em Covertype}, respectively. As these datasets are fairly large, we could not report nonlinear kernel (GMM and RBF) SVM results, although we can still report linear SVM results; see Table~\ref{tab_data}. Again, these figures confirm that 1) GMM-NYS is substantially more accurate than RBF-RFF; and 2) GMM-NYS becomes more accurate than linear SVM once $k$ is large enough. \\

\begin{figure}
\begin{center}
\mbox{
\includegraphics[width=2.3in]{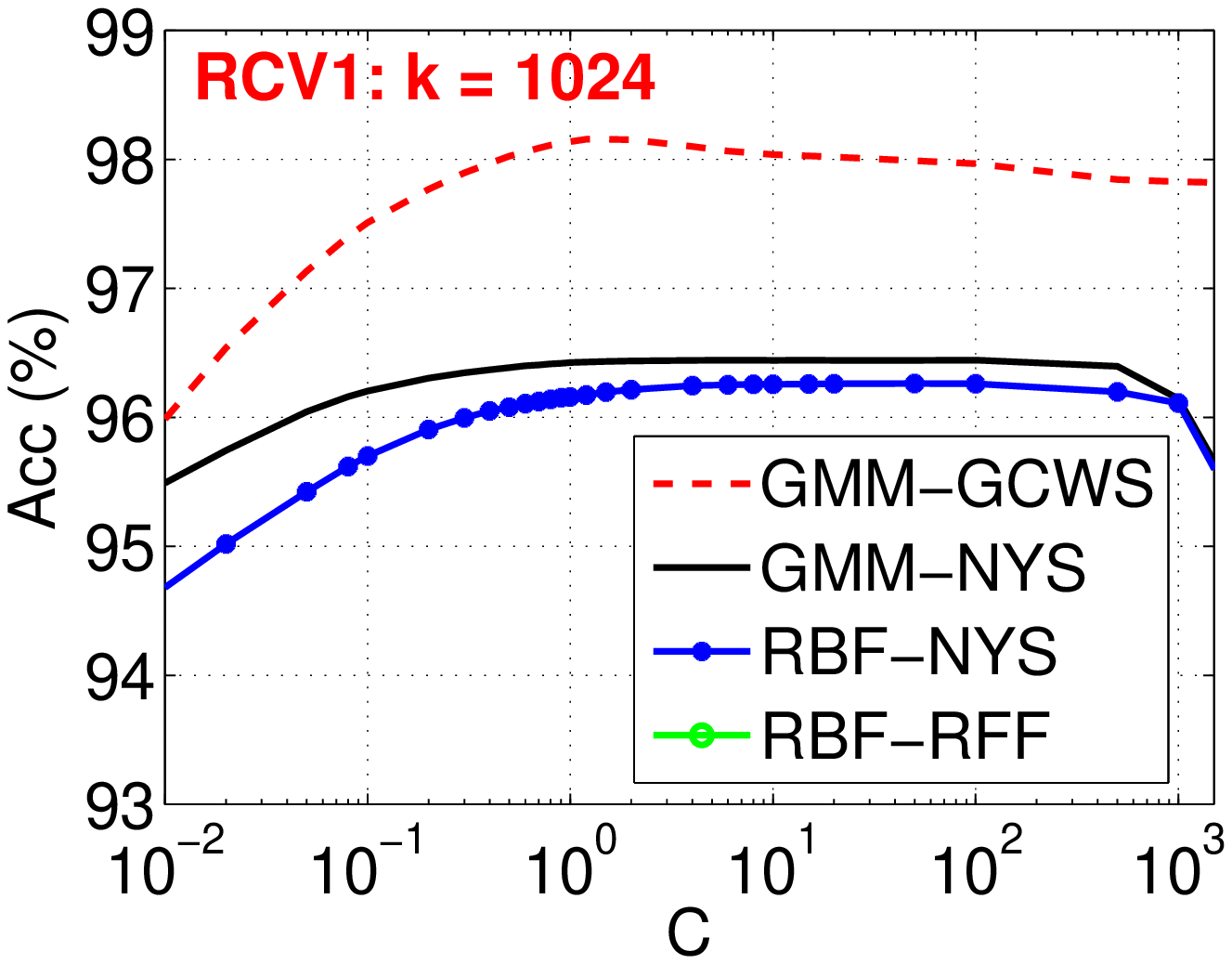}\hspace{-0.12in}
\includegraphics[width=2.3in]{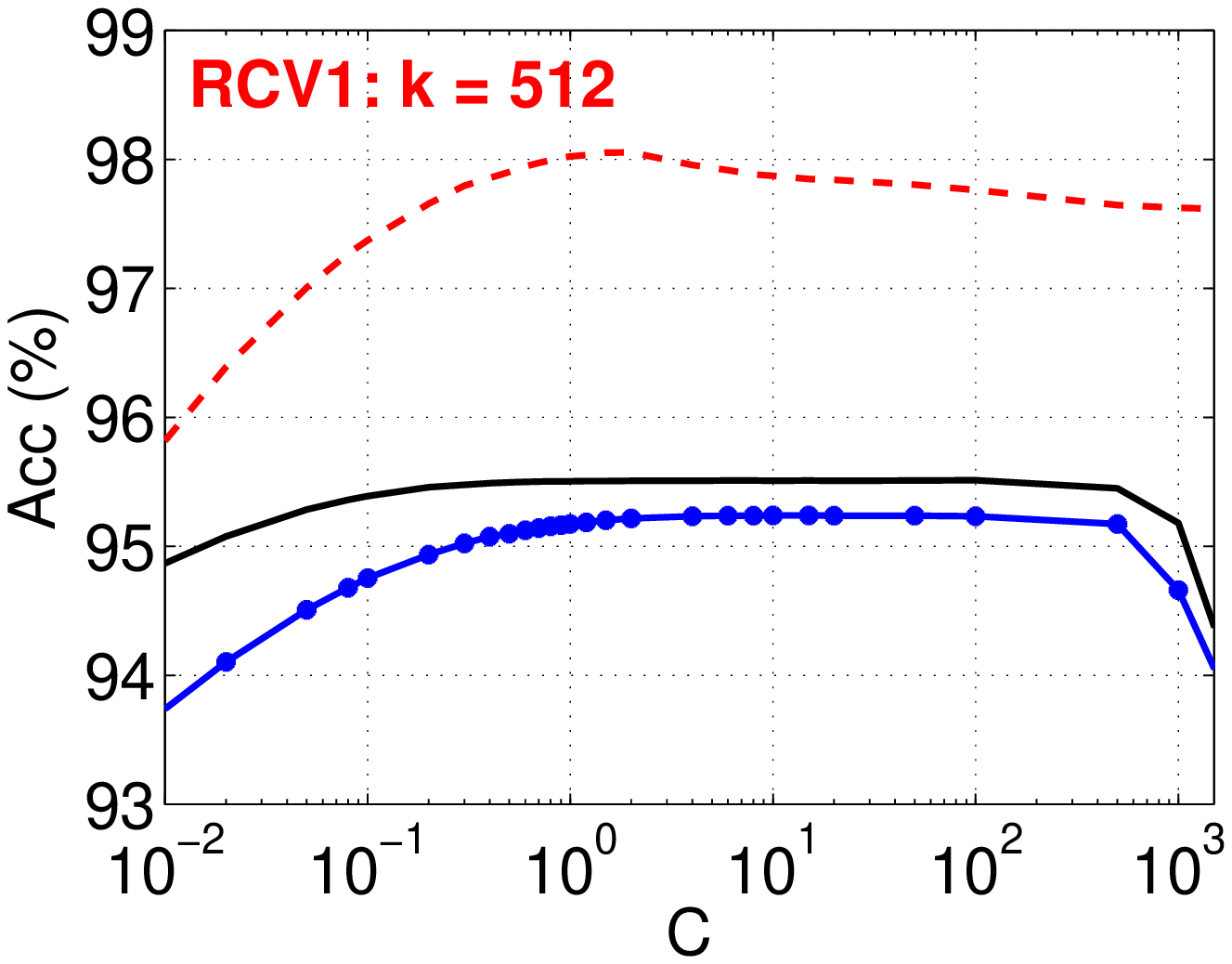}\hspace{-0.12in}
\includegraphics[width=2.3in]{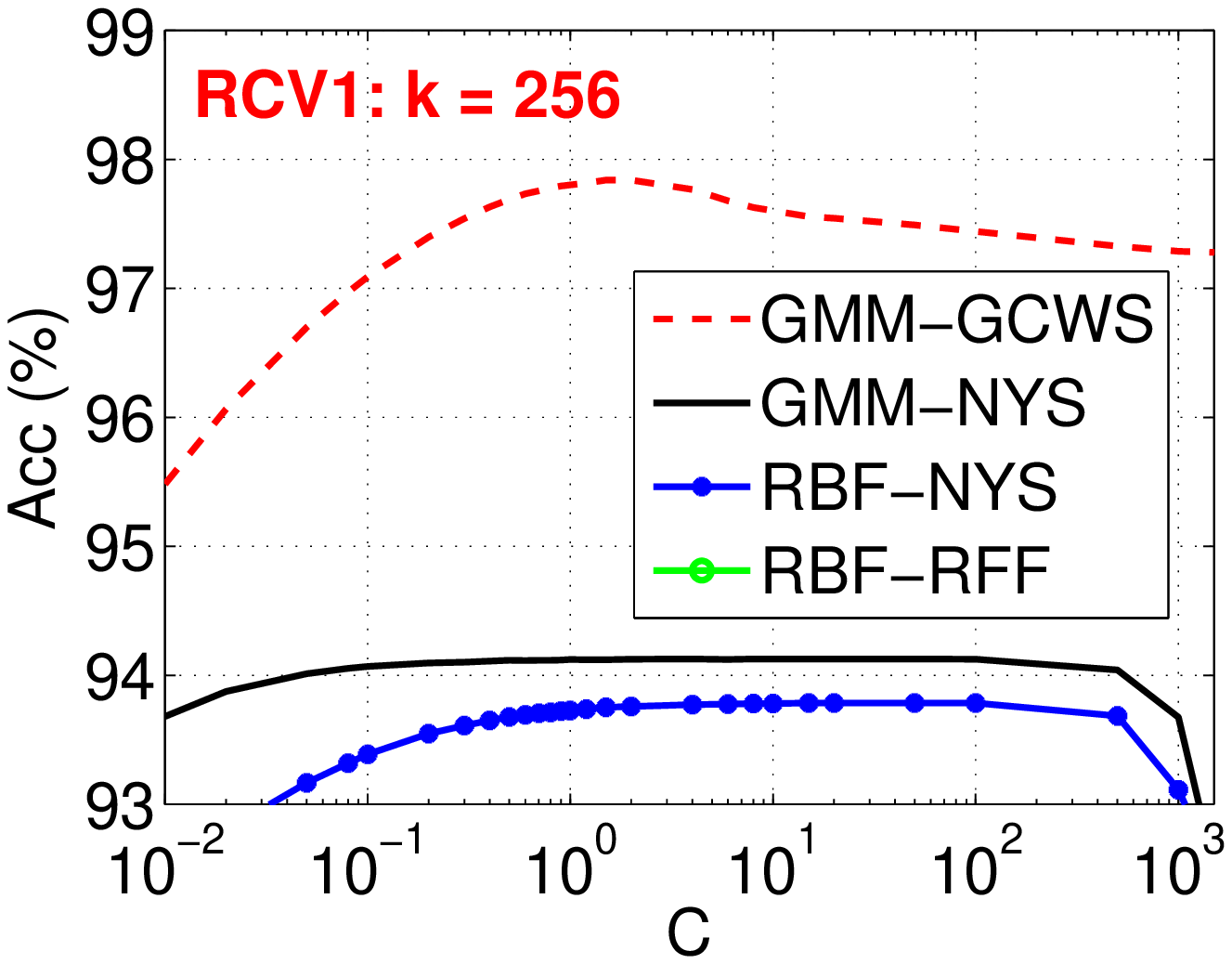}
}

\mbox{
\includegraphics[width=2.3in]{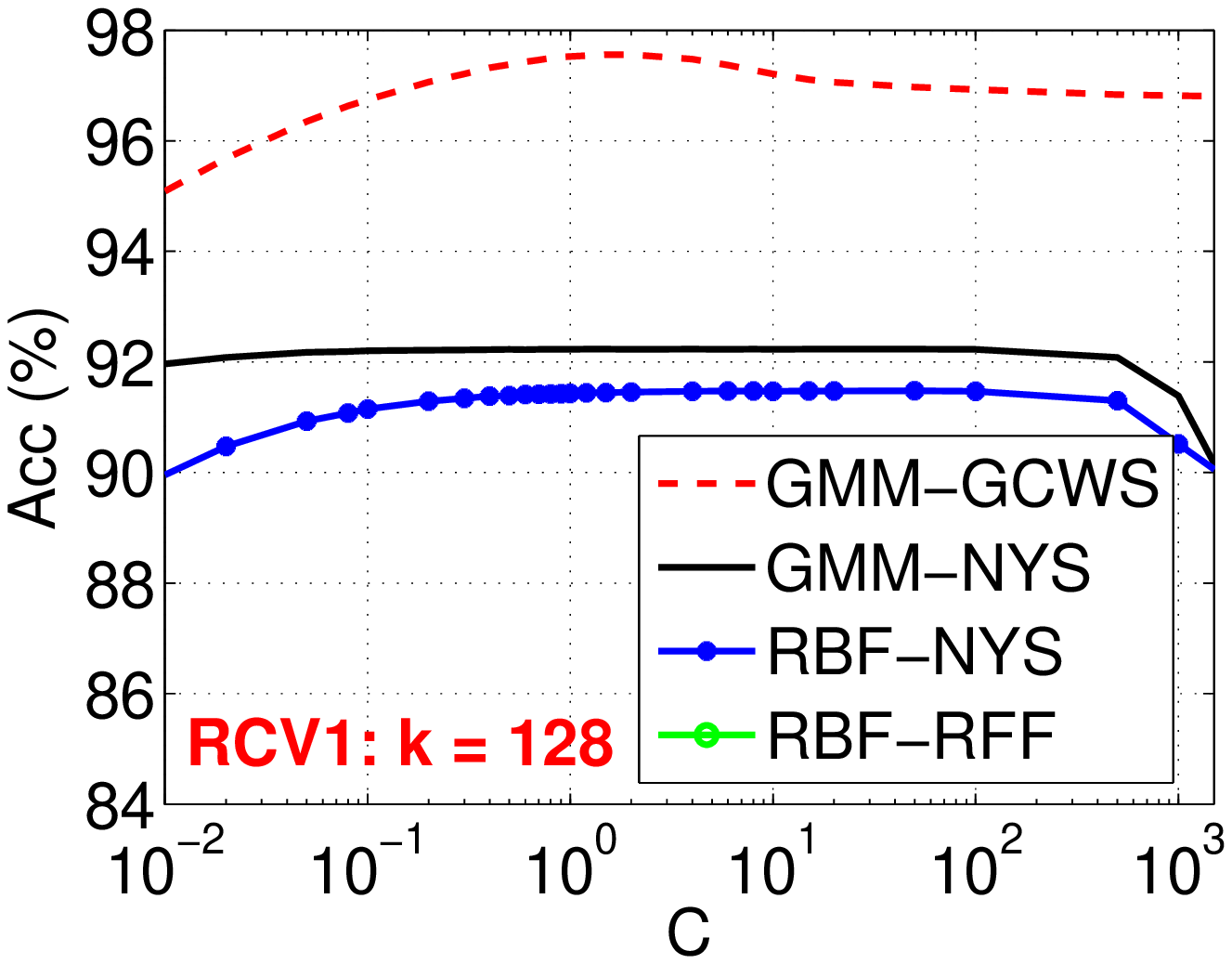}\hspace{-0.12in}
\includegraphics[width=2.3in]{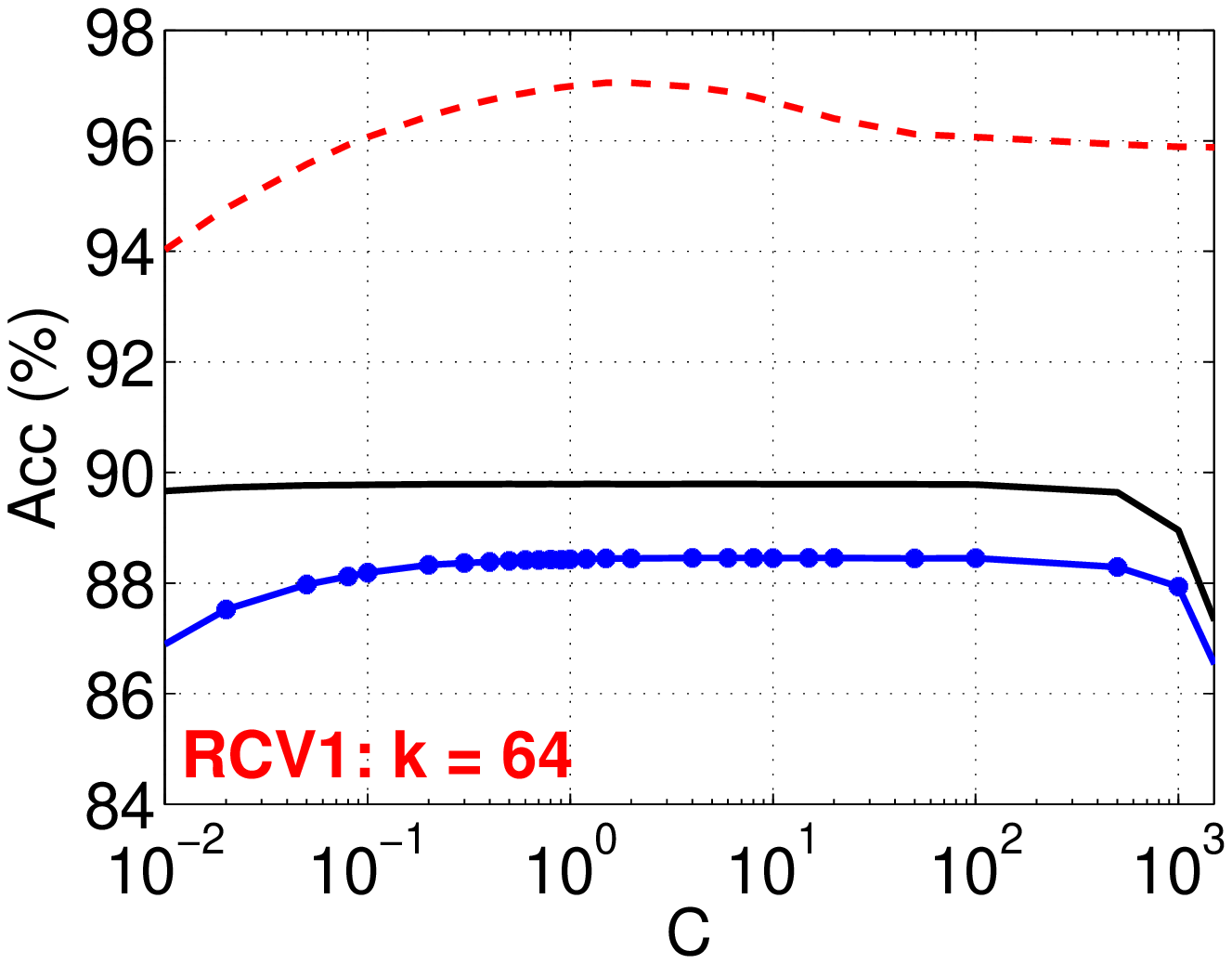}\hspace{-0.12in}
\includegraphics[width=2.3in]{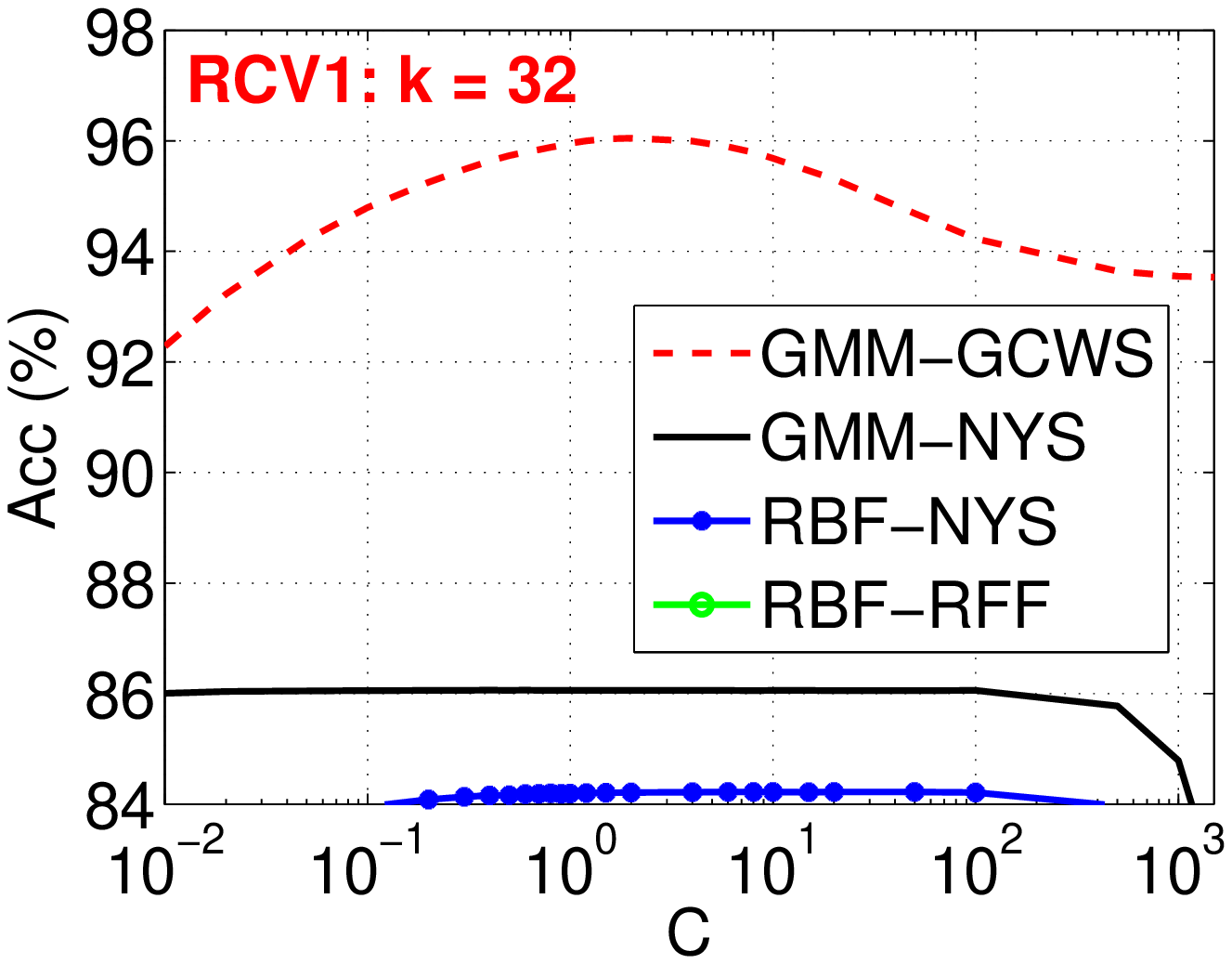}
}
\end{center}
\vspace{-0.3in}
\caption{\textbf{RCV1:} \ Test classification accuracies for 6  $k$ values. For better clarify we did not display the results for RBF-RFF because they are much worse than the results of other methods. See Figure~\ref{fig_RCV1_2} for the results of RBF-RFF. }\label{fig_RCV1_1}
\end{figure}

\begin{figure}
\begin{center}
\mbox{
\includegraphics[width=2.3in]{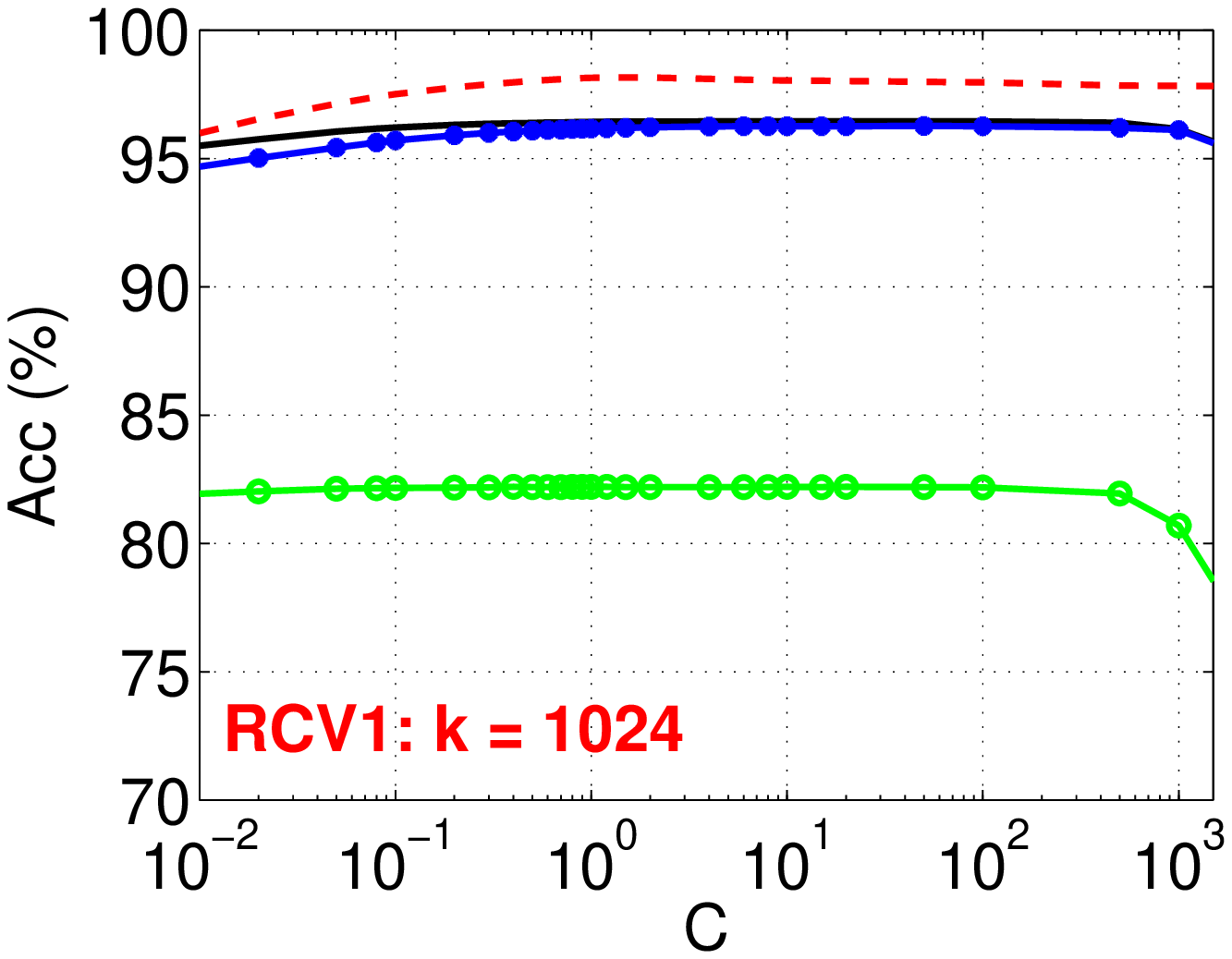}\hspace{-0.12in}
\includegraphics[width=2.3in]{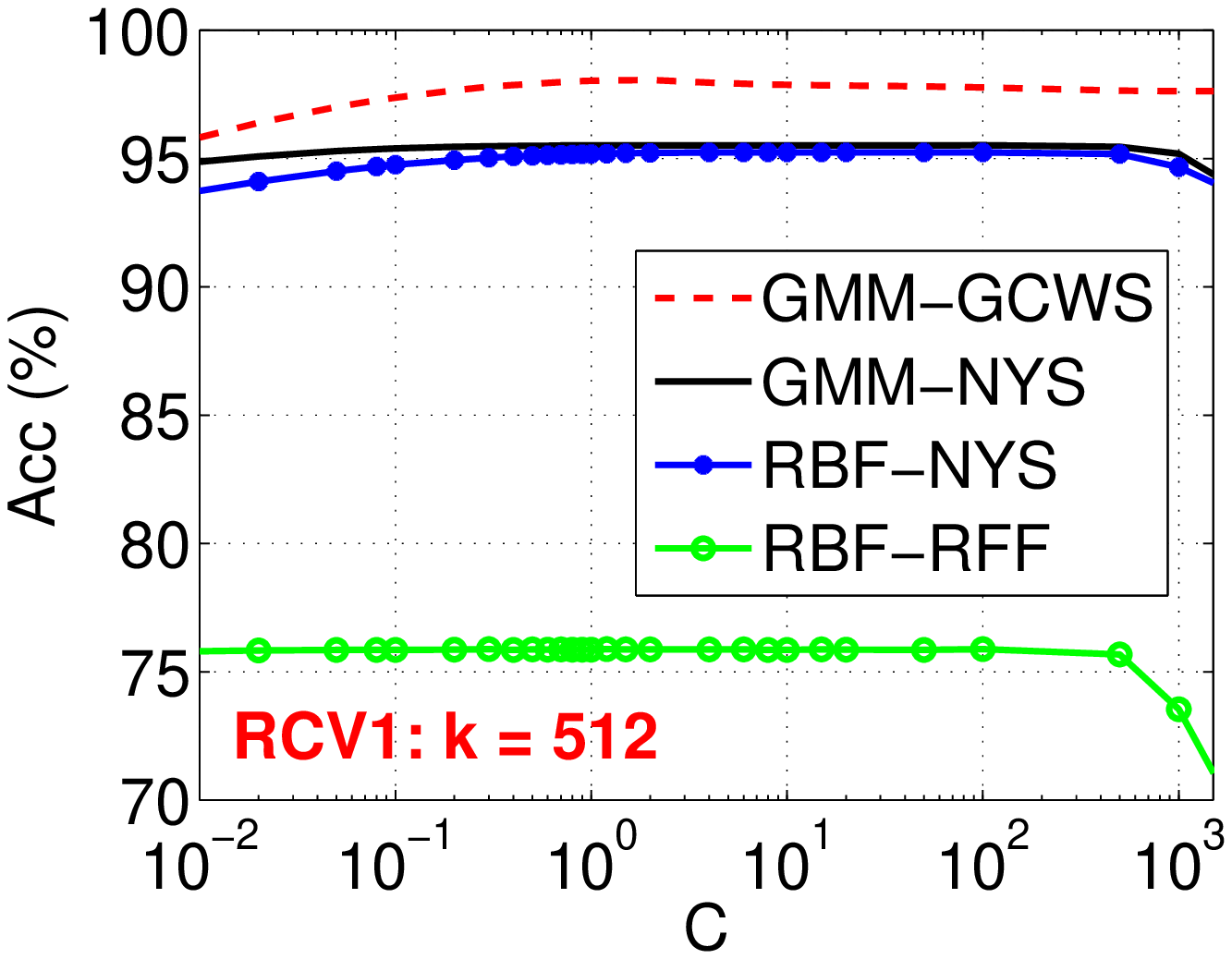}\hspace{-0.12in}
\includegraphics[width=2.3in]{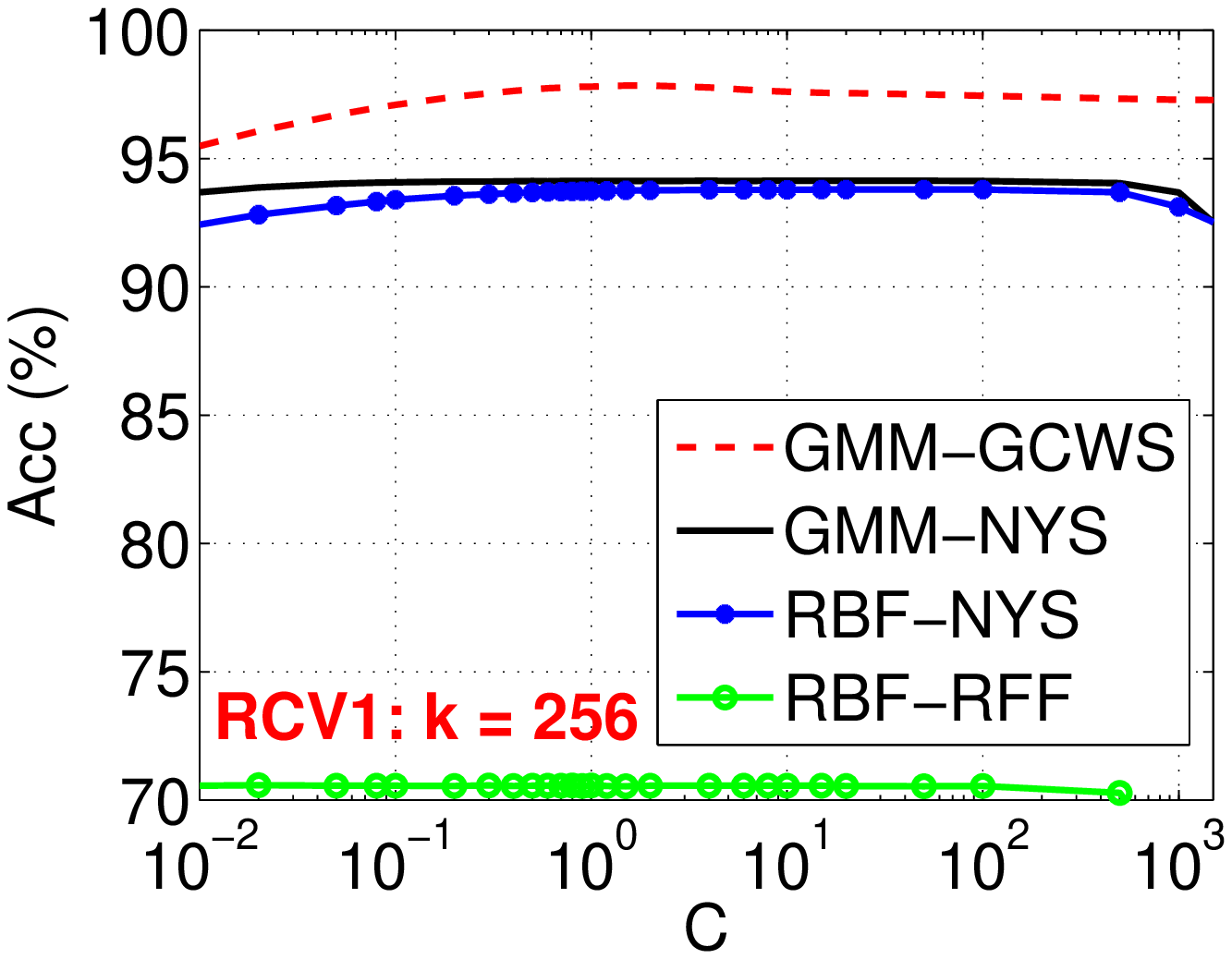}
}

\mbox{
\includegraphics[width=2.3in]{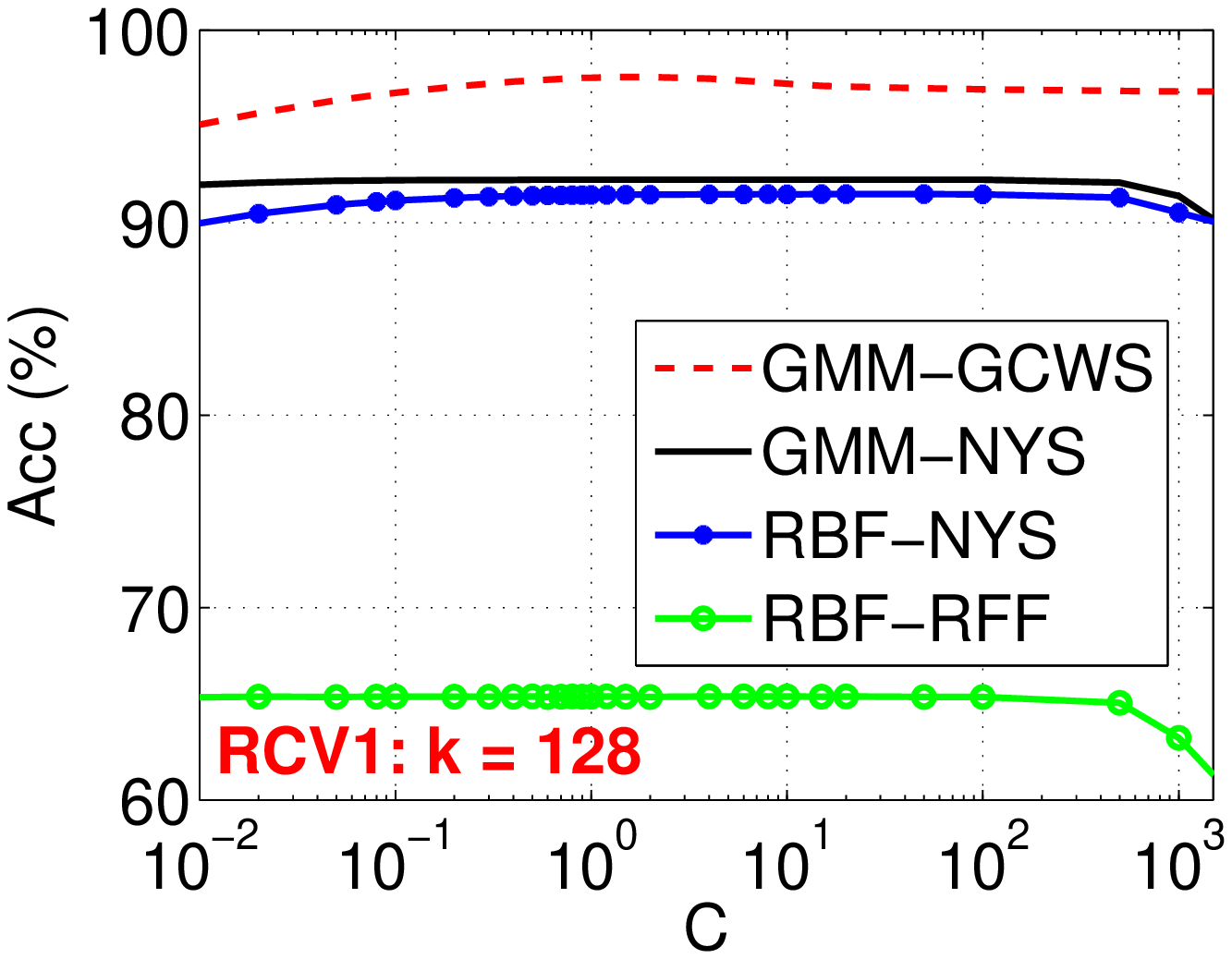}\hspace{-0.12in}
\includegraphics[width=2.3in]{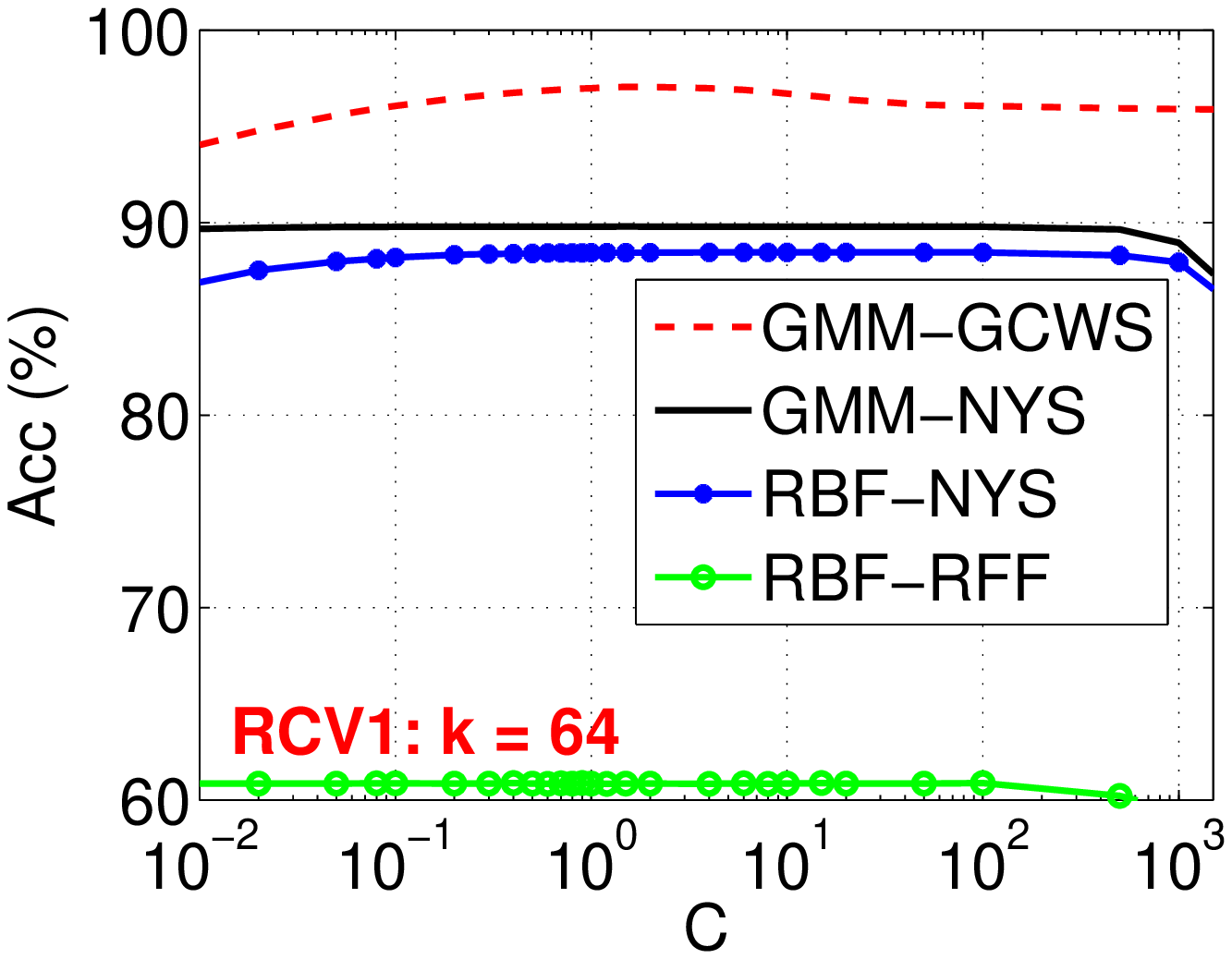}\hspace{-0.12in}
\includegraphics[width=2.3in]{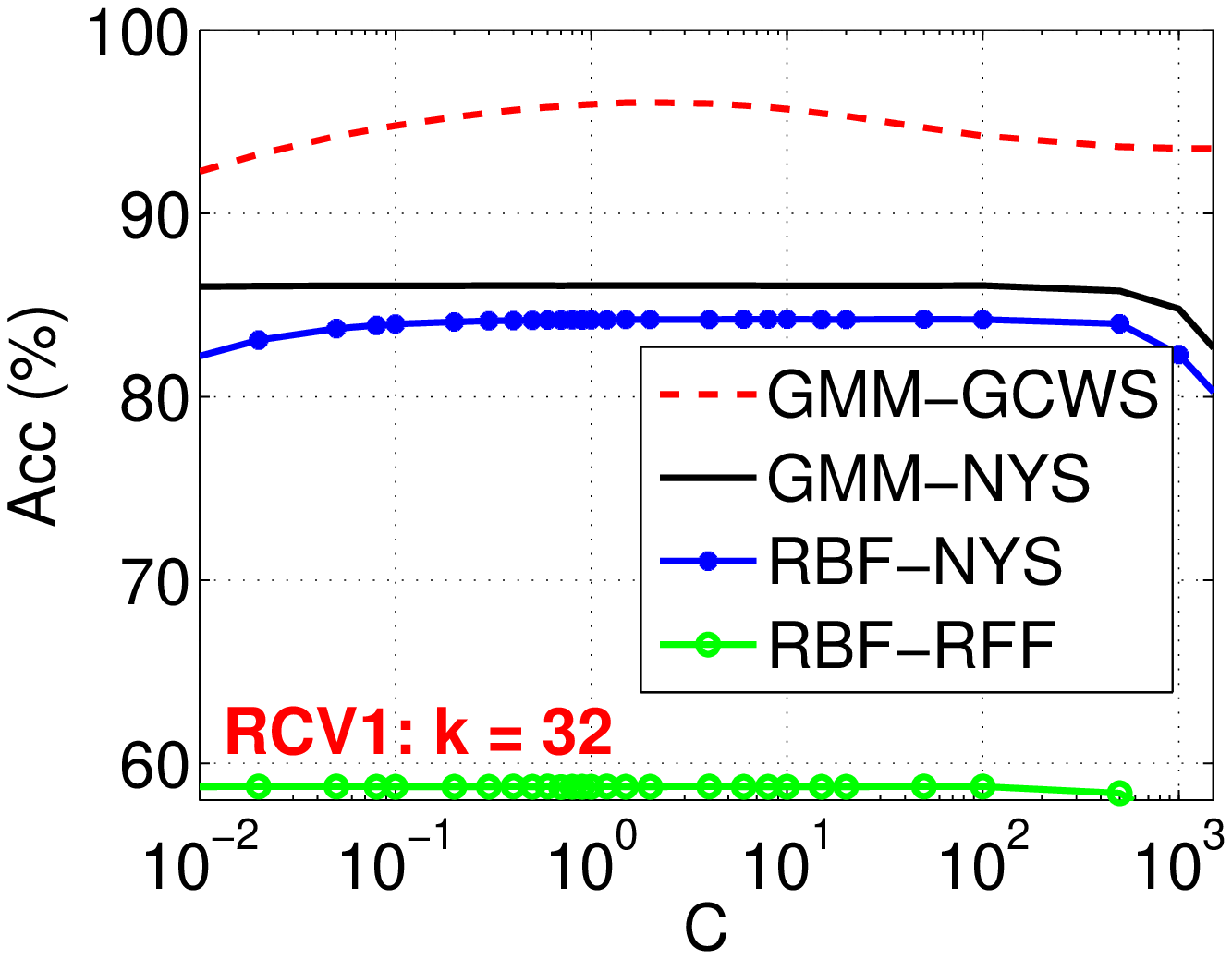}
}
\end{center}
\vspace{-0.3in}
\caption{\textbf{RCV1:}\ Test classification accuracies for 6 $k$ values and 4 different  algorithms.}\label{fig_RCV1_2}
\end{figure}

\begin{figure}[h!]
\begin{center}
\mbox{
\includegraphics[width=2.3in]{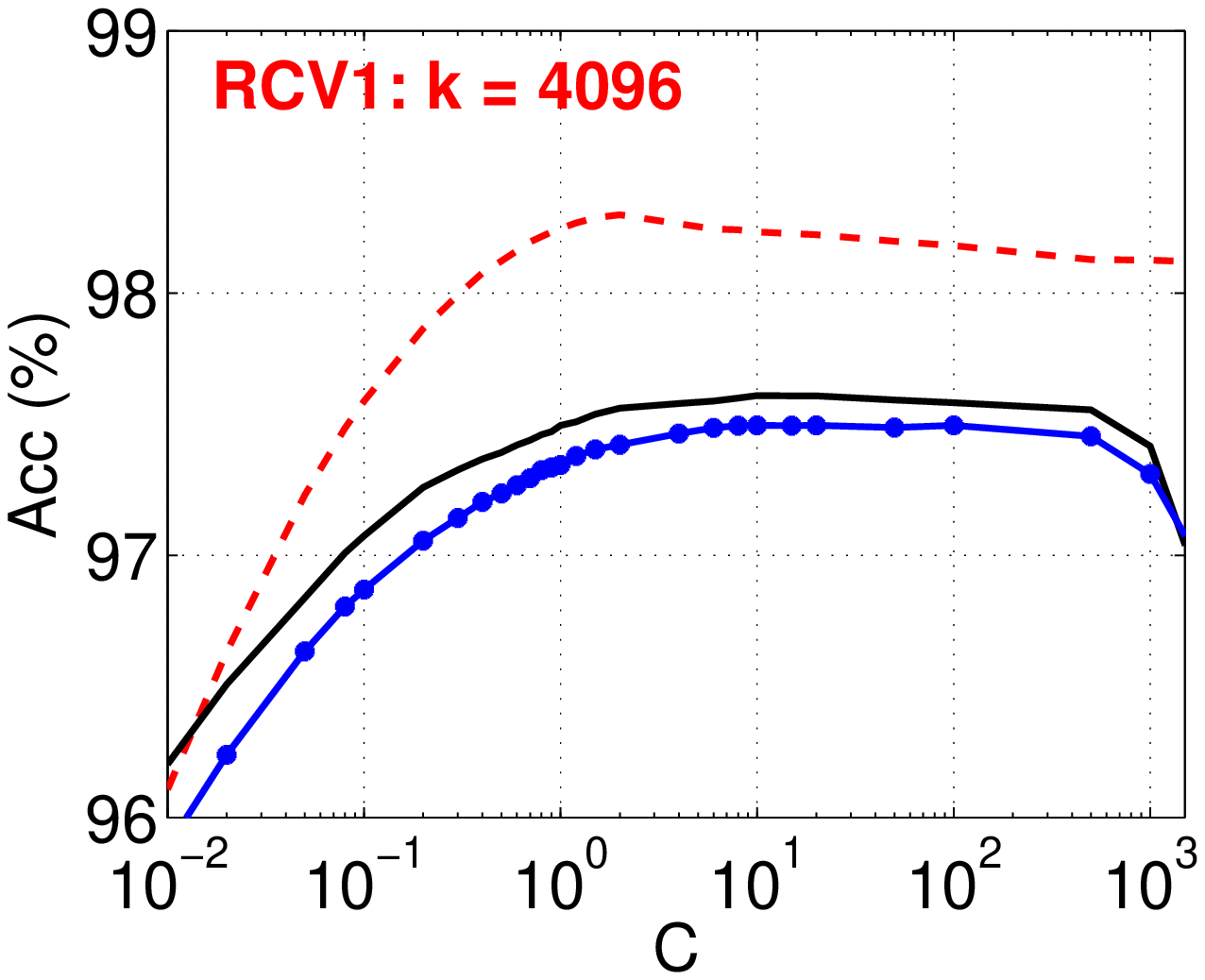}\hspace{-0.12in}
\includegraphics[width=2.3in]{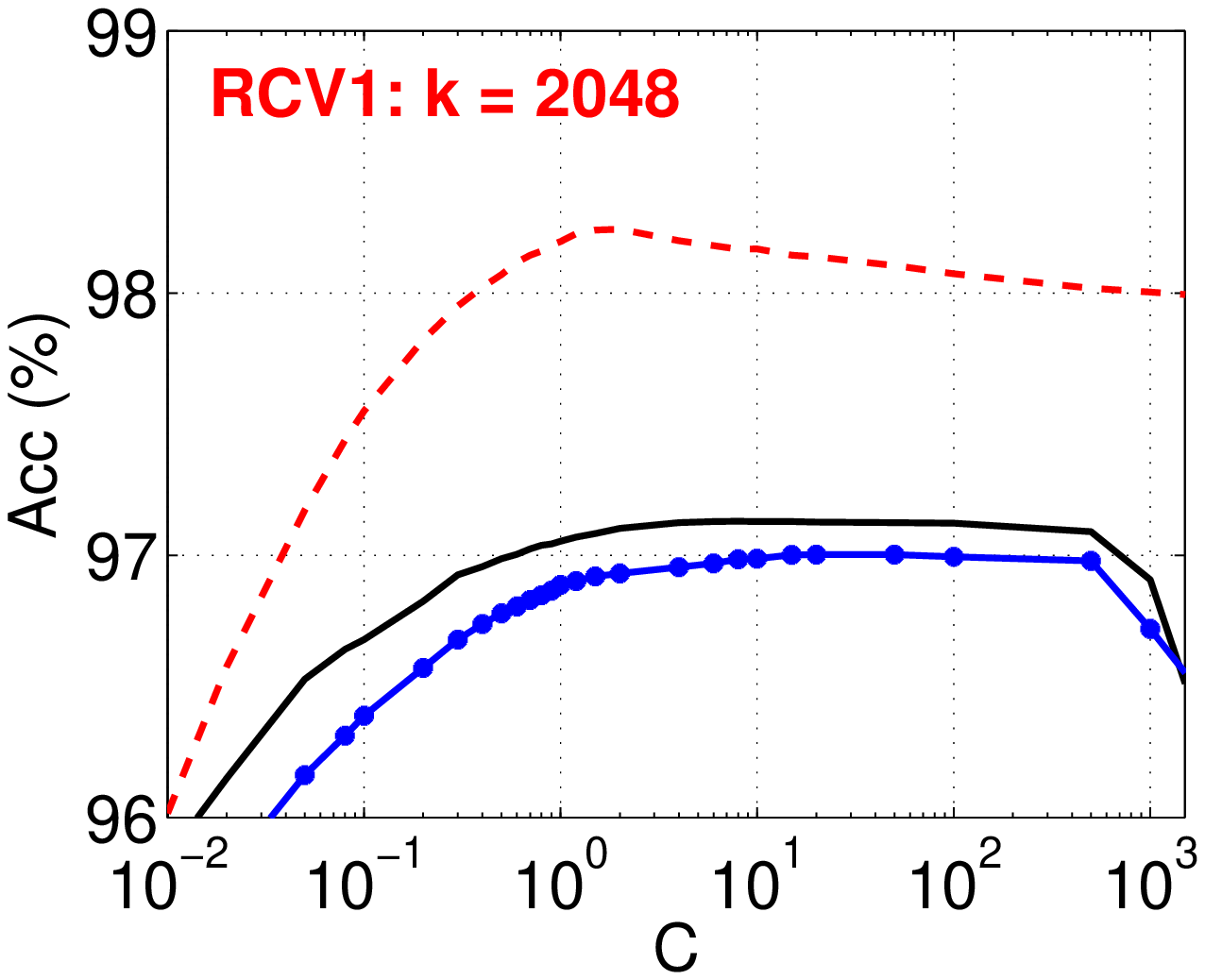}\hspace{-0.12in}
\includegraphics[width=2.3in]{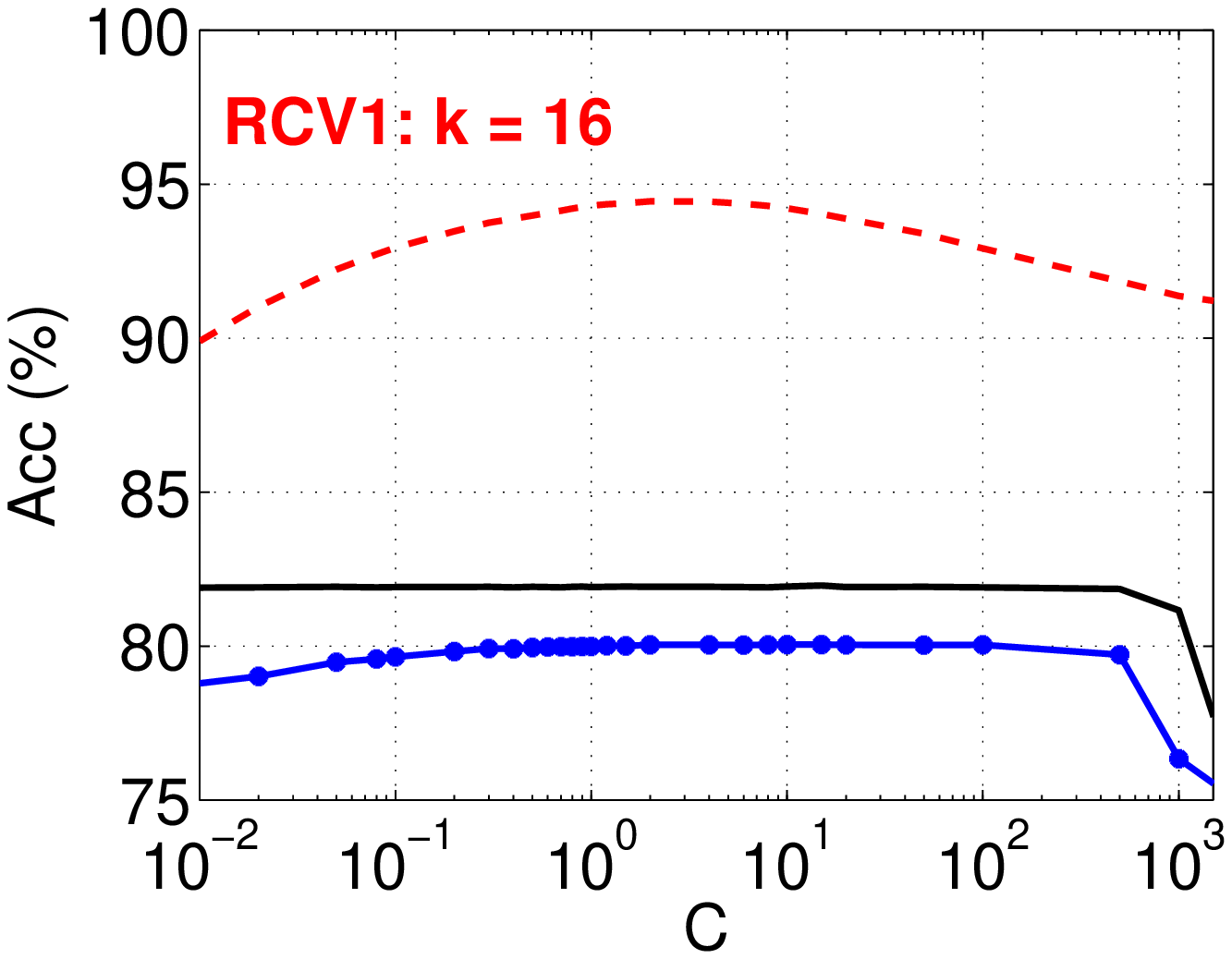}
}

\mbox{
\includegraphics[width=2.3in]{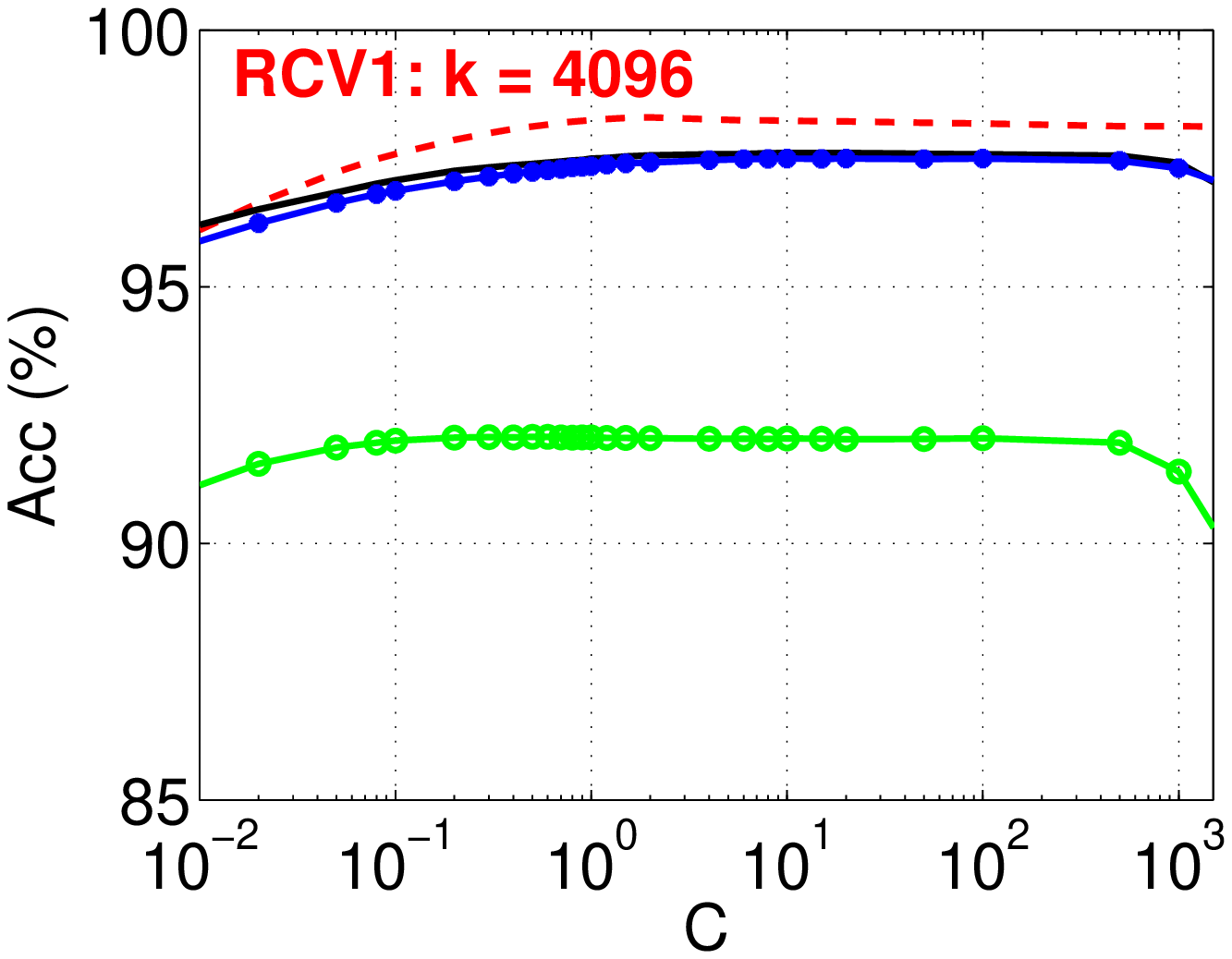}\hspace{-0.12in}
\includegraphics[width=2.3in]{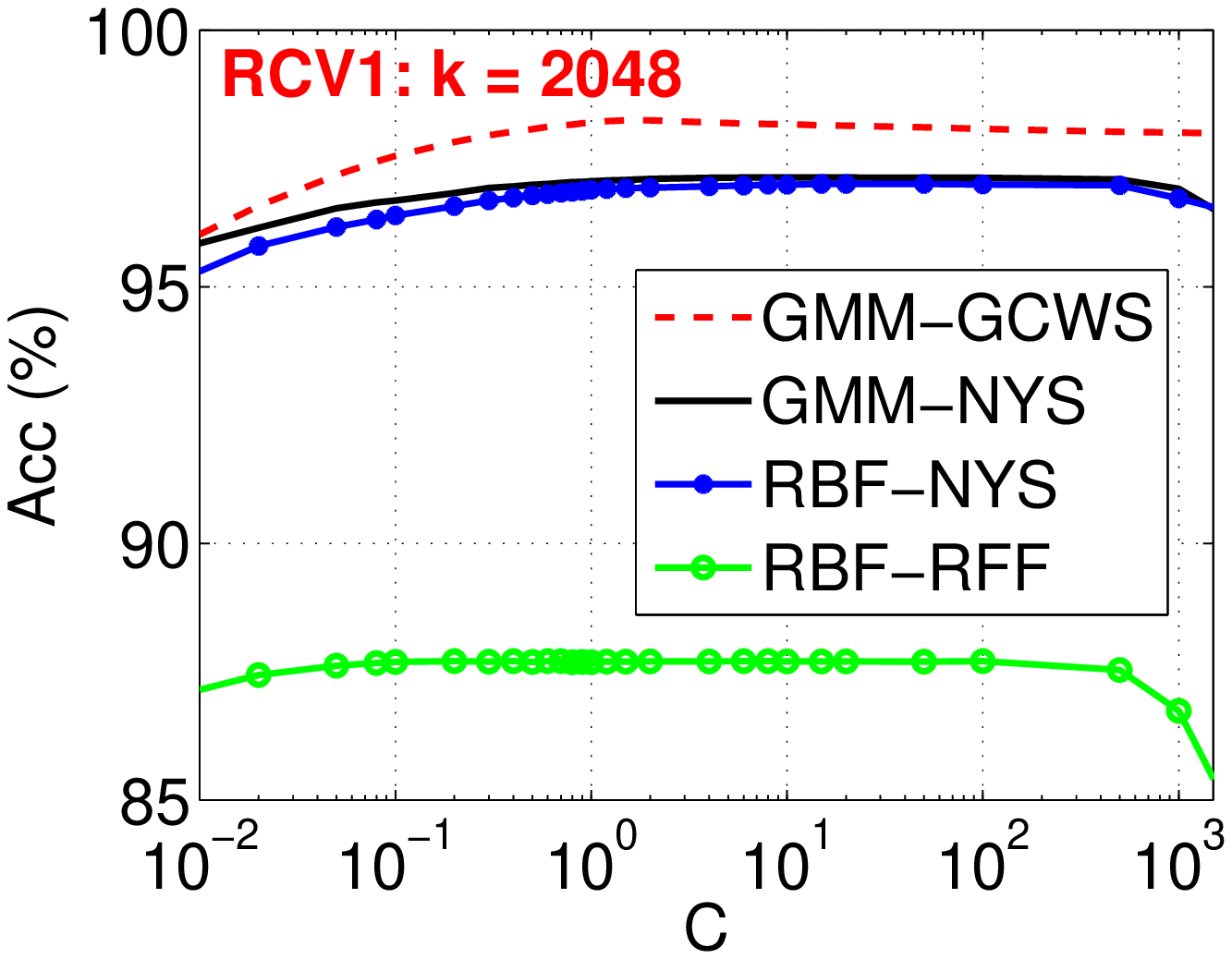}\hspace{-0.12in}
\includegraphics[width=2.3in]{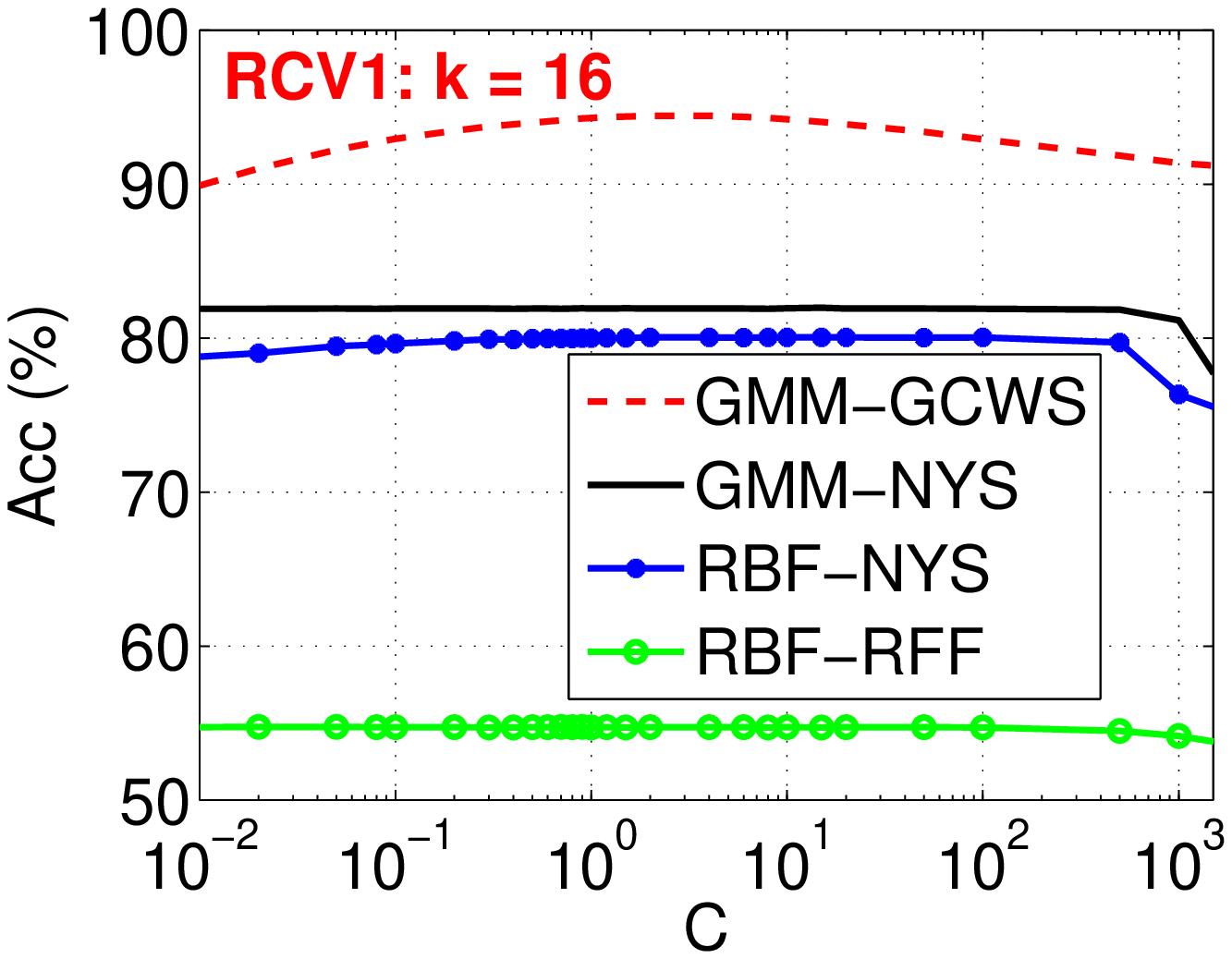}
}
\end{center}
\vspace{-0.3in}
\caption{\textbf{RCV1:}\ Test classification accuracies for $k\in\{16, 2048, 4096\}$ and 4 different  algorithms.}\label{fig_RCV1_3}
\end{figure}

Finally, Figures~\ref{fig_RCV1_1}, ~\ref{fig_RCV1_2}, and ~\ref{fig_RCV1_3} report the test classification results on the {\em RCV1} dataset. Because the performance of RBF-RFF is so much worse than other methods, we report in Figure~\ref{fig_RCV1_1} only the results for GMM-GCWS, GMM-NYS, and RBF-NYS, for better clarity. In addition, we report the  results for $k\in\{4096, 2048, 16\}$ in Figure~\ref{fig_RCV1_3}, to provide a more comprehensive comparison study. All these results confirm that GMM-NYS is a strong competitor of RBF-RFF.

\section{Conclusion}

The recently proposed GMM kernel has  proven effective as a measure of data similarity, through extensive experiments in the prior work~\cite{Report:Li_GMM16}. For large-scale machine learning, it is crucial to be able to linearize nonlinear kernels. The  work~\cite{Report:Li_GMM16} demonstrated that the GCWS hashing method for linearizing the GMM kernel (GMM-GCWS) typically produces substantially more accurate results than the well-known random Fourier feature (RFF) approach for linearizing the RBF kernel (RBF-RFF). \\

In this study, we apply the general and well-known Nystrom method for approximating the GMM kernel (GMM-NYS) and we show, through extensive experiments, that the results produced by GMM-NYS are substantially more accurate than the results obtained using RBF-RFF. This phenomenon is largely expected because random projection based algorithms often have much larger variances than sampling based methods~\cite{Proc:Li_Church_Hastie_NIPS06}.
{
\bibliographystyle{abbrv}
\bibliography{../bib/mybibfile}

}

\end{document}